\documentclass[letterpaper,english,smallextended]{svjour3}
\usepackage[T1]{fontenc}
\usepackage[latin9]{inputenc}
\usepackage{color}
\usepackage{url}
\usepackage{amsmath}
\usepackage{amssymb}
\usepackage{graphicx}
\usepackage{setspace}
\makeatletter

\pdfpageheight\paperheight
\pdfpagewidth\paperwidth

\RequirePackage{fix-cm}

\smartqed  

\usepackage{bbm}
\DeclareMathOperator*{\argmin}{arg\,min}

\newtheorem{condition}{Condition}

\makeatother

\usepackage{babel}

\begin{document}
\title{Variational approach for learning Markov processes from time series
data\thanks{This work was funded by Deutsche Forschungsgemeinschaft (SFB 1114/A4)
and European Research Commission (ERC StG 307494 \textquotedblleft pcCell\textquotedblright ).}}
\titlerunning{Variational approach for Markov processes}
\author{Hao Wu$^{1,2,\text{a)}}$ \and Frank Noé$^{2,3,4,\text{b)}}$}
\authorrunning{H. Wu and F. Noé}
\institute{1: Tongji University, School of Mathematical Sciences, 1239 Siping
Road, 200092 Shanghai, China\\
2: Freie Universität Berlin, Department of Mathematics and Computer
Science, Arnimallee 6, 14195 Berlin, Germany\\
3: Freie Universität Berlin, Department of Physics, Arnimallee 14,
14195 Berlin, Germany\\
4: Rice University, Department of Chemistry, Houston, Texas 77005,
United States\\
\\
Corresponding authors: a) hwu@tongji.edu.cn b) frank.noe@fu-berlin.de\\
}
\date{Received: date / Accepted: date}
\maketitle
\begin{abstract}
Inference, prediction and control of complex dynamical systems from
time series is important in many areas, including financial markets,
power grid management, climate and weather modeling, or molecular
dynamics. The analysis of such highly nonlinear dynamical systems
is facilitated by the fact that we can often find a (generally nonlinear)
transformation of the system coordinates to features in which the
dynamics can be excellently approximated by a linear Markovian model.
Moreover, the large number of system variables often change collectively
on large time- and length-scales, facilitating a low-dimensional analysis
in feature space. In this paper, we introduce a variational approach
for Markov processes (VAMP) that allows us to find optimal feature
mappings and optimal Markovian models of the dynamics from given time
series data. The key insight is that the best linear model can be
obtained from the top singular components of the Koopman operator.
This leads to the definition of a family of score functions called
VAMP-$r$ which can be calculated from data, and can be employed to
optimize a Markovian model. In addition, based on the relationship
between the variational scores and approximation errors of Koopman
operators, we propose a new VAMP-E score, which can be applied to
cross-validation for hyper-parameter optimization and model selection
in VAMP. VAMP is valid for both reversible and nonreversible processes
and for stationary and non-stationary processes or realizations. 
\end{abstract}

\section{Introduction\label{sec:Introduction}}

Extracting dynamical models and their main characteristics from time
series data is a recurring problem in many areas of science and engineering.
In the particularly popular approach of Markovian models, the future
evolution of the system, e.g. state $\mathbf{x}_{t+\tau}$, only depends
on the current state $\mathbf{x}_{t}$, where $t$ is the time step
and $\tau$ is the delay or lag time. Markovian models are easier
to analyze than models with explicit memory terms. They are justified
by the fact that many physical processes \textendash{} including both
deterministic and stochastic processes \textendash{} are inherently
Markovian. Even when only a subset of the variables in which the system
is Markovian are observed, a variety of physics and engineering processes
have been shown to be accurately modeled by Markovian models on sufficiently
long observation lag times $\tau$. Examples include molecular dynamics
\cite{ChoderaNoe_COSB14_MSMs,PrinzEtAl_JCP10_MSM1}, wireless communications
\cite{konrad2001markov,ma2001composite} and fluid dynamics \cite{mezic2013analysis,froyland2016optimal}.

In the past decades, a collection of closely related Markov modeling
methods were developed in different fields, including Markov state
models (MSMs) \cite{SchuetteFischerHuisingaDeuflhard_JCompPhys151_146,PrinzEtAl_JCP10_MSM1,BowmanPandeNoe_MSMBook},
Markov transition models \cite{WuNoe_JCP15_GMTM}, Ulam's Galerkin
method \cite{dellnitz2001algorithms,bollt2013applied,froyland2014computational},
blind-source separation \cite{Molgedey_94,ZieheMueller_ICANN98_TDSEP},
the variational approach of conformation dynamics (VAC) \cite{NoeNueske_MMS13_VariationalApproach,NueskeEtAl_JCTC14_Variational},
time-lagged independent component analysis (TICA) \cite{PerezEtAl_JCP13_TICA,SchwantesPande_JCTC13_TICA},
dynamic mode decomposition (DMD) \cite{RowleyEtAl_JFM09_DMDSpectral,Schmid_JFM10_DMD,TuEtAl_JCD14_ExactDMD},
extended dynamic mode decomposition (EDMD) \cite{WilliamsKevrekidisRowley_JNS15_EDMD},
variational Koopman models \cite{WuEtAl_JCP17_VariationalKoopman},
variational diffusion maps \cite{BoninsegnaEtAl_JCTC15_VariationalDM},
sparse identification of nonlinear dynamics \cite{brunton2016discovering}
and corresponding kernel embeddings \cite{harmeling2003kernel,song2013kernel,SchwantesPande_JCTC15_kTICA}
and tensor formulations \cite{NueskeEtAl_JCP15_Tensor,KlusSchuette_Arxiv15_Tensor}.
All these models approximate the Markov dynamics at a lag time $\tau$
by a linear model in the following form:
\begin{equation}
\mathbb{E}\left[\mathbf{g}(\mathbf{x}_{t+\tau})\right]=\mathbf{K}^{\top}\mathbb{E}\left[\mathbf{f}(\mathbf{x}_{t})\right].\label{eq:linear_model}
\end{equation}
Here $\mathbf{f}(\mathbf{x})=(f_{1}(\mathbf{x}),f_{2}(\mathbf{x}),...)^{\top}$
and $\mathbf{g}(\mathbf{x})=(g_{1}(\mathbf{x}),g_{2}(\mathbf{x}),...)^{\top}$
are feature transformations that transform the state variables $\mathbf{x}$
into the feature space in which the dynamics are approximately linear.
$\mathbb{E}$ denotes an expectation value over time that accounts
for stochasticity in the dynamics, and can be omitted for deterministic
dynamical systems. In some methods, such as DMD, the feature transformation
is an identity transformation: $\mathbf{f}\left(\mathbf{x}\right)=\mathbf{g}\left(\mathbf{x}\right)=\mathbf{x}$
\textendash{} and then Eq. \eqref{eq:linear_model} defines a linear
dynamical system in the original state variables. If $\text{\ensuremath{\mathbf{f}}}$
and $\mathbf{g}$ are indicator functions that partition $\Omega$
into substates, such that $f_{i}\left(\mathbf{x}\right)=g_{i}\left(\mathbf{x}\right)=1$
if $\mathbf{x}\in A_{i}$ and $0$ otherwise, Eq. \eqref{eq:linear_model}
is the propagation law of an MSM, or equivalently of Ulam's Galerkin
method, as the expectation values $\mathbb{E}\left[\mathbf{f}(\mathbf{x}_{t})\right]$
and $\mathbb{E}\left[\mathbf{g}(\mathbf{x}_{t+\tau})\right]$ represent
the vector of probabilities to be in any substate at times $t$ and
$t+\tau$, and $K_{ij}$ is the probability to transition from set
$A_{i}$ to set $A_{j}$ in time $\tau$. In general, \eqref{eq:linear_model}
can be interpreted as a finite-rank approximation of the so-called
Koopman operator \cite{Koopman_PNAS31_Koopman,Mezic_NonlinDyn05_Koopman},
which governs the time evolution of observables of the system state
and that can fully characterize the Markovian dynamics. As shown in
\cite{korda2017convergence}, this approximation becomes exact in
the limit of infinitely-sized feature transformations with $\mathbf{f}=\mathbf{g}$,
and a similar conclusion can also be obtained when $\mathbf{f},\mathbf{g}$
are infinite-dimensional feature functions deduced from a characteristic
kernel \cite{song2013kernel}.

A direct method to estimate the matrix $\mathbf{K}$ from data is
to solve the linear regression problem $\mathbf{g}(\mathbf{x}_{t+\tau})\approx\mathbf{K}^{\top}\mathbf{f}(\mathbf{x}_{t})$,
which facilitates the use of regularized solution methods, such as
the LASSO method \cite{tibshirani1996regression}. Alternatively,
feature functions $\mathbf{f}$ and $\mathbf{g}$ that allow Eq. \eqref{eq:linear_model}
to have a probabilistic interpretation (e.g. in MSMs), $\mathbf{K}$
can be estimated by a maximum-likelihood or Bayesian methods \cite{PrinzEtAl_JCP10_MSM1,Noe_JCP08_TSampling}. 

However, as yet, it is still unclear \emph{what are the optimal choices
for $\mathbf{f}$ and $\mathbf{g}$ - either given a fixed dimension
or a fixed amount of data}. Notice that this problem cannot be solved
by minimizing the regression error of Eq. \eqref{eq:linear_model},
because a regression error of zero can be trivially achieved by choosing
a completely uninformative model with $\mathbf{f}\left(\mathbf{x}\right)\equiv\mathbf{g}\left(\mathbf{x}\right)\equiv1$
and $\mathbf{K}=1$. An approach that can be applied to deterministic
systems and for stochastic systems with additive white noise is to
set $\mathbf{g}\left(\mathbf{x}\right)=\mathbf{x}$, and then choose
$\mathbf{f}$ as the transformation with smallest modeling error \cite{brunton2016discovering,brunton2016koopman}. 

A more general approach is to optimize the dominant spectrum of the
Koopman operator. At long timescales, the dynamics of the system are
usually dominated by the Koopman eigenfunctions of the Koopman operator
with large eigenvalues. If the dynamics obey detailed balance, those
eigenvalues are real-valued, and the variational approach for reversible
Markov processes can be applied that has made great progress in the
field of molecular dynamics \cite{NoeNueske_MMS13_VariationalApproach,NueskeEtAl_JCTC14_Variational}.
In such processes, the smallest modeling error of \eqref{eq:linear_model}
is achieved by setting $\mathbf{f}=\mathbf{g}$ equal to the corresponding
eigenfunctions. Ref. \cite{NoeNueske_MMS13_VariationalApproach} describes
a general approach to approximate the unknown eigenfunction from time
series data of a reversible Markov process: Given a set of orthogonal
candidate functions, $\mathbf{f}$, it can be shown that their time-autocorrelations
are lower bounds to the corresponding Koopman eigenvalues, and are
equal to them exactly if, and only if $\mathbf{f}$ are equal to the
Koopman eigenfunctions. This approach provides a variational score,
such as the sum of estimated eigenvalues (the Rayleigh trace), that
can be optimized to approximate the eigenfunctions. If $\mathbf{f}$
is defined by a linear superposition of a given set of basis functions,
then the optimal coefficients are found equivalently by either maximizing
the variational score, or minimizing the regression error in the feature
space as done in EDMD \cite{WilliamsKevrekidisRowley_JNS15_EDMD}
\textendash{} see \cite{WuEtAl_JCP17_VariationalKoopman}. However,
the regression error cannot be used to select the form and the number
of basis functions themselves, whereas the variational score can.
When working with a finite dataset, however, it is important to avoid
overfitting, and to this end a cross-validation method has been proposed
to compute variational scores that take the statistical error into
account \cite{McGibbonPande_JCP15_CrossValidation}. Such cross-validated
variational scores can be used to determine the size and type of the
function classes and the other hyper-parameters of the dynamical model.

While this approach is extremely powerful for stationary and data
and reversible Markov processes, almost all real-world dynamical processes
and time-series thereof are irreversible and often even non-stationary.
In this paper, we introduce a variational approach for Markov processes
(VAMP) that can be employed to optimize parameters and hyper-parameters
of arbitrary Markov processes. VAMP is based on the singular value
decomposition of the Koopman operator, which overcomes the limited
usefulness of the eigenvalue decomposition of time-irreversible and
non-stationary processes. We first show that the approximation error
of the Koopman operator deduced from the linear model \eqref{eq:linear_model}
can be minimized by setting $\mathbf{f}$ and $\mathbf{g}$ to be
the top left and right singular functions of the Koopman operator.
Then, by using the variational description of singular components,
a class of variational scores, VAMP-$r$ for $r=1,2,\ldots$, are
proposed to measure the similarity between the estimated singular
functions and the true ones. Maximization of any of these variational
scores leads to optimal model parameters and is algorithmically identical
to Canonical Correlation Analysis (CCA) between the featurized time-lagged
pair of variables $\mathbf{x}_{t}$ and $\mathbf{x}_{t+\tau}$. This
approach can also be employed to learn the feature transformations
by nonlinear function approximators, such as deep neural networks.
Furthermore, we establish a relationship between the VAMP-2 score
and the approximation error of the dynamical model with respect to
the true Koopman operator. We show that this approximation error can
be practically computed up to a constant, and define its negative
as the VAMP-E score. Finally, we demonstrate that optimizing the VAMP-E
score in a cross-validation framework leads to an optimal choice of
hyperparameters. 

\section{Theory\label{sec:Theory}}

\subsection{Koopman analysis of dynamical systems and its singular value decomposition}

The Koopman operator $\mathcal{K}_{\tau}$ of a Markov process is
a linear operator defined by
\begin{equation}
\mathcal{K}_{\tau}g(\mathbf{x})\triangleq\mathbb{E}\left[g(\mathbf{x}_{t+\tau})\mid\mathbf{x}_{t}=\mathbf{x}\right].\label{eq:koopman}
\end{equation}
For given $\mathbf{x}_{t}$, the Koopman operator can be used to compute
the conditional expected value of an arbitrary observable $g$ at
time $t+\tau$. For the special choice that $g$ is the Dirac delta
function $\delta_{\mathbf{y}}$ centered at $\mathbf{y}$, application
of the Koopman operator evaluates the transition density of the dynamics,
$\mathcal{K}_{\tau}\delta_{\mathbf{y}}(\mathbf{x})=\mathbb{P}(\mathbf{x}_{t+\tau}=\mathbf{y}|\mathbf{x}_{t}=\mathbf{x})$
(see Appendix \ref{subsec:ptau-and-Koopman}). Thus, the Koopman operator
is a complete description of the dynamical properties of a Markovian
system. For convenience of analysis, we consider here $\mathcal{K}_{\tau}$
as a mapping from $\mathcal{L}_{\rho_{1}}^{2}=\left\{ g|\left\langle g,g\right\rangle _{\rho_{1}}<\infty\right\} $
to $\mathcal{L}_{\rho_{0}}^{2}=\left\{ f|\left\langle f,f\right\rangle _{\rho_{0}}<\infty\right\} $,
where $\rho_{0}$ and $\rho_{1}$ are empirical distributions of $\mathbf{x}_{t}$
and $\mathbf{x}_{t+\tau}$ of all transition pairs $\{(\mathbf{x}_{t},\mathbf{x}_{t+\tau})\}$
occurring in the given time series (see Appendix \ref{sec:empirical-distributions}),
and the inner products are defined by
\begin{equation}
\left\langle f,g\right\rangle _{\rho_{0}}=\int f\left(\mathbf{x}\right)g\left(\mathbf{x}\right)\rho_{0}\left(\mathbf{x}\right)\mathrm{d}\mathbf{x},\quad\left\langle f,g\right\rangle _{\rho_{1}}=\int f\left(\mathbf{x}\right)g\left(\mathbf{x}\right)\rho_{1}\left(\mathbf{x}\right)\mathrm{d}\mathbf{x}.
\end{equation}

How is the finite-dimensional linear model \eqref{eq:linear_model}
related to the Koopman operator description? Let us consider $\mathbf{f}(\mathbf{x}_{t})$
to be a sufficient statistics for $\mathbf{x}_{t}$, and let $\mathbf{g}$
be a dictionary of observables, then the value of an arbitrary observable
$h$ in the subspace of $\mathbf{g}$, i.e. $h=\mathbf{c}^{\top}\mathbf{g}$,
with some coefficients $\mathbf{c}$, can be predicted from $\mathbf{x}_{t}$
as $\mathbb{E}\left[h(\mathbf{x}_{t+\tau})|\mathbf{x}_{t}\right]=\mathbf{c}^{\top}\mathbf{K}^{\top}\mathbf{f}(\mathbf{x}_{t})$.
This implies that Eq. \eqref{eq:linear_model} is an algebraic representation
of the projection of the Koopman operator onto the subspace spanned
by functions $\mathbf{f}$ and $\mathbf{g}$, and the matrix $\mathbf{K}$
is therefore called the Koopman matrix. Combining this insight with
the generalized Eckart-Young Theorem \cite{hsing2015theoretical}
leads to our first result, namely what is the optimal choice of functions
$\mathbf{f}$ and $\mathbf{g}$:\emph{ }
\begin{theorem}
\textbf{\emph{\label{thm:Optimal-approximation}Optimal approximation
of Koopman operator}}. If $\mathcal{K}_{\tau}$ is a Hilbert-Schmidt
operator between the separable Hilbert spaces $\mathcal{L}_{\rho_{1}}^{2}$
and $\mathcal{L}_{\rho_{0}}^{2}$, the linear model \eqref{eq:linear_model}
with the smallest modeling error in Hilbert-Schmidt norm is given
by $\mathbf{f}=(\psi_{1},\ldots,\psi_{k})^{\top}$, $\mathbf{g}=(\phi_{1},\ldots,\phi_{k})^{\top}$
and $\mathbf{K}=\mathrm{diag}(\sigma_{1},\ldots,\sigma_{k})$, i.e.,
\begin{equation}
\mathbb{E}\left[\phi_{i}(\mathbf{x}_{t+\tau})\right]=\sigma_{i}\mathbb{E}\left[\psi_{i}(\mathbf{x}_{t})\right],\quad\text{for }i=1,\ldots,k\label{eq:optimal-linear-model}
\end{equation}
under the constraint $\mathrm{dim}(\mathbf{f}),\mathrm{dim}(\mathbf{g})\le k$,
and the projected Koopman operator deduced from \eqref{eq:optimal-linear-model}
is
\begin{equation}
\hat{\mathcal{K}}_{\tau}g=\sum_{i=1}^{k}\sigma_{i}\left\langle g,\phi_{i}\right\rangle _{\rho_{1}}\psi_{i},\label{eq:svd-truncated}
\end{equation}
where the singular value $\sigma_{i}>0$ is the square root of the
$i$th largest eigenvalue of $\mathcal{K}_{\tau}^{*}\mathcal{K}_{\tau}$
or $\mathcal{K}_{\tau}\mathcal{K}_{\tau}^{*}$, the left and right
singular function $\psi_{i},\phi_{i}$ are the $i$th eigenfunctions
of $\mathcal{K}_{\tau}^{*}\mathcal{K}_{\tau}$ and $\mathcal{K}_{\tau}\mathcal{K}_{\tau}^{*}$
with
\begin{equation}
\left\langle \psi_{i},\psi_{j}\right\rangle _{\rho_{0}}=1_{i=j},\quad\left\langle \phi_{i},\phi_{j}\right\rangle _{\rho_{1}}=1_{i=j},
\end{equation}
and the first singular component is always given by $(\sigma_{1},\phi_{1},\psi_{1})=(1,\mathbbm1,\mathbbm1)$
with $\mathbbm1\left(\mathbf{x}\right)\equiv1$.
\end{theorem}
\begin{proof}
See Appendix \ref{subsec:Singular-value-decomposition}.
\end{proof}

This theorem is universal for Markov processes, and the major assumption
is that the Koopman operator is Hilbert-Schmidt, which is required
for the existences of the singular value decomposition (SVD) of $\mathcal{K}_{\tau}$
and the finite Hilbert-Schmidt norm $\Vert\mathcal{K}_{\tau}\Vert_{\mathrm{HS}}$.
Appendix \ref{subsec:Sufficient-conditions} provides two sufficient
conditions for the assumption. However, it is worth noting that the
Koopman operators of deterministic systems are not Hilbert-Schmidt
or even compact in usual cases (see Appendix \ref{subsec:deterministic-Koopman-operators}),
thus all conclusions and methods in this paper are not applicable
to deterministic systems.

\begin{figure}
\begin{centering}
\includegraphics[width=1\textwidth]{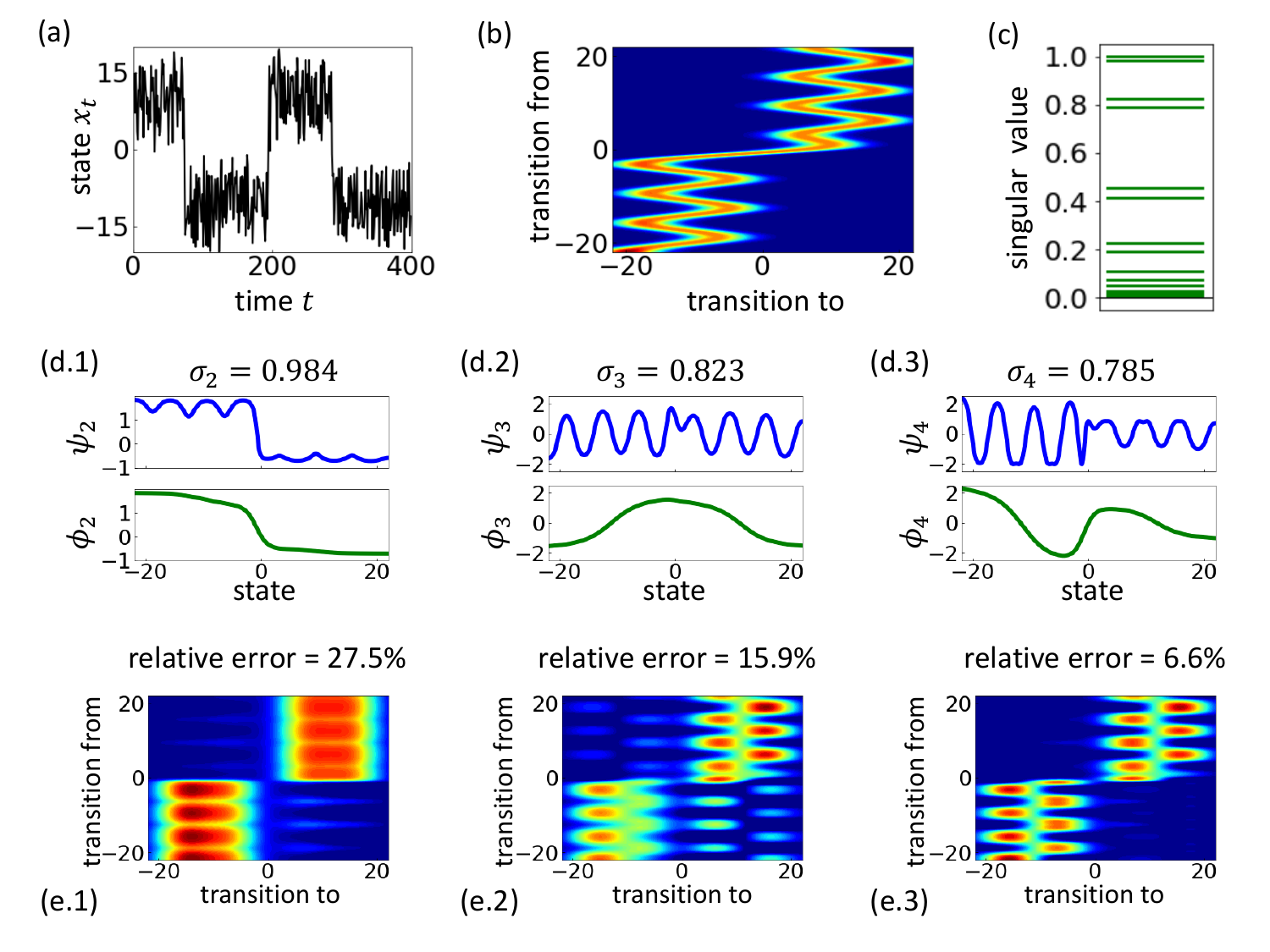}
\par\end{centering}
\caption{Analysis results of the dynamical system \eqref{eq:one-dimensional-example}
with lag time $\tau=1$. (a) A typical simulation trajectory. (b)
Transition density $\mathbb{P}(x_{t+1}|x_{t})$. (c) The singular
values. (d) The first three nontrivial left and right singular functions.
(The first singular component is $(\sigma_{1},\phi_{1},\psi_{1})=(1,\mathbbm1,\mathbbm1)$.)
(e) Approximate transition densities obtained from the projected Koopman
operator $\hat{\mathcal{K}}_{\tau}$ consisting of first $k$ singular
components defined by \eqref{eq:svd-truncated} for $k=2,3,4$, where
the relative error is calculated as $\Vert\hat{\mathcal{K}}_{\tau}-\mathcal{K}_{\tau}\Vert_{\mathrm{HS}}/\Vert\mathcal{K}_{\tau}\Vert_{\mathrm{HS}}$.\label{fig:one-dimensional-svd}}
\end{figure}

\begin{figure}
\begin{centering}
\includegraphics[width=1\textwidth]{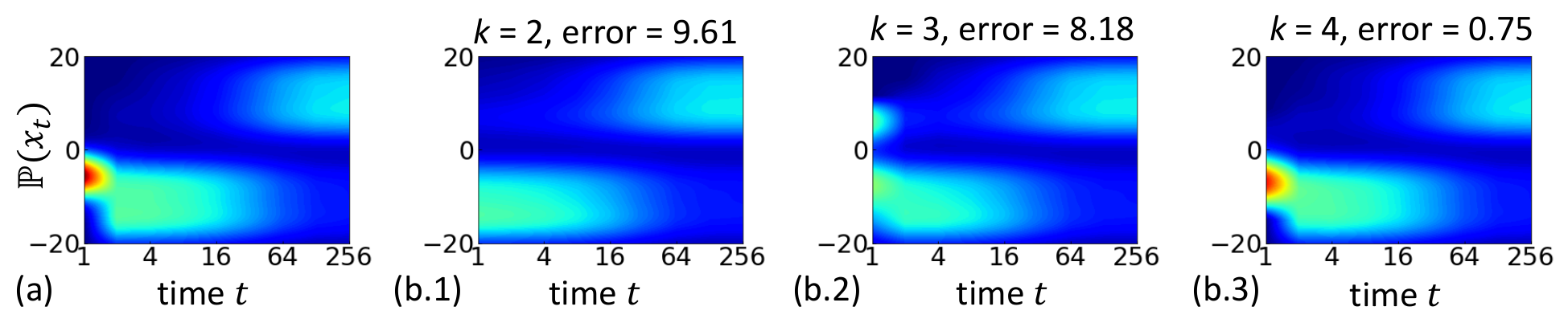}
\par\end{centering}
\caption{Probability density of state $x_{t}$ predicted by (a) the full model
and (b) the projected Koopman operator $\hat{\mathcal{K}}_{\tau}$
with rank $k=2,3,4$, where the initial state is $x_{0}=-12$.\label{fig:one-dimensional-evolution}}
\end{figure}

In addition, we prove in Appendix \ref{subsec:ptau-and-Koopman} that
$\Vert\hat{\mathcal{K}}_{\tau}-\mathcal{K}_{\tau}\Vert_{\mathrm{HS}}$
is equal to a weighted $\mathcal{L}^{2}$ error of the transition
density, which provides a more meaningful interpretation of the modeling
error in Hilbert-Schmidt norm.
\begin{example}
\label{exa:one-dimensional-example-svd}Consider a one-dimensional
dynamical system
\begin{equation}
x_{t+1}=\frac{x_{t}}{2}+\frac{7x_{t}}{1+0.12x_{t}^{2}}+6\cos x_{t}+\sqrt{10}u_{t}\label{eq:one-dimensional-example}
\end{equation}
evolving in the state space $[-20,20]$, where $u_{t}$ is a standard
Gaussian white noise zero mean and unit variance (see Appendix \ref{subsec:app-One-dimensional-system}
for details on the numerical simulations and analysis). This system
has two metastable states with the boundary close to $x=0$ as shown
in Fig.~\ref{fig:one-dimensional-svd}a, and the singular components
are summarized in Figs.~\ref{fig:one-dimensional-svd}c and \ref{fig:one-dimensional-svd}d.
As shown in the figures, the sign structures of the second left and
right singular functions clearly indicate the metastable states, and
the third and forth singular functions provide more detailed information
on the dynamics. An accurate estimate of the transition density can
be obtained by combining the first four singular components and the
corresponding relative approximation error of the Koopman operator
is only $6.6\%$ (see Figs.~\ref{fig:one-dimensional-svd}b and \ref{fig:one-dimensional-svd}e).
In addition, we utilize the finite-rank approximate Koopman operators
to predict the time evolution of the distribution of $x_{t}$ for
$t=1,\ldots,256$ with the initial state $x_{0}=12$, and a small
error can also be achieved when the rank is only $4$ as displayed
in Fig.~\ref{fig:one-dimensional-evolution}, where
\begin{equation}
\mathrm{error}=\sum_{t=1}^{256}\int\rho_{1}(x_{t})^{-1}\left(\hat{\mathbb{P}}(x_{t}|x_{0})-\mathbb{P}(x_{t}|x_{0})\right)^{2}\mathrm{d}x_{t}
\end{equation}
is the cumulative kinetic distance \cite{NoeClementi_JCTC15_KineticMap}
between the transition density $\mathbb{P}(x_{t}|x_{0})$ and its
estimate $\hat{\mathbb{P}}(x_{t}|x_{0})$, and $\rho_{1}$ is the
stationary distribution.
\end{example}
There are other formalisms to describe Markovian dynamics, for example,
the Markov propagator or the weighted Markov propagator, also called
transfer operator \cite{SchuetteFischerHuisingaDeuflhard_JCompPhys151_146}.
These propagators are commonly used for modeling physical processes
such as molecular dynamics, and describe the evolution of probability
densities instead of observables. We show in Appendix \ref{sec:svd-P}
that all conclusions in this paper can be equivalently established
by interpreting $(\sigma_{i},\rho_{1}\phi_{i},\rho_{0}\psi_{i})$
as the singular components of the Markov propagator.

\subsection{Variational principle for Markov processes\label{subsec:Variational-principle}}

In order to allow the optimal model \eqref{eq:optimal-linear-model}
to be estimated from data, we develop a variational principle for
the approximation of singular values and singular functions of Markov
processes.

According to the Rayleigh variational principle of singular values,
the first singular component maximizes the generalized Rayleigh quotient
of $\mathcal{K}_{\tau}$ as
\begin{equation}
(\psi_{1},\phi_{1})=\arg\max_{f,g}\frac{\left\langle f,\mathcal{K}_{\tau}g\right\rangle _{\rho_{0}}}{\sqrt{\left\langle f,f\right\rangle _{\rho_{0}}\cdot\left\langle g,g\right\rangle _{\rho_{1}}}}
\end{equation}
and the maximal value of the generalized Rayleigh quotient is equal
to the first singular value $\sigma_{1}=\left\langle \psi_{1},\,\mathcal{K}_{\tau}\phi_{1}\right\rangle _{\rho_{0}}$.
For the $i$th singular component with $i>1$, we have
\begin{equation}
(\psi_{i},\phi_{i})=\arg\max_{f,g}\frac{\left\langle f,\mathcal{K}_{\tau}g\right\rangle _{\rho_{0}}}{\sqrt{\left\langle f,f\right\rangle _{\rho_{0}}\cdot\left\langle g,g\right\rangle _{\rho_{1}}}}
\end{equation}
under constraints
\begin{equation}
\left\langle f,\psi_{j}\right\rangle _{\rho_{0}}=\left\langle g,\phi_{j}\right\rangle _{\rho_{1}}=0,\quad\forall j=1,\ldots,i-1
\end{equation}
and the maximal value is equal to $\sigma_{i}=\left\langle \psi_{i},\,\mathcal{K}_{\tau}\phi_{i}\right\rangle _{\rho_{0}}$.
These insights can be summarized by the following variational theorem
for seeking all top $k$ singular components simultaneously:
\begin{theorem}
\textbf{\emph{\label{thm:VAMP-variational-principle}VAMP variational
principle}}\emph{. }The $k$ dominant singular components of a Koopman
operator are the solution of the following maximization problem:
\begin{align}
\sum_{i=1}^{k}\sigma_{i}^{r} & =\max_{\mathbf{f},\mathbf{g}}\mathcal{R}_{r}\left[\mathbf{f},\mathbf{g}\right],\nonumber \\
s.t. & \left\langle f_{i},f_{j}\right\rangle _{\rho_{0}}=1_{i=j},\nonumber \\
 & \left\langle g_{i},g_{j}\right\rangle _{\rho_{1}}=1_{i=j},\label{eq:variational}
\end{align}
where $r\ge1$ can be any positive integer. The maximal value is achieved
by the singular functions $f_{i}=\psi_{i}$ and $g_{i}=\phi_{i}$
and
\begin{equation}
\mathcal{R}_{r}\left[\mathbf{f},\mathbf{g}\right]=\sum_{i=1}^{k}\left\langle f_{i},\mathcal{K}_{\tau}g_{i}\right\rangle _{\rho_{0}}^{r}\label{eq:Rr}
\end{equation}
is called the VAMP-r score of $\mathbf{f}$ and $\mathbf{g}$.
\end{theorem}
\begin{proof}
See Appendix \ref{sec:Proof-of-Proposition-variational-svd}.
\end{proof}

This theorem generalizes Proposition 2 in \cite{froyland2013analytic}
where only the case of $k=2$ is considered. It is important to note
that this theorem has direct implications for the data-driven estimation
of dynamical models. For $r=1$, $\mathcal{R}_{r}\left[\mathbf{f},\mathbf{g}\right]$
is actually the time-correlation between $\mathbf{f}(\mathbf{x}_{t})$
and $\mathbf{g}(\mathbf{x}_{t+\tau})$ since $\left\langle f_{i},\mathcal{K}_{\tau}g_{i}\right\rangle _{\rho_{0}}=\mathbb{E}_{t}[f_{i}(\mathbf{x}_{t})g_{i}(\mathbf{x}_{t+\tau})]$
and $\mathbb{E}_{t}[\cdot]$ denotes the expectation value over all
transition pairs $(x_{t},x_{t+\tau})$ in the time series. Hence the
maximization of VAMP-$r$ is analogous to the problem of seeking orthonormal
transformations of $\mathbf{x}_{t}$ and $\mathbf{x}_{t+\tau}$ with
maximal time-correlations, and we can thus utilize the canonical correlation
analysis (CCA) algorithm \cite{hardoon2004canonical} in order to
estimate the singular components from data.

\subsection{Comparison with related analysis approaches}

The SVD of the Koopman operator is equivalent to the eigenvalue decomposition
when the Markov process is time-reversible and stationary with $\rho_{0}=\rho_{1}$,
and therefore the variational principle presented here is a generalization
of that developed for reversible conformation dynamics \cite{NoeNueske_MMS13_VariationalApproach,NueskeEtAl_JCTC14_Variational}.
Specifically, VAMP-1 maximizes the \emph{Rayleigh trace}, i.e. the
sum of the estimated eigenvalues \cite{NoeNueske_MMS13_VariationalApproach,McGibbonPande_JCP15_CrossValidation},
and VAMP-2 maximizes the \emph{kinetic variance} introduced in \cite{NoeClementi_JCTC15_KineticMap}.
See Appendix \ref{sec:Variational-principle-reversible} for a detailed
derivation of the reversible variational principle from the VAMP variational
principle. For irreversible Markov processes, the singular functions
can provide low-dimensional embeddings of kinetic distances between
states like eigenfunctions of reversible processes \cite{Paul2018identification}.
Furthermore, the coherent sets of nonstationary Markov processes,
which are the generalization of metastable states, can be identified
from dominant singular functions \cite{koltai2018optimal}.

The dynamics of an irreversible Markov process can also be analyzed
through solving the eigenvalue problem $\mathcal{K}_{\tau}g=\lambda g$
(see, e.g., \cite{WilliamsKevrekidisRowley_JNS15_EDMD,williams2014kernel,KlusKoltaiSchuette_ApproximationKoopman,KlusSchuette_Arxiv15_Tensor}),
and the eigenfunctions form an invariant subspace of the Koopman operator
for multiple lag times since the eigenvalue problem satisfies
\begin{equation}
\mathcal{K}_{\tau}g=\lambda g\Rightarrow\mathcal{K}_{n\tau}g=\lambda^{n}g,\quad\forall n\ge1.
\end{equation}
However, as far as we know, there is no variational principle for
approximate eigenfunctions of irreversible Markov processes, and it
is difficult to evaluate errors of projections of Koopman operators
to the invariant subspaces. The SVD based analysis approach overcomes
the above problems and yields the optimal finite-rank approximate
models. The major limitation of this approach comes from the fact
that the singular functions are dependent on the choice of the lag
time and the optimality of model \eqref{eq:optimal-linear-model}
holds only for a fixed $\tau$. The optimization and error analysis
of Koopman models for multiple lag times will be studied in our future
work.

\section{Estimation algorithms}

We introduce algorithms to estimate optimal dynamical models from
time series data. We make the Ansatz to represent the feature functions
$\mathbf{f}$ and $\mathbf{g}$ as linear combinations of \emph{basis}
functions $\boldsymbol{\chi}_{0}=(\chi_{0,1},\chi_{0,2},\ldots)^{\top}$
and $\boldsymbol{\chi}_{1}=(\chi_{1,1},\chi_{1,2},\ldots)^{\top}$:
\begin{align}
\mathbf{f} & =\mathbf{U}^{\top}\boldsymbol{\chi}_{0},\nonumber \\
\mathbf{g} & =\mathbf{V}^{\top}\boldsymbol{\chi}_{1}.\label{eq:linear_Ansatz}
\end{align}
Here, $\mathbf{U}$ and $\mathbf{V}$ are matrices of size $m\times k$
and $m^{\prime}\times k$, i.e. we are trying to approximate $k$
singular components by linearly combining $m$ and $m^{\prime}$ basis
functions. For the sake of generality we have assumed that $\mathbf{f}$
and $\mathbf{g}$ are represented by different basis sets. However,
in practice one can justify using a single basis set the joint set
$\boldsymbol{\chi}^{\top}=(\boldsymbol{\chi}_{0}^{\top},\boldsymbol{\chi}_{1}^{\top})$
as an Ansatz for both $\mathbf{f}$ and $\mathbf{g}$. Please note
that despite the linear Ansatz \eqref{eq:linear_Ansatz}, the feature
functions may be strongly nonlinear in the system's state variables
$\mathbf{x}$, thus we are not restricting the generality of the functions
$\mathbf{f}$ and $\mathbf{g}$ that can be represented. In this section,
we consider three problems: (i) optimizing $\mathbf{U}$ and $\mathbf{V}$,
(ii) optimizing $\boldsymbol{\chi}_{0}$ and $\boldsymbol{\chi}_{1}$
and (iii) assessing the quality of the resulting dynamical model.

For convenience of notation, we denote by $\mathbf{C}_{00},\mathbf{C}_{11},\mathbf{C}_{01}$
the covariance matrices and time-lagged covariance matrices of basis
functions, which can be computed from a trajectory $\{x_{1},\ldots,x_{T}\}$
by
\begin{eqnarray}
\mathbf{C}_{00} & \triangleq & \mathbb{E}_{t}\left[\boldsymbol{\chi}_{0}\left(\mathbf{x}_{t}\right)\boldsymbol{\chi}_{0}\left(\mathbf{x}_{t}\right)^{\top}\right]\approx\frac{1}{T-\tau}\sum_{t=1}^{T-\tau}\boldsymbol{\chi}_{0}\left(\mathbf{x}_{t}\right)\boldsymbol{\chi}_{0}\left(\mathbf{x}_{t}\right)^{\top},\label{eq:C00}\\
\mathbf{C}_{11} & \triangleq & \mathbb{E}_{t}\left[\boldsymbol{\chi}_{1}\left(\mathbf{x}_{t+\tau}\right)\boldsymbol{\chi}_{1}\left(\mathbf{x}_{t+\tau}\right)^{\top}\right]\approx\frac{1}{T-\tau}\sum_{t=1+\tau}^{T}\boldsymbol{\chi}_{1}\left(\mathbf{x}_{t}\right)\boldsymbol{\chi}_{1}\left(\mathbf{x}_{t}\right)^{\top},\label{eq:C11}\\
\mathbf{C}_{01} & \triangleq & \mathbb{E}_{t}\left[\boldsymbol{\chi}_{0}\left(\mathbf{x}_{t}\right)\boldsymbol{\chi}_{1}\left(\mathbf{x}_{t+\tau}\right)^{\top}\right]\approx\frac{1}{T-\tau}\sum_{t=1}^{T-\tau}\boldsymbol{\chi}_{0}\left(\mathbf{x}_{t}\right)\boldsymbol{\chi}_{1}\left(\mathbf{x}_{t+\tau}\right)^{\top}.\label{eq:C01}
\end{eqnarray}
If there are multiple trajectories, the covariance matrices can be
computed in the same manner by averaging over all trajectories. Instead
of the direct estimators (\ref{eq:C00}-\ref{eq:C01}), more elaborated
estimation methods such as regularization methods \cite{tibshirani1996regression}
and reweighting estimators \cite{WuEtAl_JCP17_VariationalKoopman}
may be used.

\subsection{Feature TCCA: finding the best linear model in a given feature space\label{subsec:feature-TCCA}}

We first propose a solution for the problem of finding the optimal
parameter matrices $\mathbf{U}$ and $\mathbf{V}$ given that the
basis functions $\boldsymbol{\chi}_{0}$ and $\boldsymbol{\chi}_{1}$
are known. Substituting the linear Ansatz \eqref{eq:linear_Ansatz}
into the VAMP variational principle, shows that $\mathbf{U}$ and
$\mathbf{V}$ can be computed as the solutions of the maximization
problem:
\begin{align}
\max_{\mathbf{U},\mathbf{V}} & \mathcal{R}_{r}(\mathbf{U},\mathbf{V})\nonumber \\
\mathrm{s.t.} & \mathbf{U}^{\top}\mathbf{C}_{00}\mathbf{U}=\mathbf{I}\nonumber \\
 & \mathbf{V}^{\top}\mathbf{C}_{11}\mathbf{V}=\mathbf{I},\label{eq:vamp-matrix}
\end{align}
where
\begin{equation}
\mathcal{R}_{r}(\mathbf{U},\mathbf{V})=\sum_{i=1}^{k}\left(\mathbf{u}_{i}^{\top}\mathbf{C}_{01}\mathbf{v}_{i}\right)^{r}
\end{equation}
is a matrix representation of VAMP-$r$ score, and $\mathbf{u}_{i}$
and $\mathbf{v}_{i}$ are the $i$th columns of $\mathbf{U}$ and
$\mathbf{V}$. This problem can be solved by applying linear CCA \cite{hardoon2004canonical}
in the feature spaces defined by the basis sets $\boldsymbol{\chi}_{0}(\mathbf{x}_{t})$
and $\boldsymbol{\chi}_{1}(\mathbf{x}_{t+\tau})$, and the same solution
will be obtained for any other choice of $r$. (See Appendices \ref{sec:Correctness-of-feature-TCCA}
and \ref{subsec:Feature-TCCA-proj} for more detailed proof and analysis.)
The resulting algorithm for finding the best linear model is a CCA
in feature space, applied on time-lagged data. Hence we briefly call
this algorithm feature TCCA:
\begin{enumerate}
\item Compute covariance matrices $\mathbf{C}_{00},\mathbf{C}_{01},\mathbf{C}_{11}$
via (\ref{eq:C00}-\ref{eq:C01}).
\item Perform the truncated SVD
\[
\bar{\mathbf{K}}=\mathbf{C}_{00}^{-\frac{1}{2}}\mathbf{C}_{01}\mathbf{C}_{11}^{-\frac{1}{2}}\approx\mathbf{U}^{\prime}\mathbf{K}\mathbf{V}^{\prime\top},
\]
where $\bar{\mathbf{K}}$ is the Koopman matrix for the normalized
basis functions $\mathbf{C}_{00}^{-\frac{1}{2}}\boldsymbol{\chi}_{0}$
and $\mathbf{C}_{11}^{-\frac{1}{2}}\boldsymbol{\chi}_{1}$, $\mathbf{K}=\mathrm{diag}(K_{11},\ldots,K_{kk})$
is a diagonal matrix of the first $k$ singular values that approximate
the true singular values $\sigma_{1},...,\sigma_{k}$, and $\mathbf{U}^{\prime}$
and $\mathbf{V}^{\prime}$ consist of the $k$ corresponding left
and right singular vectors respectively.
\item Compute $\mathbf{U}=\mathbf{C}_{00}^{-\frac{1}{2}}\mathbf{U}^{\prime}$
and $\mathbf{V}=\mathbf{C}_{11}^{-\frac{1}{2}}\mathbf{V}^{\prime}$.
\item Output the linear model \eqref{eq:linear_model} with $K_{ii}$, $f_{i}=\mathbf{u}_{i}^{\top}\boldsymbol{\chi}_{0}$
and $g_{i}=\mathbf{v}_{i}^{\top}\boldsymbol{\chi}_{1}$ being the
estimates of the $i$th singular value, left singular function and
right singular function of the Koopman operator.
\end{enumerate}
Please note that this pseudocode is given only for illustrative purposes
and cannot be executed literally if $\mathbf{C}_{00}$ and $\mathbf{C}_{11}$
do not have full rank, i.e. are not invertible. To handle this problem,
we ensure that the basis functions are linearly independent by applying
a de-correlation (whitening) transformation that ensures that $\mathbf{C}_{00}$
and $\mathbf{C}_{11}$ will both have full rank. We then add the constant
function $\mathbbm1\left(x\right)\equiv1$ to the decorrelated basis
sets to ensure that $\mathbbm1$ belongs to the subspaces spanned
by $\boldsymbol{\chi}_{0}$ and by $\boldsymbol{\chi}_{1}$. It can
be shown that the singular values given by the feature TCCA algorithm
with these numerical modifications are bounded by $1$, and the first
estimated singular component is exactly $(K_{11},f_{1},g_{1})=(1,\mathbbm1,\mathbbm1)$
even in the presence of statistical noise and modeling error \textendash{}
see Appendix \ref{sec:Proof-of-Proposition-linear-variational-bound}
for details.

In the case of $k=\mathrm{dim}(\boldsymbol{\chi}_{0})=\mathrm{dim}(\boldsymbol{\chi}_{1})$
and full rank $\mathbf{C}_{00},\mathbf{C}_{11}$, the output of the
feature TCCA can be equivalently written as
\begin{align}
 & \mathbb{E}\left[\mathbf{V}^{\top}\boldsymbol{\chi}_{1}\left(\mathbf{x}_{t+\tau}\right)\right]=\mathbf{K}^{\top}\mathbb{E}\left[\mathbf{U}^{\top}\boldsymbol{\chi}_{0}\left(\mathbf{x}_{t}\right)\right]\nonumber \\
\Rightarrow & \mathbb{E}\left[\boldsymbol{\chi}_{1}\left(\mathbf{x}_{t+\tau}\right)\right]=\mathbf{K}_{\chi}^{\top}\mathbb{E}\left[\boldsymbol{\chi}_{0}\left(\mathbf{x}_{t}\right)\right]\label{eq:chi-K-chi}
\end{align}
where
\begin{eqnarray}
\mathbf{K}_{\chi} & = & \mathbf{U}\mathbf{K}\mathbf{V}^{-1}\nonumber \\
 & = & \mathbf{C}_{00}^{-1}\mathbf{C}_{01}\label{eq:K}
\end{eqnarray}
is equal to the least square solution to the regression problem $\boldsymbol{\chi}_{1}\left(\mathbf{x}_{t+\tau}\right)\approx\mathbf{K}_{\chi}^{\top}\boldsymbol{\chi}_{0}\left(\mathbf{x}_{t}\right)$.
Note that if we further assume that $\boldsymbol{\chi}_{0}=\boldsymbol{\chi}_{1}$,
\eqref{eq:chi-K-chi} is identical to the linear model of EDMD. Thus,
the feature TCCA can be seen as a generalization of EDMD that can
provide approximate Markov models for different basis $\boldsymbol{\chi}_{0}$
and $\boldsymbol{\chi}_{1}$. More discussion on the relationship
between the two methods is provided in Appendix \ref{sec:Relationship-between-VAMP-EDMD}.

\subsection{Nonlinear TCCA: optimizing the basis functions}

We now extend feature TCCA to a more flexible representation of the
transformation functions $\mathbf{f}$ and $\mathbf{g}$ by optimizing
the basis functions themselves:
\begin{align}
\mathbf{f}\left(\mathbf{x}\right) & =\mathbf{U}^{\top}\boldsymbol{\chi}_{0}\left(\mathbf{x};\mathbf{w}\right),\nonumber \\
\mathbf{g}\left(\mathbf{x}\right) & =\mathbf{V}^{\top}\boldsymbol{\chi}_{1}\left(\mathbf{x};\mathbf{w}\right).\label{eq:nonlinear_Ansatz}
\end{align}
Here, $\mathbf{w}$ represents a set of parameters that determines
the form of the basis functions. As a simple example, consider $\mathbf{w}$
to represent the mean vectors and covariance matrices of a Gaussian
basis set. However, $\boldsymbol{\chi}_{0}\left(\mathbf{x};\mathbf{w}\right)$
and $\boldsymbol{\chi}_{1}\left(\mathbf{x};\mathbf{w}\right)$ can
also represent very complex and nonlinear learning structures, such
as neural networks and decision trees.

\begin{figure}
\begin{centering}
\includegraphics[width=1\textwidth]{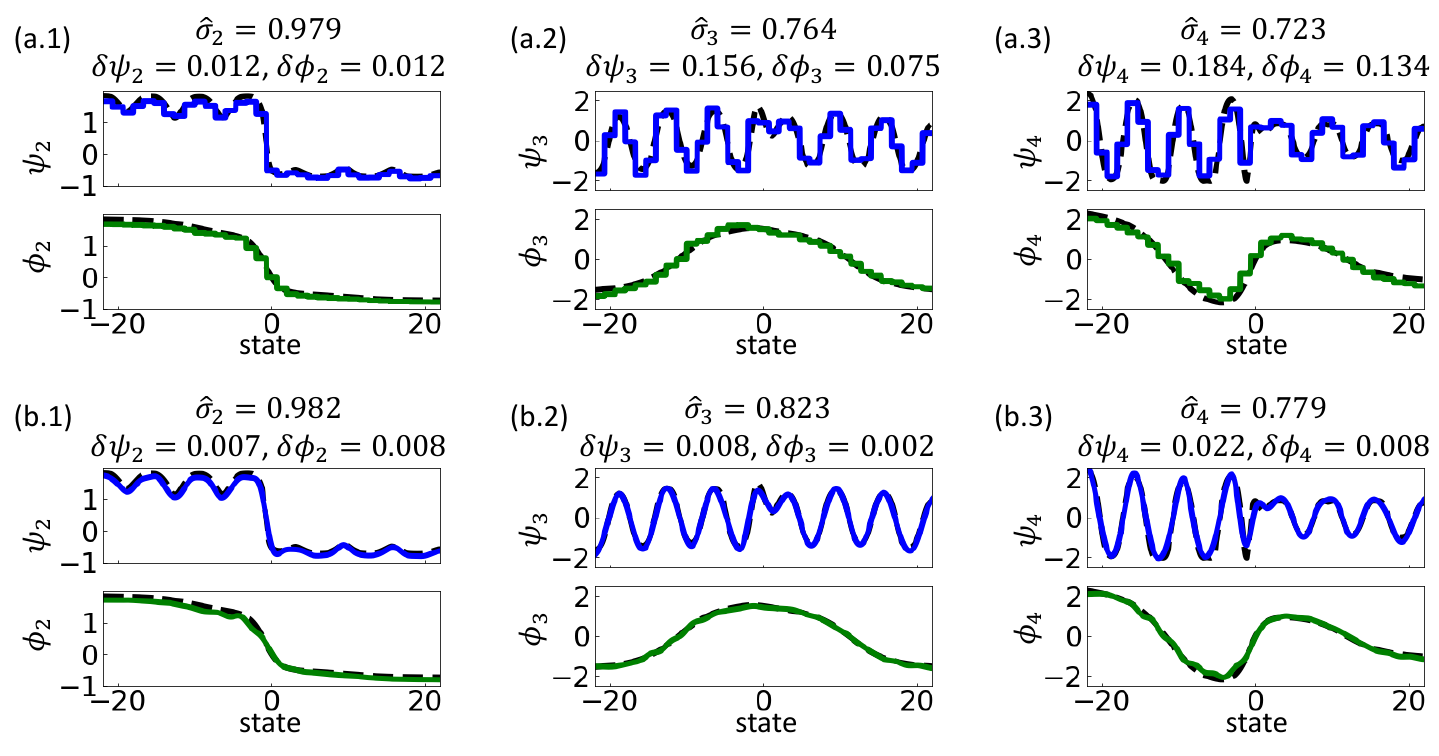}
\par\end{centering}
\caption{Estimated singular components of the system in Example \ref{exa:one-dimensional-example-svd},
where dash lines represent true singular functions, and the estimation
errors of singular functions are defined as $\delta\psi_{i}=\int\left(f_{i}(x)-\psi_{i}(x)\right)^{2}\rho_{0}(x)\mathrm{d}x$,
$\delta\phi_{i}=\int\left(g_{i}(x)-\phi_{i}(x)\right)^{2}\rho_{1}(x)\mathrm{d}x$
with $\rho_{0}=\rho_{1}$ being the stationary distribution. (a) Estimates
provided by feature TCCA with basis functions \eqref{eq:one-dimensional-example-indicator-basis}.
(b) Estimates provided by nonlinear TCCA with basis functions \eqref{eq:one-dimensional-example-rbf}.\label{fig:one-dimensional-estimation}}
\end{figure}

The parameters $\mathbf{w}$ could conceptually be determined together
with the linear expansion coefficients $\mathbf{U},\mathbf{V}$ by
solving \eqref{eq:vamp-matrix} with $\mathbf{C}_{00}$, $\mathbf{C}_{11}$,
$\mathbf{C}_{01}$ treated as functions of $\mathbf{w}$, but this
method is not practical due to the nonlinear equality constraints
are involved. In practice, we can set $k$ to be $\min\{\mathrm{dim}\left(\boldsymbol{\chi}_{0}\right),\mathrm{dim}\left(\boldsymbol{\chi}_{1}\right)\}$,
i.e., the largest number of singular components that can be approximated
given the basis set. Then the maximal VAMP-$r$ score for a fixed
$\mathbf{w}$ can be represented as
\begin{equation}
\max_{\mathbf{U},\mathbf{V}}\mathcal{R}_{r}=\left\Vert \mathbf{C}_{00}\left(\mathbf{w}\right)^{-\frac{1}{2}}\mathbf{C}_{01}\left(\mathbf{w}\right)\mathbf{C}_{11}\left(\mathbf{w}\right)^{-\frac{1}{2}}\right\Vert _{r}^{r},\label{eq:Rr-w}
\end{equation}
which can also be interpreted as the sum over the $r$'th power of
all singular values of the projected Koopman operator on subspaces
of $\boldsymbol{\chi}_{0},\boldsymbol{\chi}_{1}$ (see Eq.~\eqref{eq:max-Rr-matrix-norm}
in Appendix \ref{sec:Correctness-of-feature-TCCA} and Eq.~\eqref{eq:sum-sir-matrix-norm}
in Appendix \ref{subsec:Feature-TCCA-proj}). Here $\left\Vert \mathbf{A}\right\Vert _{r}$
denotes the $r$-Schatten norm of matrix $\mathbf{A}$, which is the
$\ell^{r}$ norm of singular values of $\mathbf{A}$, and $\left\Vert \mathbf{A}\right\Vert _{2}$
equals the Frobenius norm of $\mathbf{A}$. The parameters $\mathbf{w}$
can be optimized without computing $\mathbf{U}$ and $\mathbf{V}$
explicitly. Using these ideas, nonlinear TCCA can be performed as
follows:
\begin{enumerate}
\item Compute $\mathbf{w}^{*}=\arg\max_{\mathbf{w}}\left\Vert \mathbf{C}_{00}\left(\mathbf{w}\right)^{-\frac{1}{2}}\mathbf{C}_{01}\left(\mathbf{w}\right)\mathbf{C}_{11}\left(\mathbf{w}\right)^{-\frac{1}{2}}\right\Vert _{r}^{r}$
by gradient descent or other nonlinear optimization methods.
\item Approximate the Koopman singular values and singular functions using
the feature TCCA algorithm with basis sets $\boldsymbol{\chi}_{0}\left(\mathbf{x};\mathbf{w}^{*}\right)$
and $\boldsymbol{\chi}_{1}\left(\mathbf{x};\mathbf{w}^{*}\right)$.
\end{enumerate}
Unlike the estimated singular components generated by the feature
TCCA, the estimation results of the nonlinear TCCA do generally depend
on the value of $r$. (An example is given in Appendix \ref{subsec:example-nonlinear-TCCA},
where the VAMP scores can be analytically computed.) We suggest to
set $r=2$ in applications for the direct relationship between the
VAMP-2 score and the approximation error of Koopman operators and
the convenience of cross-validation (see below). The details of the
nonlinear TCCA, including the optimization algorithm and regularization,
are beyond the scope of this paper. Appendix \ref{subsec:Parameter-optimization-nonlinear-TCCA}
provides a brief description of the implementation, and related work
based on kernel methods and deep networks can be found in \cite{andrew2013deep,vampnet}.
\begin{example}
\label{exa:one-dimensional-example-estimation}Let us consider the
stochastic system described in Example \ref{exa:one-dimensional-example-svd}
again. We generate $10$ simulation trajectories of length $500$,
and approximate the dominant singular components by the feature TCCA.
Here, the basis functions are
\begin{equation}
\chi_{0,i}(x)=\chi_{1,i}(x)=1_{\frac{40\cdot(i-1)}{m}-20\le x\le\frac{40\cdot i}{m}-20},\text{ for }i=1,\ldots,m,\label{eq:one-dimensional-example-indicator-basis}
\end{equation}
which define a partition of the domain $[-20,20]$ into $m=33$ disjoint
intervals. In other words, the approximation is performed based on
an MSM with $33$ discrete states. Estimation results are given in
Fig.~\ref{fig:one-dimensional-estimation}a, where the discretization
errors arising from indicator basis functions are clearly shown. For
comparison, we also implement the nonlinear TCCA algorithm with radial
basis functions 
\begin{equation}
\chi_{0,i}(x;w)=\chi_{1,i}(x;w)=\frac{\exp\left(-w\left(x-c_{i}\right)^{2}\right)}{\sum_{j=1}^{m}\exp\left(-w\left(x-c_{j}\right)^{2}\right)}\label{eq:one-dimensional-example-rbf}
\end{equation}
with smoothing parameter $w\ge0$, where $c_{i}=\frac{40\cdot(i-0.5)}{m}-20$
for $i=1,\ldots,m$ are uniformly distributed in $[-20,20]$. Notice
that the basis functions given in \eqref{eq:one-dimensional-example-indicator-basis}
are a specific case of the radial basis functions with $w=\infty$,
and it is therefore possible to achieve better approximation by optimizing
$w$. As can be seen from Fig.~\ref{fig:one-dimensional-estimation}b,
the nonlinear TCCA provides more accurate estimates of singular functions
and singular values (see Appendix \ref{subsec:app-One-dimensional-system}
for more details). In addition, both feature TCCA and nonlinear TCCA
underestimate the dominant singular values as stated by the variational
principle.
\end{example}
The nonlinear TCCA is similar to the EDMD with dictionary learning
(EDMD-DL) \cite{li2017extended}, where the feature transformations
are optimized by minimizing the regression error of \eqref{eq:linear_model}.
The major advantages of the nonlinear TCCA over EDMD-DL are: First,
the uninformative model with zero regression error can be systematically
excluded without any extra constraints on features. Second, the optimization
objective is directly related to the approximation error of the Koopman
operator (see Section \ref{subsec:Error-analysis}). Some recent methods
extend EDMD-DL for modeling Koopman operators of deterministic systems
\cite{takeishi2017learning,lusch2018deep,otto2019linearly}, and solve
the first problem by using the prediction error between the observed
$x_{t}$ and that predicted by the low-dimensional model. But they
cannot be applied to stochastic Koopman operators of Markov processes
directly.

\subsection{Error analysis\label{subsec:Error-analysis}}

According to \eqref{eq:svd-truncated}, both feature TCCA and nonlinear
TCCA lead to a rank $k$ approximation
\begin{equation}
\hat{\mathcal{K}}_{\tau}g=\sum_{i=1}^{k}K_{ii}\left\langle g,g_{i}\right\rangle _{\rho_{1}}f_{i}=\sum_{i=1}^{k}K_{ii}\left\langle g,\mathbf{v}_{i}^{\top}\boldsymbol{\chi}_{1}\right\rangle _{\rho_{1}}\mathbf{u}_{i}^{\top}\boldsymbol{\chi}_{0}\label{eq:finite-rank-approximation}
\end{equation}
to $\mathcal{K}_{\tau}$. We consider here the approximation error
of \eqref{eq:finite-rank-approximation} in a general case where $\mathbf{f}=\mathbf{U}^{\top}\boldsymbol{\chi}_{0}$
and $\mathbf{g}=\mathbf{V}^{\top}\boldsymbol{\chi}_{1}$ may not satisfy
the orthonormal constraints due to statistical noise and numerical
errors. After a few steps of derivation, the approximation error can
be expressed as

\begin{equation}
\left\Vert \hat{\mathcal{K}}_{\tau}-\mathcal{K}_{\tau}\right\Vert _{\mathrm{HS}}^{2}=-\mathcal{R}_{E}[\mathbf{K},\mathbf{f},\mathbf{g}]+\left\Vert \mathcal{K}_{\tau}\right\Vert _{\mathrm{HS}}^{2}\label{eq:error}
\end{equation}
with
\begin{equation}
\mathcal{R}_{E}[\mathbf{K},\mathbf{f},\mathbf{g}]=2\sum_{i}K_{ii}\left\langle f_{i},\mathcal{K}_{\tau}g_{i}\right\rangle _{\rho_{0}}-\sum_{i,j}K_{ii}K_{jj}\left\langle f_{i},f_{j}\right\rangle _{\rho_{0}}\left\langle g_{i},g_{j}\right\rangle _{\rho_{1}}.
\end{equation}
Remarkably, this error decomposes into a unknown constant part (the
square of Hilbert-Schmidt norm of $\mathcal{K}_{\tau}$), and a model-dependent
part $\mathcal{R}_{E}$ that can be entirely estimated from data by
its matrix representation:
\begin{equation}
\mathcal{R}_{E}(\mathbf{K},\mathbf{U},\mathbf{V})=\mathrm{tr}\left[2\mathbf{K}\mathbf{U}^{\top}\mathbf{C}_{01}\mathbf{V}-\mathbf{K}\mathbf{U}^{\top}\mathbf{C}_{00}\mathbf{U}\mathbf{K}\mathbf{V}^{\top}\mathbf{C}_{11}\mathbf{V}\right].
\end{equation}
$\mathcal{R}_{E}$, is thus a score that can be used alternatively
to the VAMP-$r$ scores, and we call $\mathcal{R}_{E}$ VAMP-E score.
It can be proved that the maximization of $\mathcal{R}_{E}$ is equivalent
to maximization of $\mathcal{R}_{2}$ in feature TCCA or nonlinear
TCCA. However, these scores will behave differently in terms of hyper-parameter
optimization (see Sec. \ref{sec:Cross-validation}). Proofs and analysis
are given in Appendix \ref{sec:Analysis-VAMP-E}.

\section{Model validation}

\subsection{Cross-validation for hyper-parameter optimization\label{sec:Cross-validation}}

For a data-driven estimation of dynamical models, either using feature
TCCA or nonlinear TCCA, we have to strike a balance between the modeling
or discretization error and the statistical or overfitting error.
The choice of number and type of basis functions is critical for both.
If basis sets are very small and not flexible enough to capture singular
functions, the approximation results may be inaccurate with large
biases. We can improve the variational score and reduce the modeling
error by larger and more flexible basis sets. But too complicated
basis sets will produce unstable estimates with large statistical
variances, and in particular poor predictions on data that has not
been used in the estimation process \textendash{} this problem is
known as \emph{overfitting} in the machine learning community. A popular
way to achieve the balance between the statistical bias and variance
are resampling methods, including bootstrap and cross-validation \cite{friedman2001elements}.
They iteratively fit a model in a training set, which are sampled
from the data with or without replacement, and validate the model
in the complementary dataset. Alternatively, there are also Bayesian
hyper-parameter optimization methods. See \cite{ArlotCelisse_StatSurv10_CVReview,snoek2012practical}
for an overview. Here, we will focus on cross-validation and describe
how to use the VAMP scores in this and similar resampling frameworks.

Let $\boldsymbol{\theta}$ be hyper-parameters in feature TCCA or
nonlinear TCCA that need to be specified. For example, $\boldsymbol{\theta}$
includes the number and functional form of basis functions used in
feature TCCA, or the architecture and connectivity of a neural network
used for nonlinear TCCA. Generally speaking, different values of $\boldsymbol{\theta}$
correspond to different dynamical models that we want to rank, and
these models may be of completely different types. The cross-validation
of $\boldsymbol{\theta}$ can be performed as follows:
\begin{enumerate}
\item Separate the available trajectories into $J$ disjoint folds $\mathcal{D}_{1},\ldots,\mathcal{D}_{J}$
with approximately equal size. If there are only a small number of
long trajectories, we can divide each trajectory into blocks of length
$L$ with $\tau<L\ll T$ and create folds based on the blocks. This
defines a number of $J$ training sets, with training set $j$ consisting
of all data except the $j$th fold, $\mathcal{D}_{j}^{\mathrm{train}}=\cup_{l\neq j}\mathcal{D}_{l}$,
and the $j$th fold used as test set $\mathcal{D}_{j}^{\mathrm{test}}=\mathcal{D}_{j}$.
\item For each hyper-parameter set $\boldsymbol{\theta}$: 
\begin{enumerate}
\item For $j=1,...,J:$
\begin{enumerate}
\item \emph{Train on }$\mathcal{D}_{j}^{\mathrm{train}}$\emph{n}: training
set $\mathcal{D}_{j}^{\mathrm{train}}$, construct the best $k$-dimensional
linear model consisting of $(\mathbf{K},\mathbf{U}^{\top}\boldsymbol{\chi}_{0},\mathbf{V}^{\top}\boldsymbol{\chi}_{1})$
by applying the feature TCCA or nonlinear TCCA with hyper-parameters
$\boldsymbol{\theta}$
\item \emph{Validate on $\mathcal{D}_{\mathrm{test}}$}: measure the performance
of the estimated singular components by a score
\begin{equation}
\mathrm{CV}_{j}\left(\boldsymbol{\theta}\right)=\mathrm{CV}\left(\mathbf{K},\mathbf{U},\mathbf{V}|\mathcal{D}_{\mathrm{test}}\right)\label{eq:CV-i}
\end{equation}
\end{enumerate}
\item Compute cross validation score
\begin{equation}
\mathrm{MCV}\left(\boldsymbol{\theta}\right)=\frac{1}{J}\sum_{j=1}^{J}\mathrm{CV}_{j}\left(\boldsymbol{\theta}\right)
\end{equation}
\end{enumerate}
\item Select model / hyper-parameter set with maximal $\mathrm{MCV}\left(\boldsymbol{\theta}\right)$.
\end{enumerate}
The key to the above procedure is how to evaluate the estimated singular
components for given test set. It is worth pointing out that we \emph{cannot}
simply define the validation score directly as the VAMP-$r$ score
of estimated singular functions for the test data, because the singular
functions obtained from training data are usually not orthonormal
with respect to the test data.

A feasible way is to utilize the subspace variational score as proposed
for reversible Markov processes in \cite{McGibbonPande_JCP15_CrossValidation}.
For VAMP-$r$ this score becomes:
\begin{eqnarray}
\mathrm{CV}\left(\mathbf{K},\mathbf{U},\mathbf{V}|\mathcal{D}_{\mathrm{test}}\right) & = & \mathcal{R}_{r}^{\mathrm{space}}\left(\mathbf{U},\mathbf{V}|\mathcal{D}_{\mathrm{test}}\right)\nonumber \\
 & = & \left\Vert \left(\mathbf{U}^{\top}\mathbf{C}_{00}^{\mathrm{test}}\mathbf{U}\right)^{-\frac{1}{2}}\left(\mathbf{U}^{\top}\mathbf{C}_{01}^{\mathrm{test}}\mathbf{V}\right)\left(\mathbf{V}^{\top}\mathbf{C}_{11}^{\mathrm{test}}\mathbf{V}\right)^{-\frac{1}{2}}\right\Vert _{r}^{r},\label{eq:CV-subspace}
\end{eqnarray}
where $\mathcal{R}_{r}^{\mathrm{space}}$ measures the consistency
between the singular subspace and the estimated one without the constraint
of orthonormality. However, this scheme suffers from the following
limitations in practical applications: Firstly, the value of $k$
must be chosen a priori and kept fixed during the cross-validation
procedure, which implies that models with a different number of singular
components cannot be compared by the validation scores. Secondly,
computation of the validation score possibly suffers from numerical
instability. (See Appendix \ref{sec:Singular-space-based-vamp} for
detailed analysis.)

We suggest in this paper to perform the cross-validation based on
the approximation error of Koopman operators. According to conclusions
in Section \ref{subsec:Error-analysis}, feature TCCA and VAMP-2 base
nonlinear TCCA both maximize the VAMP-E score $\mathcal{R}_{E}\left(\mathbf{K},\mathbf{U},\mathbf{V}|\mathcal{D}_{\mathrm{train}}\right)$
for a given training set $\mathcal{D}_{\mathrm{train}}$. 

Therefore, we can score the performance of estimated singular components
on the test set by
\begin{equation}
\mathrm{CV}\left(\mathbf{K},\mathbf{U},\mathbf{V}|\mathcal{D}_{\mathrm{test}}\right)=\mathcal{R}_{E}\left(\mathbf{K},\mathbf{U},\mathbf{V}|\mathcal{D}_{\mathrm{test}}\right).\label{eq:validation-score-vamp-E}
\end{equation}
In contrast with the validation score \eqref{eq:CV-subspace} deduced
from the subspace VAMP-$r$ score, the validation score defined by
\eqref{eq:validation-score-vamp-E} allows us to choose $k$ according
to practical requirements: If we are only interested in a small number
of dominant singular components, we can select a fixed value of $k$.
If we want to evaluate the statistical performance of the approximate
model consisting of all available estimated singular components as
in the EDMD method, we can set $k=\min\{\mathrm{dim}(\boldsymbol{\chi}_{0}),\mathrm{dim}(\boldsymbol{\chi}_{1})\}$.
We can even view $k$ as a hyper-parameter and select a suitable rank
of the model via cross-validation. Another advantage of the VAMP-E
based validation score is that it does not involve any inverse operation
of matrices and can be stably computed.

\begin{figure}
\begin{centering}
\includegraphics[width=0.7\textwidth]{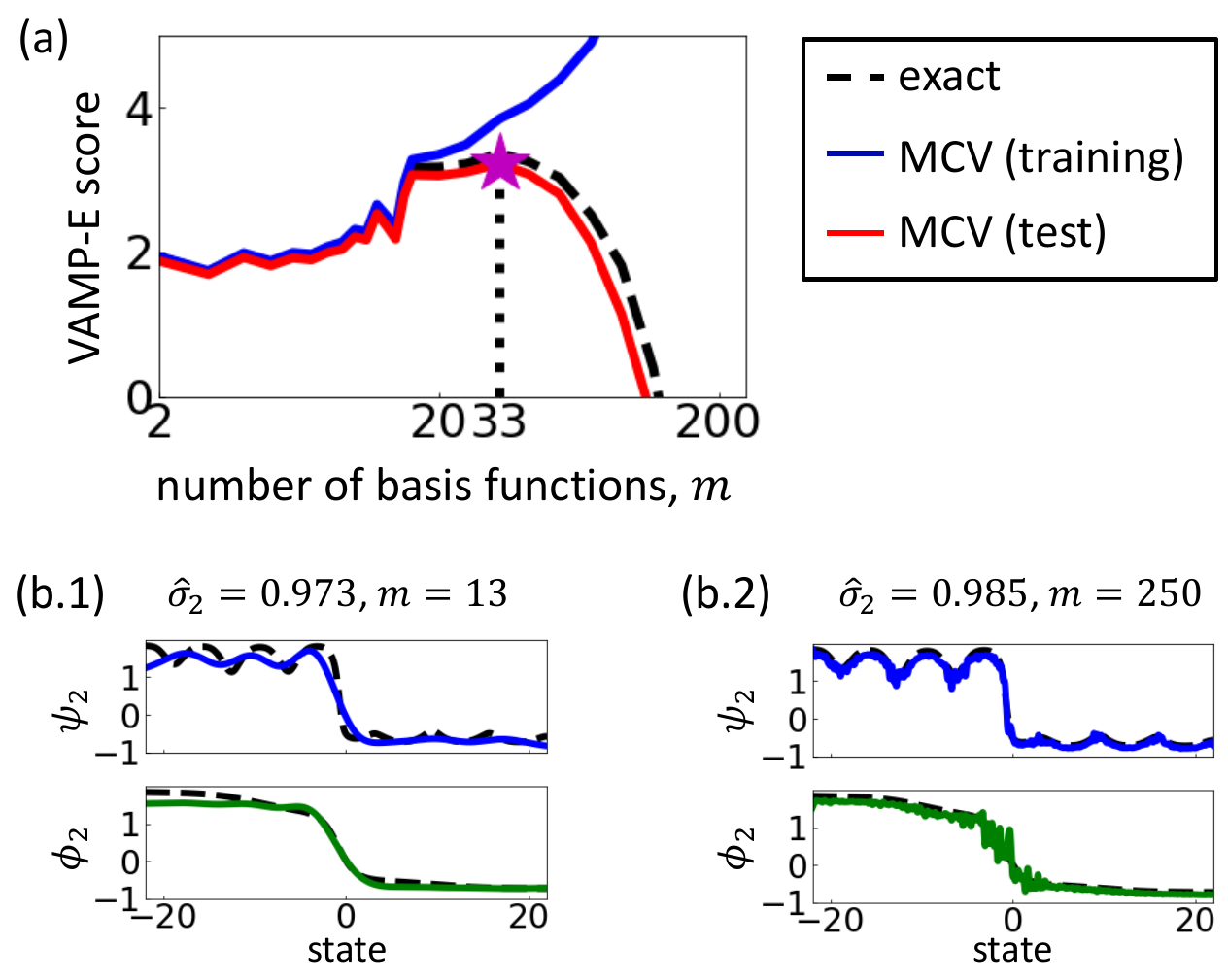}
\par\end{centering}
\caption{Cross validation for modeling the system in Example \ref{exa:one-dimensional-example-svd}.
(a) Cross-validated VAMP-E scores for the choice of the number of
basis functions $m$. The black line indicates the exact VAMP-E score
calculated according to the true model. Using cross-validation we
compute the average VAMP-E scores computed from the training sets
(blue) and the test sets (red). (b) Estimated $\psi_{2}$ and $\phi_{2}$
obtained by the nonlinear TCCA with $m=13$ and $250$.\label{fig:one-dimensional-cv}}
\end{figure}

It is worth pointing out that a validation score is proposed \cite{kurebayashi370optimal}
for cross-validation of kernel DMD based on the analysis of approximation
error of transition densities, which has a similar form to that of
VAMP-E. The theoretical and empirical comparisons between the two
scores will be performed in our future work.
\begin{example}
\label{exa:one-dimensional-example-cv}We consider here the choice
of the basis function number $m$ for the nonlinear TCCA in Example
\ref{exa:one-dimensional-example-estimation}. We use 5-fold cross-validation
with the VAMP-E score to compare different values of $m$. While the
average score computed by training sets keeps increasing with $m$,
both the cross validation score and the exact VAMP-E score achieve
their maximum value at $m=33$ as in Example \ref{exa:one-dimensional-example-estimation}
(see Fig.~\ref{fig:one-dimensional-cv}a). The optimality can also
be demonstrated by comparing Fig.~\ref{fig:one-dimensional-estimation}b
and Fig.~\ref{fig:one-dimensional-cv}b. A much smaller basis set
with $m=13$ yields large errors in the approximation of singular
functions. When $m=250$, the estimation of singular functions suffers
from overfitting and the estimated singular value is even larger than
the true value due to the statistical noise.
\end{example}

\subsection{Chapman-Kolmogorov test for choice of lag times}

Besides hyper-parameters mentioned in above, the lag time $\tau$
is also an essential parameter especially for time-continuous Markov
processes. If $\tau\to0$, the $\mathcal{K}_{\tau}$ is usually close
to the identity operator and cannot be accurately approximated by
a low-rank model, whereas a too high value of $\tau$ can cause the
loss of kinetic information in data since $\mathbb{P}(\mathbf{x}_{t+\tau}|\mathbf{x}_{t})$
is approximately independent of $\mathbf{x}_{t}$ in the case of ergodic
processes. However, the variational approach presented in this paper
is based on analysis of the approximation error of the Koopman operator
for a fixed $\tau$, so we cannot compare models with different lag
times and choose $\tau$ by the VAMP scores.

In order to address this problem, the Chapman-Kolmogorov test can
be used, which is common in building Markov state models \cite{PrinzEtAl_JCP10_MSM1}.
Let us consider the covariance
\begin{eqnarray}
\mathrm{cov}(f,g;n\tau) & \triangleq & \left\langle f,\mathcal{K}_{n\tau}g\right\rangle _{\rho_{0}(n\tau)}\nonumber \\
 & = & \mathbb{E}_{\mathbf{x}_{t}\sim\rho_{0}(n\tau)}\left[f(\mathbf{x}_{t})g(\mathbf{x}_{t+n\tau})\right]\label{eq:cov-ntau}
\end{eqnarray}
between observables $f$ and $g$ of lag time $n\tau$, which can
be estimated from data as
\begin{equation}
\mathrm{cov}^{\mathrm{emp}}(f,g;n\tau)=\frac{1}{T-n\tau}\sum_{t=1}^{T-n\tau}f\left(\mathbf{x}_{t}\right)g\left(\mathbf{x}_{t+n\tau}\right)^{\top},
\end{equation}
where $\rho_{0}(n\tau)$ is the empirical distribution of the simulation
data excluding $\{x_{t}|t>T-n\tau\}$. If our methods provide an ideal
Markov model of lag time $\tau$, the Koopman operator $\mathcal{K}_{n\tau}$
can be approximated by $\hat{\mathcal{K}}_{\tau}^{n}$, and the covariance
can also be predicted as
\begin{eqnarray}
\mathrm{cov}^{\mathrm{pred}}(f,g;n\tau) & = & \left\langle f,\hat{\mathcal{K}}_{\tau}^{n}g\right\rangle _{\rho_{0}(n\tau)}\nonumber \\
 & = & \mathbb{E}_{\mathbf{x}_{t}\sim\rho_{0}(n\tau)}\left[f(\mathbf{x}_{t})\boldsymbol{\chi}_{0}(\mathbf{x}_{t})^{\top}\right]\nonumber \\
 &  & \cdot\mathbf{U}\mathbf{R}^{n-1}\mathbf{K}\mathbf{V}^{\top}\nonumber \\
 &  & \cdot\mathbb{E}_{\mathbf{x}_{t}\sim\rho_{1}}\left[\boldsymbol{\chi}_{1}(\mathbf{x}_{t})g(\mathbf{x}_{t})\right]\label{eq:cov-pred}
\end{eqnarray}
(see Appendix \ref{sec:Calculation-Ktau-n}), where
\begin{equation}
\mathbf{R}=\mathbf{K}\cdot\mathbb{E}_{\mathbf{x}_{t}\sim\rho_{1}}\left[\mathbf{g}(\mathbf{x}_{t})^{\top}\mathbf{f}(\mathbf{x}_{t})\right].
\end{equation}

Therefore, the lag time $\tau$ can be selected according to the following
criteria in applications: (i) The lag time is smaller than the timescale
that we are interested in. (ii) The equation
\begin{equation}
\mathrm{cov}^{\mathrm{pred}}(f,g;n\tau)=\mathrm{cov}^{\mathrm{emp}}(f,g;n\tau)
\end{equation}
holds approximately for multiple observables $f,g$ and lag times
$n\tau$. In this paper, we simply set $f,g$ to be the estimated
leading singular functions since they dominate the dynamics of the
Markov process.

\section{Numerical examples}

\subsection{Double-gyre system\label{subsec:Double-gyre-system}}

\begin{figure}
\begin{centering}
\includegraphics[width=1\textwidth]{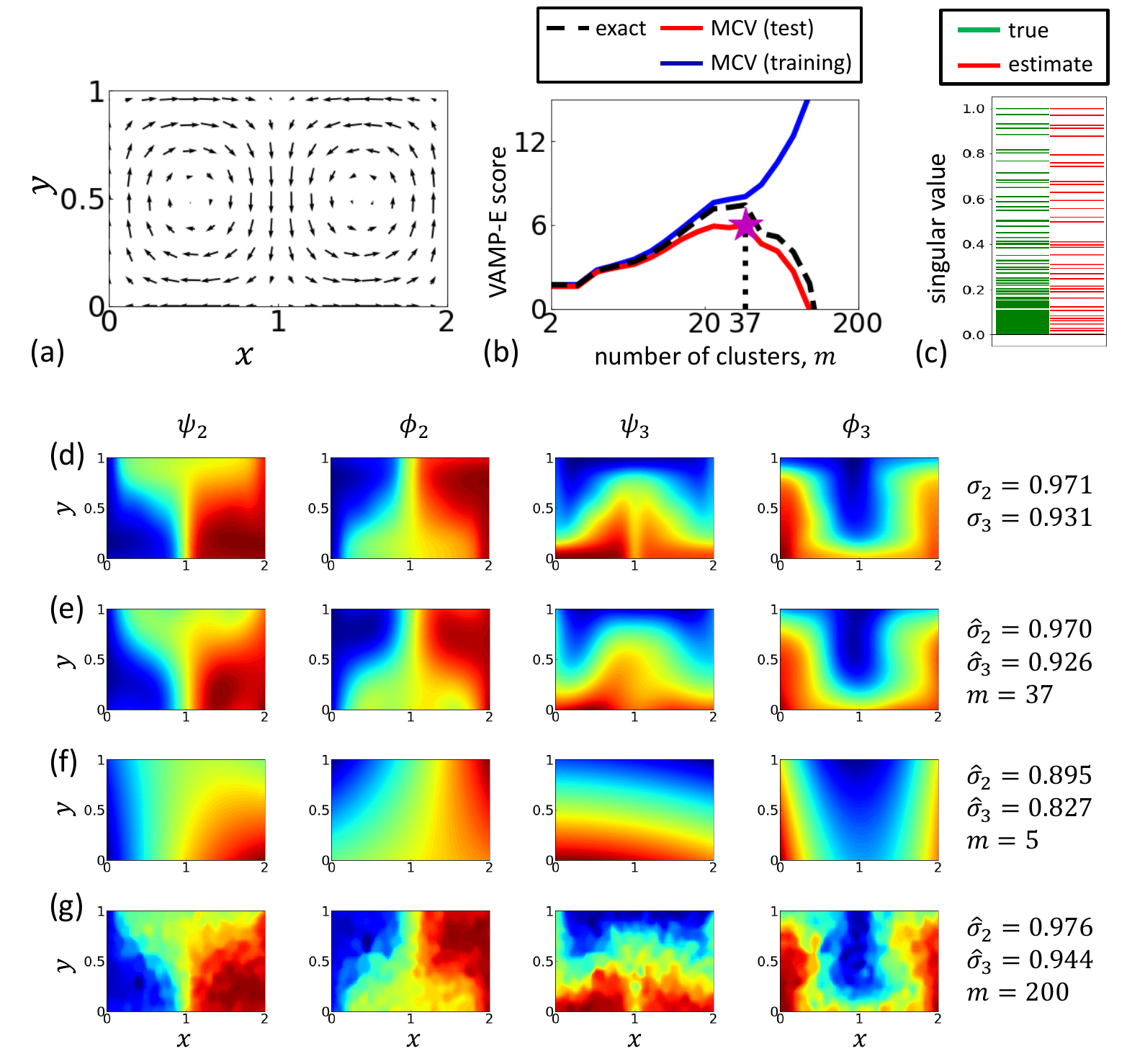}
\par\end{centering}
\caption{Modeling of the double-gyre system \eqref{eq:double-gyre}. (a) Flow
field of the system, where the arrows represent directions and magnitudes
of $(\mathrm{d}x_{t},\mathrm{d}y_{t})$ with $\epsilon=0$. (b) VAMP-E
scores of estimated models obtained from the train sets, test sets
and true model respectively. The largest MCV on test sets and exact
VAMP-E score are both achieved with $m=37$ basis functions. (c) The
true singular values and estimated ones given by the nonlinear TCCA
with $m=37$. (d) The first two nontrivial singular components. (e-g)
The estimated singular components obtained by the nonlinear TCCA with
$m=37$, $5$ and $200$.\label{fig:double-gyre}}
\end{figure}

Let's consider a stochastic double-gyre system defined by: 
\begin{eqnarray}
\mathrm{d}x_{t} & = & -\pi\,A\,\sin(\pi\,x_{t})\,\cos(\pi\,y_{t})\,\mathrm{d}t+\varepsilon\sqrt{x_{t}/4+1}\,\mathrm{d}\mathbf{W}_{t,1},\nonumber \\
\mathrm{d}y_{t} & = & \phantom{-}\pi\,A\,\cos(\pi\,x_{t})\,\sin(\pi\,y_{t})\,\mathrm{d}t+\varepsilon\,\mathrm{d}\mathbf{W}_{t,2},\label{eq:double-gyre}
\end{eqnarray}
where $\mathbf{W}_{t,1}$ and $\mathbf{W}_{t,2}$ are two independent
standard Wiener processes. The dynamics are defined on the domain
$[0,2]\times[0,1]$ with reflecting boundary. For $\varepsilon=0$,
it can be seen from the flow field depicted in Fig.~\ref{fig:double-gyre}a
that there is no transport between the left half and the right half
of the domain and both subdomains are invariant sets with measure
$\frac{1}{2}$~\cite{froyland2009almost,froyland2014almost}. For
$\varepsilon>0$, there is a small amount of transport due to diffusion
and the subdomains are almost invariant. Here we used the parameters
$A=0.25$, $\epsilon=0.1$, and lag time $\tau=2$ in analysis and
simulations. The first two nontrivial singular components are shown
in Fig.~\ref{fig:double-gyre}c, where the two almost invariant sets
are clearly visible in $\psi_{2},\phi_{2}$ and $\psi_{3},\phi_{3}$
are associated with the rotational kinetics within the almost invariant
sets.

\begin{figure}
\begin{centering}
\includegraphics[width=1\textwidth]{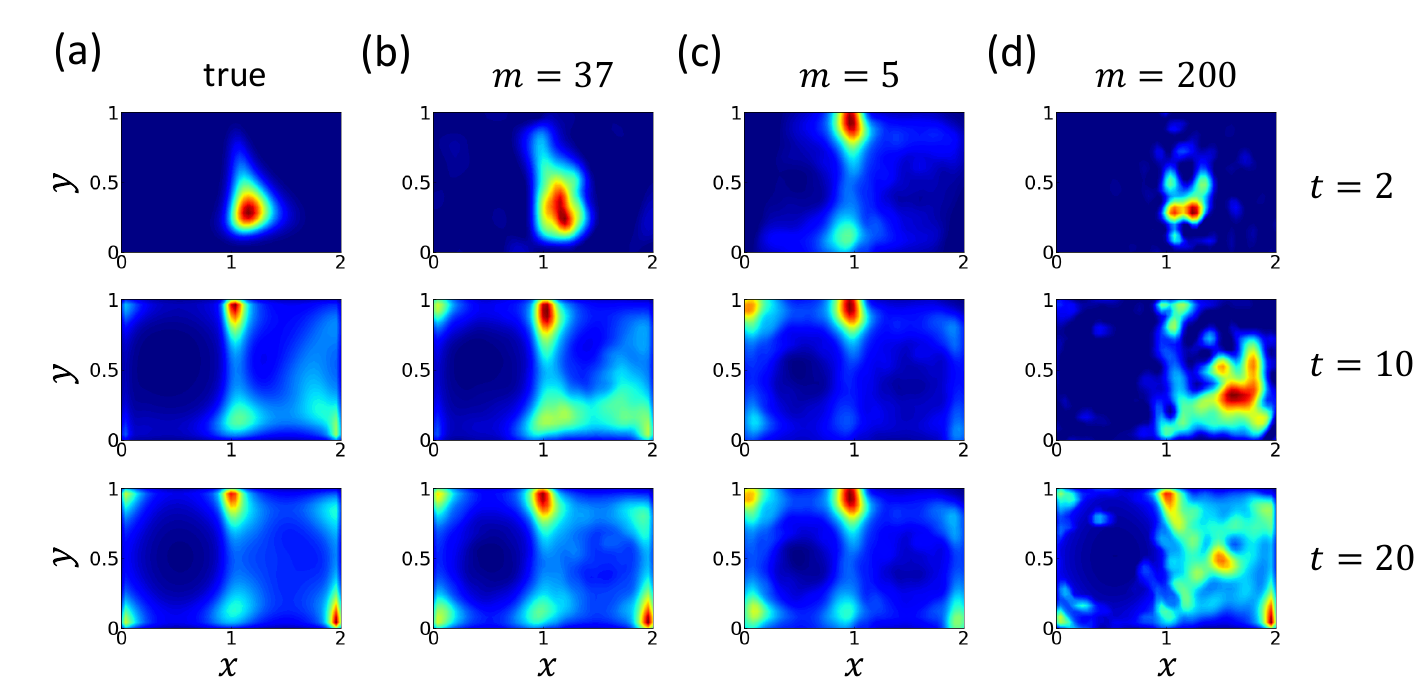}
\par\end{centering}
\caption{Probability density of state $(x_{t},y_{t})$ of the double-gyre system
predicted by (a) the full simulation model and (b) the estimated Koopman
operator obtained by the nonlinear TCCA with $m=37$, $5$ and $200$,
where the initial state is $(x_{0},y_{0})=(1.48,0.8)$.\textcolor{red}{\label{fig:double-gyre-evolution}}}
\end{figure}

\begin{figure}
\begin{centering}
\includegraphics[width=1\textwidth]{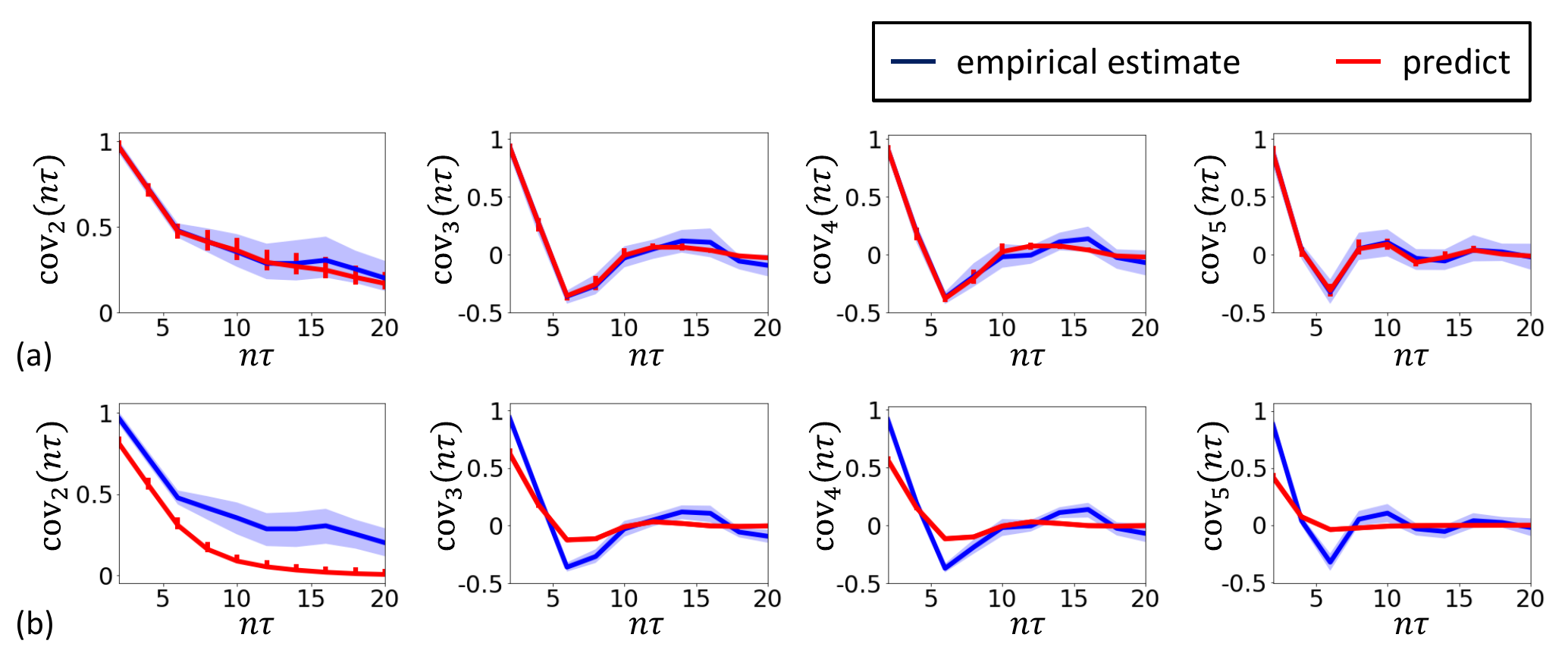}
\par\end{centering}
\caption{Chapman-Kolmogorov test for modeling the double-gyre system by nonlinear
TCCA with (a) $\tau=2$ and (b) $\tau=0.1$. The number of basis functions
$m=37$ for both cases, $\mathrm{cov}_{i}(n\tau)=\mathrm{cov}(\hat{\psi}_{i},\hat{\phi}_{i};n\tau)$
is the time-lagged covariance between $\hat{\psi}_{i},\hat{\phi}_{i}$
as defined in \eqref{eq:cov-ntau}, and $\hat{\psi}_{i},\hat{\phi}_{i}$
are the $i$th singular functions estimated with $\tau=2$. Blue lines
indicate empirical covariances directly calculated from data, red
lines indicate the predicted values given by $\hat{\mathcal{K}}_{\tau}$
as in \eqref{eq:cov-pred}, and error bars represent standard deviations
calculated from $100$ bootstrapping replicates of simulation data.\label{fig:double-gyre-CK}}
\end{figure}

We generate $10$ trajectories of length $4$ with step size $0.02$,
and perform modeling by nonlinear TCCA with basis functions
\begin{equation}
\chi_{0,i}(x,y;w)=\chi_{1,i}(x,y;w)=\frac{\exp\left(-w\left\Vert (x,y)^{\top}-\mathbf{c}_{i}\right\Vert ^{2}\right)}{\sum_{j=1}^{m}\exp\left(-w\left\Vert (x,y)^{\top}-\mathbf{c}_{j}\right\Vert ^{2}\right)},\quad\text{for }i=1,\ldots,m
\end{equation}
where $\mathbf{c}_{1},\ldots,\mathbf{c}_{m}$ are cluster centers
given by k-means algorithm, and the smoothing parameter $w$ is determined
via maximizing the VAMP-2 score given in \eqref{eq:Rr-w} (see Appendix
\ref{subsec:app-Double-gyre-system} for more details of numerical
computations). The size of basis set $m=37$ is selected by the VAMP-E
based cross-validation proposed in \ref{sec:Cross-validation} with
$5$ folds (see Fig~\ref{fig:double-gyre}b), and it can be observed
from Figs.~\ref{fig:double-gyre}c, \ref{fig:double-gyre}d and \ref{fig:double-gyre}e
that the leading singular components are accurately estimated. In
contrast, as shown in Figs.~\ref{fig:double-gyre}f and \ref{fig:double-gyre}g,
a much small value of $m$ leads to significant approximation errors
of singular components, while for a much larger value, the estimates
are obviously influenced by statistical noise. Fig.~\ref{fig:double-gyre-evolution}
illustrates that the Koopman operator estimated by nonlinear TCCA
can successfully predict the long-time evolution of the distribution
of the state. The Chapman-Kolmogorov test results are displayed in
Fig.~\ref{fig:double-gyre-CK}, which confirm that $\tau=2$ is a
suitable choice of the lag time.


\subsection{Stochastic Lorenz system\label{subsec:Stochastic-Lorenz-system}}

\begin{figure}
\begin{centering}
\includegraphics[width=1\textwidth]{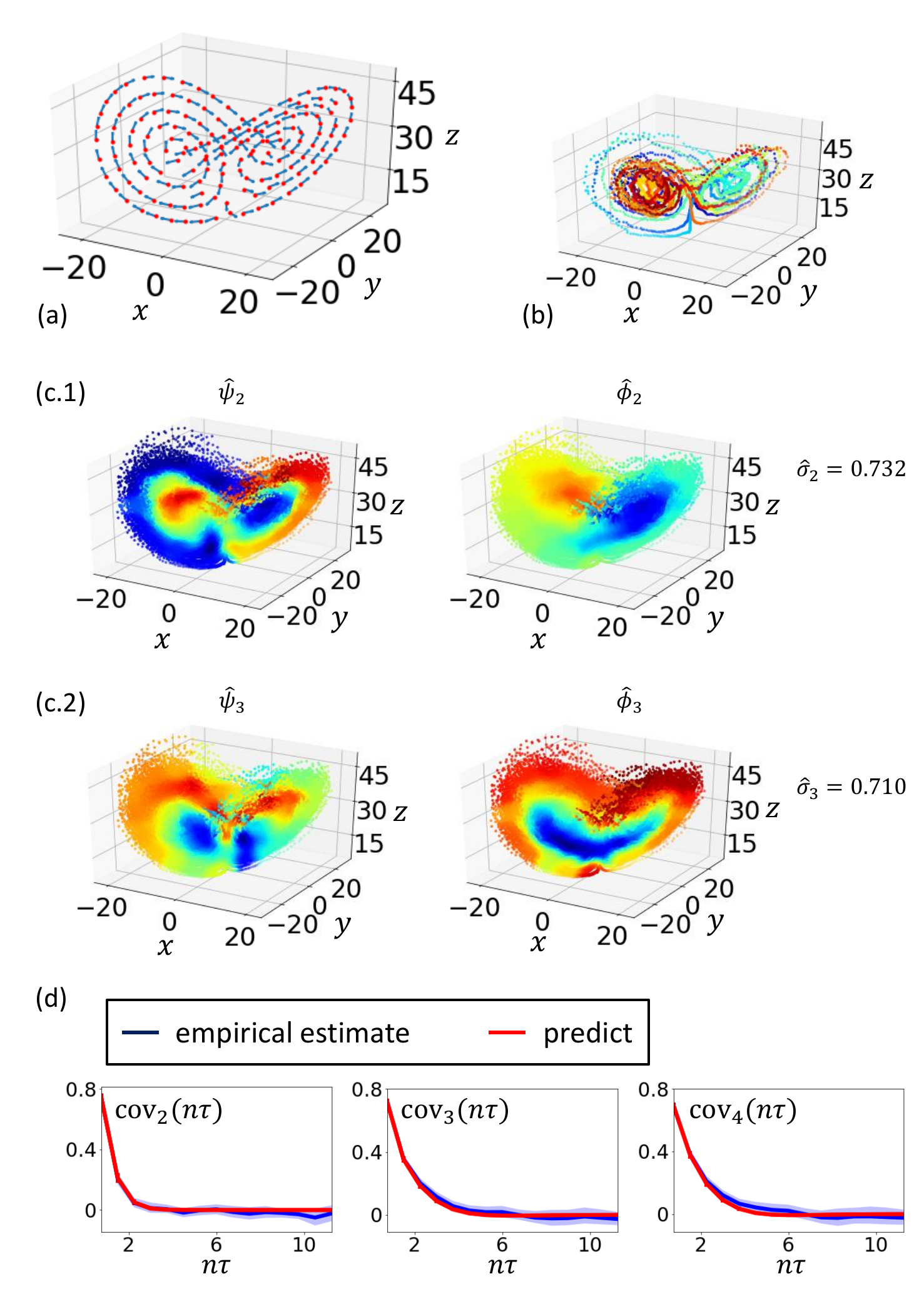}
\par\end{centering}
\caption{Modeling of the stochastic Lorenz system \eqref{eq:lorenz}. (a) Flow
field of the system, where the arrows represent the mean directions
of $(\mathrm{d}x_{t},\mathrm{d}y_{t},\mathrm{d}z_{t})$. (b) A typical
trajectory with $\epsilon=0.3$ generated by the Euler\textendash Maruyama
scheme, which is colored according to time (from blue to red). (c)
The first two nontrivial singular components computed by nonlinear
TCCA. (d) Chapman-Kolmogorov test results for $\tau=0.75$, where
$\mathrm{cov}_{i}(n\tau)=\mathrm{cov}(\hat{\psi}_{i},\hat{\phi}_{i};n\tau)$,
where error bars represent standard deviations calculated from $100$
bootstrapping replicates of simulation data.\label{fig:lorenz}}
\end{figure}

As the last example, we investigate the stochastic Lorenz system which
obeys the following stochastic differential equation:
\begin{eqnarray}
\mathrm{d}x_{t} & = & s(y-x)\,\mathrm{d}t+\epsilon x_{t}\,\mathrm{d}\mathbf{W}_{t,1},\nonumber \\
\mathrm{d}y_{t} & = & (rx_{t}-y_{t}-x_{t}z_{t})\,\mathrm{d}t+\varepsilon y_{t}\,\mathrm{d}\mathbf{W}_{t,2},\nonumber \\
\mathrm{d}z_{t} & = & (-bz_{t}+x_{t}y_{t})\,\mathrm{d}t+\varepsilon z_{t}\,\mathrm{d}\mathbf{W}_{t,2},\label{eq:lorenz}
\end{eqnarray}
with parameters $s=10$, $r=28$ and $b=8/3$. The deterministic Lorenz
system with $\epsilon=0$ is known to exhibit chaotic behavior \cite{sparrow1982lorenz}
with a strange attractor characterized by two lobes as illustrated
in Fig.~\ref{fig:lorenz}a. We generate $20$ trajectories of length
$25$ with $\epsilon=0.3$ by using the Euler\textendash Maruyama
scheme with step size $0.005$, and one of them is shown in Fig.~\ref{fig:lorenz}b.
As stated in \cite{chekroun2011stochastic}, all the trajectories
move around the deterministic attractor with small random perturbations
and switch between the two lobes.

The leading singular components computed from the simulation data
by the nonlinear TCCA are summarized in Fig~\ref{fig:lorenz}c, where
the lag time $\tau=0.75$ is determined via the Chapman-Kolmogorov
test (see Fig.~\ref{fig:lorenz}d), $\boldsymbol{\chi}_{0}=\boldsymbol{\chi}_{1}$
consist of $m$ normalized radial basis functions similar to those
used in Section \ref{subsec:Double-gyre-system}, and the selection
of $m$ is also implemented by $5$-fold cross-validation. According
to the patterns of the singular functions, the stochastic Lorenz system
can be coarse-grained into a simplified model which transitions between
four macrostates corresponding to inner and outer basins of the two
attractor lobes. In particular, the sign-boundary of $\psi_{1}$ closely
matches that between the almost invariant sets of the Lorenz flow
\cite{froyland2009almost}.

\begin{figure}
\begin{centering}
\includegraphics[width=1\textwidth]{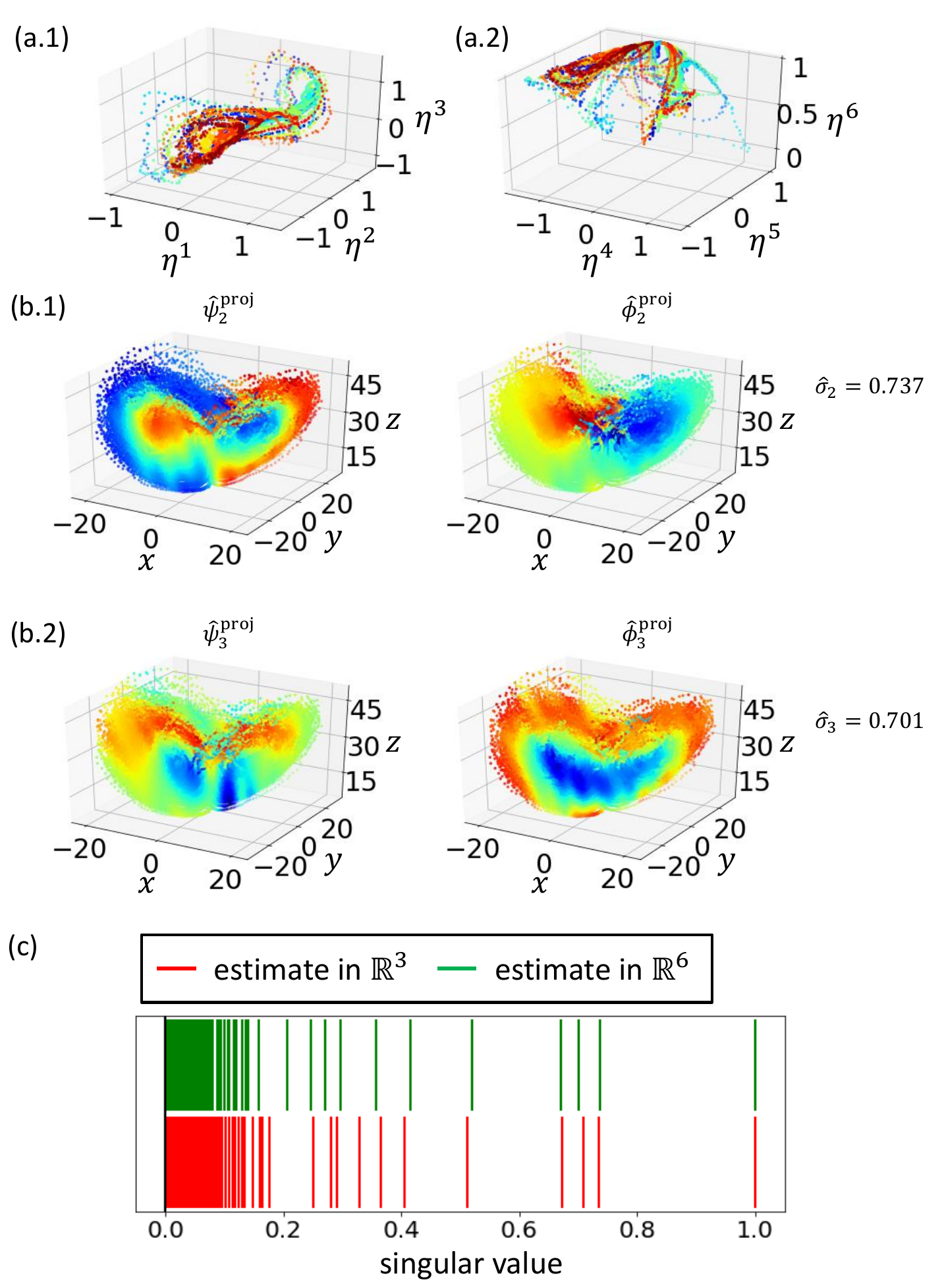}
\par\end{centering}
\caption{Modeling of the stochastic Lorenz system \eqref{eq:lorenz} in the
space of $\boldsymbol{\eta}_{t}$. (a) Plots of a typical trajectory
in spaces of $(\eta_{t}^{1},\eta_{t}^{2},\eta_{t}^{3})$ and $(\eta_{t}^{4},\eta_{t}^{5},\eta_{t}^{6})$,
which are colored according to time (from blue to red). Force field
of the system. (b) The projected singular functions in the space of
$(x_{t},y_{t},z_{t})$ computed by nonlinear TCCA. (c) The singular
values estimated from trajectories of $(x_{t},y_{t},z_{t})\in\mathbb{R}^{3}$
and $(\eta_{t}^{1},\ldots,\eta_{t}^{6})\in\mathbb{R}^{6}$. \label{fig:lorenz-highdim}}
\end{figure}

Next, we map the simulation data to a higher dimensional space via
the nonlinear transformation $\boldsymbol{\eta}_{t}=\boldsymbol{\eta}(x_{t},y_{t},z_{t})$
defined by

\begin{equation}
\begin{array}{lll}
\eta_{t}^{1}=\left(\frac{z_{t}}{50}+\frac{1}{2}\right)\cos\left(\frac{\pi x_{t}}{30}+\frac{z_{t}}{50}-1\right), &  & \eta_{t}^{2}=\left(\frac{z_{t}}{50}+\frac{1}{2}\right)\sin\left(\frac{\pi x_{t}}{30}+\frac{z_{t}}{50}-1\right),\\
\eta_{t}^{3}=\left(\frac{z_{t}}{50}+\frac{1}{2}\right)\cos\left(\frac{\pi y_{t}}{30}+\frac{z_{t}}{50}-1\right), &  & \eta_{t}^{4}=\left(\frac{z_{t}}{50}+\frac{1}{2}\right)\sin\left(\frac{\pi y_{t}}{30}+\frac{z_{t}}{50}-1\right),\\
\eta_{t}^{5}=\cos\frac{\pi\left(x_{t}+y_{t}\right)}{40}, &  & \eta_{t}^{6}=\cos\frac{\pi\left(x_{t}-y_{t}\right)}{40}.
\end{array}
\end{equation}
Fig.~\ref{fig:lorenz-highdim}a plots the transformed points of the
illustrative trajectory in Fig.~\ref{fig:lorenz}b. We utilize the
nonlinear TCCA to compute the singular components in the space of
$\boldsymbol{\eta}_{t}=(\eta_{t}^{1},\ldots,\eta_{t}^{6})$ by assuming
that the available observable is $\boldsymbol{\eta}_{t}$ instead
of $(x_{t},y_{t},z_{t})$, show in Fig.~\ref{fig:lorenz-highdim}b
the projections of the singular functions back on the three-dimensional
space
\begin{equation}
\psi_{i}^{\mathrm{proj}}(x_{t},y_{t},z_{t})=\psi_{i}(\boldsymbol{\eta}(x_{t},y_{t},z_{t})),\quad\phi_{i}^{\mathrm{proj}}(x_{t},y_{t},z_{t})=\phi_{i}(\boldsymbol{\eta}(x_{t},y_{t},z_{t})),
\end{equation}
and compares the singular values estimated from trajectories of $(x_{t},y_{t},z_{t})$
and $(\eta_{t}^{1},\ldots,\eta_{t}^{6})$. It can be seen the projected
leading singular components are almost the same as those directly
computed from the three-dimensional data, which illustrates the transformation
invariance of VAMP. Notice it is straightforward to prove that the
exact $\psi_{i}^{\mathrm{proj}}$ and $\phi_{i}^{\mathrm{proj}}$
are the solution to the variational problem \eqref{eq:variational}
in the space of $(x_{t},y_{t},z_{t})^{\top}$ if there is an inverse
mapping $\boldsymbol{\eta}^{-1}$ with $\boldsymbol{\eta}^{-1}(\boldsymbol{\eta}(x_{t},y_{t},z_{t}))\equiv(x_{t},y_{t},z_{t})$.


\section{Conclusion}

The linearized coarse-grained models of Markov systems are commonly
used in a broad range of fields, such as power systems, fluid mechanics
and molecular dynamics. Although the models were developed independently
in different communities, the VAMP proposed in this paper provides
a general framework for analysis of them, and the modeling accuracy
can be quantitatively evaluated by the VAMP-$r$ and VAMP-E scores.
Moreover, a set of data-driven methods, including feature TCCA, nonlinear
TCCA and VAMP-E based cross-validation, are developed to achieve optimal
modeling for given finite model dimensions and finite data sets.

The major challenge in real-world applications of VAMP is how to overcome
the curse of dimensionality and solve the variational problem effectively
and efficiently for high-dimensional systems. One feasible way of
addressing this challenge is to approximate singular components by
deep neural networks, which yields the concept of \emph{VAMPnet} \cite{vampnet}.
The optimal models can therefore be obtained by deep learning techniques.
Another possible way is to utilize tensor decomposition based approximation
approaches. Some tensor analysis methods have been presented based
on the reversible variational principle and EDMD \cite{NueskeEtAl_JCP15_Tensor,KlusSchuette_Arxiv15_Tensor,Klus_Arxiv16_Tensor},
and it is worth studying more general variational tensor method within
the framework of VAMP in future.

One drawback of the methods developed in this paper is that the resulting
models are possibly not valid probabilistic models with nonnegative
transition densities if only the operator error is considered, and
the probability-preserving modeling method requires further investigations.
Moreover, the applications of VAMP to detection of metastable states
\cite{DeuflhardWeber_LAA05_PCCA+}, coherent sets \cite{froyland2014almost}
and dominant cycles \cite{conrad2016finding} will also be explored
in next steps.

\appendix

\section*{Appendix}

For convenience of notation, we denote by $p_{\tau}(\mathbf{x},\mathbf{y})=\mathbb{P}(\mathbf{x}_{t+\tau}=\mathbf{y}|\mathbf{x}_{t}=\mathbf{x})$
the transition density which satisfies
\begin{equation}
\int_{A}p_{\tau}(\mathbf{x},\mathbf{y})\mathrm{d}\mathbf{y}=\mathbb{P}(\mathbf{x}_{t+\tau}\in A|\mathbf{x}_{t}=\mathbf{x})
\end{equation}
for every measurable set $A$, and define the matrix of scalar products:
\begin{equation}
\left\langle \mathbf{a},\mathbf{b}^{\top}\right\rangle _{\rho}=\left[\left\langle a_{i},b_{j}\right\rangle _{\rho}\right]\in\mathbb{R}^{m\times n}
\end{equation}
\begin{equation}
\mathcal{K}\mathbf{g}=(\mathcal{K}g_{1},\mathcal{K}g_{2},\ldots)^{\top}
\end{equation}
for $\mathbf{a}=(a_{1},a_{2},\ldots,a_{m})^{\top}$, $\mathbf{b}=(b_{1},b_{2},\ldots,b_{n})^{\top}$
and $\mathbf{g}=(g_{1},g_{2},\ldots)^{\top}$. In addition, $\mathcal{N}(\cdot|c,\sigma^{2})$
denotes the probability density function of the normal distribution
with mean $c$ and variance $\sigma^{2}$.

\section{Analysis of Koopman operators\label{sec:Analysis-of-Koopman}}

\subsection{Definition of empirical distributions\label{sec:empirical-distributions}}

We first consider the case where the simulation data consist of $S$
independent trajectories $\{\mathbf{x}_{t}^{1}\}_{t=1}^{T},\ldots,\{\mathbf{x}_{t}^{S}\}_{t=1}^{T}$
of length $T$ and the initial state $x_{0}^{s}\stackrel{\mathrm{iid}}{\sim}p_{0}\left(\mathbf{x}\right)$.
In this case, $\rho_{0}$ and $\rho_{1}$ can be defined by
\begin{equation}
\rho_{0}=\frac{1}{T-\tau}\sum_{t=1}^{T-\tau}\mathcal{P}_{t}p_{0},\quad\rho_{1}=\frac{1}{T-\tau}\sum_{t=1}^{T-\tau}\mathcal{P}_{t+\tau}p_{0},
\end{equation}
and they satisfy
\begin{equation}
\rho_{1}=\mathcal{P}_{\tau}\rho_{0},
\end{equation}
where $\mathcal{P}_{t}$ denotes the Markov propagator defined in
\eqref{eq:Markov-propagator}. We can then conclude that the estimates
of $\mathbf{C}_{00},\mathbf{C}_{11},\mathbf{C}_{01}$ given by (\ref{eq:C00}-\ref{eq:C01})
are unbiased and consistent as $S\to\infty$.

In more general cases where trajectories $\{\mathbf{x}_{t}^{1}\}_{t=1}^{T_{1}},\ldots,\{\mathbf{x}_{t}^{S}\}_{t=1}^{T_{S}}$
are generated with different initial conditions and different lengths,
the similar conclusions can be obtained by defining $\rho_{0},\rho_{1}$
as the averages of marginal distributions of $\{\mathbf{x}_{t}^{s}|1\le t\le T_{s}-\tau,1\le s\le S\}$
and $\{\mathbf{x}_{t}^{s}|1+\tau\le t\le T_{s},1\le s\le S\}$ respectively.

\subsection{Proof of Theorem \ref{thm:Optimal-approximation} \label{subsec:Singular-value-decomposition}}

Because $\mathcal{K}_{\tau}$ is a Hilbert-Schmidt operator from $\mathcal{L}_{\rho_{1}}^{2}$
to $\mathcal{L}_{\rho_{0}}^{2}$, there exists the following SVD of
$\mathcal{K}_{\tau}$:
\begin{equation}
\mathcal{K}_{\tau}g=\sum_{i=1}^{\infty}\sigma_{i}\left\langle g,\phi_{i}\right\rangle _{\rho_{1}}\psi_{i}.\label{eq:svd}
\end{equation}
Due to the orthonormality of right singular functions, the projection
of any function $g\in\mathcal{L}_{\rho_{1}}^{2}$ onto the space spanned
by $\{\phi_{1},\ldots,\phi_{k}\}$ can be written as $\sum_{i=1}^{k}\left\langle g,\phi_{i}\right\rangle _{\rho_{1}}\phi_{i}$.
Then $\hat{\mathcal{K}}_{\tau}$ defined by \eqref{eq:svd-truncated}
is the approximate Koopman operator deduced from model \eqref{eq:optimal-linear-model},
and it is the best rank $k$ approximation to $\mathcal{K}_{\tau}$
in Hilbert-Schmidt norm according to the generalized Eckart-Young
Theorem (see Theorem 4.4.7 in \cite{hsing2015theoretical}).

Since the adjoint operator $\mathcal{K}_{\tau}^{*}$ of $\mathcal{K}_{\tau}$
satisfies
\begin{eqnarray*}
\left\langle f,\mathcal{K}_{\tau}^{*}\mathbbm1\right\rangle _{\rho_{1}} & = & \left\langle \mathcal{K}_{\tau}f,\mathbbm1\right\rangle _{\rho_{0}}\\
 & = & \int\mathbb{E}[f(\mathbf{x}_{t+\tau})|\mathbf{x}_{t}=\mathbf{x}]\rho_{0}(\mathbf{x})\mathrm{d}\mathbf{x}\\
 & = & \int\mathbb{E}[f(\mathbf{x})]\rho_{1}(\mathbf{x})\mathrm{d}\mathbf{x}\\
 & = & \left\langle f,\mathbbm1\right\rangle _{\rho_{1}}
\end{eqnarray*}
for all $f$, we can obtain
\begin{equation}
\mathcal{K}_{\tau}^{*}\mathbbm1=\mathcal{K}_{\tau}\mathbbm1=\mathbbm1,
\end{equation}
and conclude from Proposition 2 in \cite{froyland2013analytic} that
$(\sigma_{1},\phi_{1},\psi_{1})=(1,\mathbbm1,\mathbbm1)$.

\subsection{Transition densities deduced from Koopman operators\label{subsec:ptau-and-Koopman}}

The Koopman operator can also be written as
\begin{equation}
\mathcal{K}_{\tau}g(\mathbf{x})=\int p_{\tau}(\mathbf{x},\mathbf{y})g(\mathbf{y})\mathrm{d}\mathbf{y}\label{eq:Koopman-density}
\end{equation}
if the transition density is given, which implies that
\begin{equation}
\mathcal{K}_{\tau}\delta_{\mathbf{y}}(\mathbf{x})=p_{\tau}(\mathbf{x},\mathbf{y}).
\end{equation}
Then the transition density deduced from the approximate Koopman operator
$\hat{\mathcal{K}}_{\tau}$ defined by \eqref{eq:svd-truncated} is
\begin{eqnarray}
\hat{p}_{\tau}(\mathbf{x},\mathbf{y}) & = & \hat{\mathcal{K}}_{\tau}\delta_{\mathbf{y}}(\mathbf{x})\nonumber \\
 & = & \sum_{i=1}^{k}\sigma_{i}\psi_{i}(\mathbf{x})\phi_{i}(\mathbf{y})\rho_{1}(\mathbf{y}).\label{eq:approximate-transition-density}
\end{eqnarray}
From \eqref{eq:Koopman-density}, we can show that
\begin{eqnarray}
\left\Vert \mathcal{K}_{\tau}\right\Vert _{\mathrm{HS}}^{2} & = & \sum_{i}\left\langle \mathcal{K}_{\tau}\phi_{i},\mathcal{K}_{\tau}\phi_{i}\right\rangle _{\rho_{0}}\nonumber \\
 & = & \int\sum_{i}\left(\int p(\mathbf{x},\mathbf{y})\phi_{i}(\mathbf{y})\mathrm{d}\mathbf{y}\right)^{2}\rho_{0}(\mathbf{x})\mathrm{d}\mathbf{x}\nonumber \\
 & = & \int\sum_{i}\left(\int\frac{p(\mathbf{x},\mathbf{y})}{\rho_{1}(\mathbf{y})}\cdot\phi_{i}(\mathbf{y})\cdot\rho_{1}(\mathbf{y})\mathrm{d}\mathbf{y}\right)^{2}\rho_{0}(\mathbf{x})\mathrm{d}\mathbf{x}\nonumber \\
 & = & \int\left(\int\left(\frac{p(\mathbf{x},\mathbf{y})}{\rho_{1}(\mathbf{y})}\right)^{2}\cdot\rho_{1}(\mathbf{y})\mathrm{d}\mathbf{y}\right)\rho_{0}(\mathbf{x})\mathrm{d}\mathbf{x}\nonumber \\
 & = & \iint\frac{\rho_{0}(\mathbf{x})}{\rho_{1}(\mathbf{y})}p(\mathbf{x},\mathbf{y})^{2}\mathrm{d}\mathbf{x}\mathrm{d}\mathbf{y},
\end{eqnarray}
and
\begin{equation}
\left\Vert \hat{\mathcal{K}}_{\tau}-\mathcal{K}_{\tau}\right\Vert _{\mathrm{HS}}^{2}=\iint\frac{\rho_{0}(\mathbf{x})}{\rho_{1}(\mathbf{y})}\left(\hat{p}(\mathbf{x},\mathbf{y})-p(\mathbf{x},\mathbf{y})\right)^{2}\mathrm{d}\mathbf{x}\mathrm{d}\mathbf{y},
\end{equation}
i.e., the operator error between $\hat{\mathcal{K}}_{\tau}$ and $\mathcal{K}_{\tau}$
can be represented by the error between $\hat{p}_{\tau}$ and $p_{\tau}$.

It is worth pointing out that the approximate transition density in
\eqref{eq:approximate-transition-density} satisfies the normalization
constraint with
\begin{eqnarray}
\int\hat{p}_{\tau}(\mathbf{x},\mathbf{y})\mathrm{d}\mathbf{y} & = & \sum_{i=1}^{k}\sigma_{i}\psi_{i}(\mathbf{x})\left\langle \phi_{j},\mathbbm1\right\rangle _{\rho_{1}}\nonumber \\
 & = & \sigma_{1}\psi_{1}(\mathbf{x})\nonumber \\
 & \equiv & 1,
\end{eqnarray}
but $\hat{p}_{\tau}(\mathbf{x},\mathbf{y})$ is possibly negative
for some $\mathbf{x},\mathbf{y}$. Thus, the approximate Koopman operators
and transition densities are not guaranteed to yield valid probabilistic
models, although they can still be utilized to quantitative analysis
of Markov processes.

\subsection{Sufficient conditions for Theorem \ref{thm:Optimal-approximation}\label{subsec:Sufficient-conditions}}

We show here $\mathcal{L}_{\rho_{0}}^{2},\mathcal{L}_{\rho_{1}}^{2}$
are separable Hilbert spaces and $\mathcal{K}_{\tau}:\mathcal{L}_{\rho_{1}}^{2}\mapsto\mathcal{L}_{\rho_{0}}^{2}$
is Hilbert-Schmidt if one of the following conditions is satisfied:

\begin{condition}The state space of the Markov process is a finite
set.

\end{condition}
\begin{proof}
The proof is trivial by considering $\mathcal{K}_{\tau}$ is a linear
operator between finite-dimensional spaces, and thus omitted.
\end{proof}

\begin{condition}The state space of the Markov process is $\mathbb{R}^{d}$,
$\rho_{0}(\mathbf{x}),\rho_{1}(\mathbf{y})$ are positive for all
$\mathbf{x},\mathbf{y}\in\mathbb{R}^{d}$, and there exists a constant
$M$ so that
\begin{equation}
p_{\tau}(\mathbf{x},\mathbf{y})\le M\rho_{1}(\mathbf{y}),\quad\forall\mathbf{x},\mathbf{y}
\end{equation}

\end{condition}
\begin{proof}
Let $\{e_{1},e_{2},\ldots\}$ be a orthonormal basis of $\mathcal{L}^{2}(\mathbb{R}^{d})$.
Then $\mathcal{L}_{\rho_{0}}^{2},\mathcal{L}_{\rho_{1}}^{2}$ are
separable because they have the countable orthonormal bases $\{\rho_{0}^{-\frac{1}{2}}e_{1},\rho_{0}^{-\frac{1}{2}}e_{2},\ldots\}$
and $\{\rho_{1}^{-\frac{1}{2}}e_{1},\rho_{1}^{-\frac{1}{2}}e_{2},\ldots\}$.

Now we prove that $\left\Vert \mathcal{K}_{\tau}\right\Vert _{\mathrm{HS}}<\infty$.
Because
\begin{eqnarray}
\iint\frac{\rho_{0}(\mathbf{x})}{\rho_{1}(\mathbf{y})}p_{\tau}(\mathbf{x},\mathbf{y})^{2}\mathrm{d}\mathbf{x}\mathrm{d}\mathbf{y} & \le & \iint M\rho_{0}(\mathbf{x})p_{\tau}(\mathbf{x},\mathbf{y})\mathrm{d}\mathbf{x}\mathrm{d}\mathbf{y}\nonumber \\
 & = & M,
\end{eqnarray}
the operator $\mathcal{S}$ defined by
\begin{equation}
\mathcal{S}f(\mathbf{x})=\int\sqrt{\frac{\rho_{0}(\mathbf{x})}{\rho_{1}(\mathbf{y})}}p_{\tau}(\mathbf{x},\mathbf{y})f(\mathbf{y})\mathrm{d}\mathbf{y}
\end{equation}
is a Hilbert-Schmidt integral operator from $\mathcal{L}^{2}(\mathbb{R}^{d})$
to $\mathcal{L}^{2}(\mathbb{R}^{d})$ with $\left\Vert \mathcal{S}\right\Vert _{\mathrm{HS}}^{2}\le M$
\cite{renardy2004introduction}. Therefore,
\begin{eqnarray}
\left\Vert \mathcal{K}_{\tau}\right\Vert _{\mathrm{HS}}^{2} & = & \sum_{i}\left\langle \mathcal{K}_{\tau}\rho_{1}^{-\frac{1}{2}}e_{i},\mathcal{K}_{\tau}\rho_{1}^{-\frac{1}{2}}e_{i}\right\rangle _{\rho_{0}}\nonumber \\
 & = & \sum_{i}\left\langle \mathcal{S}e_{i},\mathcal{S}e_{i}\right\rangle \nonumber \\
 & = & \left\Vert \mathcal{S}\right\Vert _{\mathrm{HS}}^{2}\le M,
\end{eqnarray}
where $\left\langle f,g\right\rangle =\int f(\mathbf{x})g(\mathbf{x})\mathrm{d}\mathbf{x}$.
\end{proof}

\subsection{Koopman operators of deterministic systems\label{subsec:deterministic-Koopman-operators}}

For the completeness of paper, we prove here the following proposition
by contradiction: \emph{The Koopman operator $\mathcal{K}_{\tau}$
of the deterministic system $\mathbf{x}_{t+\tau}=F(\mathbf{x}_{t})$
defined by
\begin{equation}
\mathcal{K}_{\tau}g(\mathbf{x})=g(F(\mathbf{x}))
\end{equation}
is not a compact operator from $\mathcal{L}_{\rho_{1}}^{2}$ to $\mathcal{L}_{\rho_{0}}^{2}$
if $\mathcal{L}_{\rho_{1}}^{2}$ is infinite-dimensional.}

Assume that $\mathcal{K}_{\tau}$ is compact. Then, the SVD \eqref{eq:svd}
of $\mathcal{K}_{\tau}$ exists with $\sigma_{i}\to0$ as $i\to\infty$,
and there is $j$ so that $0\le\sigma_{j}<1$. This implies $\left\langle \mathcal{K}_{\tau}\psi_{j},\mathcal{K}_{\tau}\psi_{j}\right\rangle _{\rho_{0}}=\sigma_{j}^{2}<1$.
However, according to the definition of the Koopman operator, $\left\langle \mathcal{K}_{\tau}\psi_{j},\mathcal{K}_{\tau}\psi_{j}\right\rangle _{\rho_{0}}=\left\langle \psi_{j},\psi_{j}\right\rangle _{\rho_{1}}=1$,
which leads to a contradiction. We can conclude that $\mathcal{K}_{\tau}$
is not compact and hence not Hilbert-Schmidt.

\section{Markov propagators\label{sec:svd-P}}

The Markov propagator $\mathcal{P}_{\tau}$ is defined by

\begin{eqnarray}
p_{t+\tau}\left(\mathbf{x}\right) & = & \mathcal{P}_{\tau}p_{t}\left(\mathbf{x}\right)\nonumber \\
 & \triangleq & \int p_{\tau}\left(\mathbf{y},\mathbf{x}\right)p_{t}\left(\mathbf{y}\right)\mathrm{d}\mathbf{y},\label{eq:Markov-propagator}
\end{eqnarray}
with $p_{t}\left(\mathbf{x}\right)=\mathbb{P}(\mathbf{x}_{t}=\mathbf{x})$
being the probability density of $\mathbf{x}_{t}$. According to the
SVD of the Koopman operator given in \eqref{eq:svd}, we have
\begin{equation}
p_{\tau}\left(\mathbf{x},\mathbf{y}\right)=\mathcal{K}_{\tau}\delta_{\mathbf{y}}\left(\mathbf{x}\right)=\sum_{i=1}^{\infty}\sigma_{i}\psi_{i}\left(\mathbf{x}\right)\phi_{i}\left(\mathbf{y}\right)\rho_{1}\left(\mathbf{y}\right).
\end{equation}
Then
\begin{eqnarray}
\mathcal{P}_{\tau}p_{t}\left(\mathbf{x}\right) & = & \int p_{\tau}\left(\mathbf{y},\mathbf{x}\right)p_{t}\left(\mathbf{y}\right)\mathrm{d}\mathbf{y}\nonumber \\
 & = & \sum_{i=1}^{\infty}\sigma_{i}\left\langle p_{t},\rho_{0}\psi_{i}\right\rangle _{\rho_{0}^{-1}}\rho_{1}\left(\mathbf{x}\right)\phi_{i}\left(\mathbf{x}\right).
\end{eqnarray}
Where the following normalizations were used:
\begin{eqnarray}
\left\langle \rho_{0}\psi_{i},\rho_{0}\psi_{j}\right\rangle _{\rho_{0}^{-1}} & = & \left\langle \psi_{i},\psi_{j}\right\rangle _{\rho_{0}}=1_{i=j}\\
\left\langle \rho_{1}\phi_{i},\rho_{1}\phi_{j}\right\rangle _{\rho_{1}^{-1}} & = & \left\langle \phi_{i},\phi_{j}\right\rangle _{\rho_{1}}=1_{i=j},
\end{eqnarray}
The SVD of $\mathcal{P}_{\tau}$ can be written as
\begin{equation}
\mathcal{P}_{\tau}p_{t}=\sum_{i=1}^{\infty}\sigma_{i}\left\langle p_{t},\rho_{0}\psi_{i}\right\rangle _{\rho_{0}^{-1}}\rho_{1}\phi_{i}.\label{eq:svd-P}
\end{equation}

\section{Proof of the variational principle\label{sec:Proof-of-Proposition-variational-svd}}

Notice that $\mathbf{f}$ and $\mathbf{g}$ can be expressed as
\begin{equation}
\mathbf{f}=\mathbf{D}_{0}^{\top}\boldsymbol{\psi},\quad\mathbf{g}=\mathbf{D}_{1}^{\top}\boldsymbol{\phi}
\end{equation}
where $\boldsymbol{\psi}=(\psi_{1},\psi_{2},\ldots)^{\top}$, $\boldsymbol{\phi}=(\phi_{1},\phi_{2},\ldots)^{\top}$
and $\mathbf{D}_{0},\mathbf{D}_{1}\in\mathbb{R}^{\infty\times k}$.

Since
\begin{eqnarray}
\left\langle \mathbf{f},\mathbf{f}^{\top}\right\rangle _{\rho_{0}} & = & \mathbf{D}_{0}^{\top}\mathbf{D}_{0}\\
\left\langle \mathbf{g},\mathbf{g}^{\top}\right\rangle _{\rho_{1}} & = & \mathbf{D}_{1}^{\top}\mathbf{D}_{1}
\end{eqnarray}
and
\begin{eqnarray*}
\left\langle \mathbf{f},\mathcal{K}_{\tau}\mathbf{g}^{\top}\right\rangle _{\rho_{0}} & = & \mathbf{D}_{0}^{\top}\left\langle \boldsymbol{\psi},\mathcal{K}_{\tau}\boldsymbol{\phi}^{\top}\right\rangle _{\rho_{0}}\mathbf{D}_{1}\\
 & = & \mathbf{D}_{0}^{\top}\left\langle \boldsymbol{\psi},\boldsymbol{\psi}^{\top}\right\rangle _{\rho_{0}}\boldsymbol{\Sigma}\mathbf{D}_{1}\\
 & = & \mathbf{D}_{0}^{\top}\boldsymbol{\Sigma}\mathbf{D}_{1},
\end{eqnarray*}
the optimization problem can be equivalently written as
\begin{equation}
\max_{\mathbf{D}_{0}^{\top}\mathbf{D}_{0}=\mathbf{I},\mathbf{D}_{1}^{\top}\mathbf{D}_{1}=\mathbf{I}}\sum_{i=1}^{k}\left(\sigma_{i}\mathbf{d}_{0,i}^{\top}\mathbf{d}_{1,i}\right)^{r},
\end{equation}
where $\boldsymbol{\Sigma}=\mathrm{diag}(\sigma_{1},\sigma_{2},\ldots)$.According
to the Cauchy-Schwarz inequality and the conclusion in Section I.3.C
of \cite{marshall1979inequalities}, we have
\begin{equation}
\sum_{i=1}^{k}\left|\sigma_{i}\mathbf{d}_{0,i}^{\top}\mathbf{d}_{1,i}\right|\le\sum_{i=1}^{k}\sigma_{i}
\end{equation}
and
\begin{equation}
\sum_{i=1}^{k}\left(\sigma_{i}\mathbf{d}_{0,i}^{\top}\mathbf{d}_{1,i}\right)^{r}\le\sum_{i=1}^{k}\left|\sigma_{i}\mathbf{d}_{0,i}^{\top}\mathbf{d}_{1,i}\right|^{r}\le\sum_{i=1}^{k}\sigma_{i}^{r}
\end{equation}
under the constraint $\mathbf{D}_{0}^{\top}\mathbf{D}_{0}=\mathbf{I},\mathbf{D}_{1}^{\top}\mathbf{D}_{1}=\mathbf{I}$.
The variational principle can then be proven by considering
\begin{equation}
\sum_{i=1}^{k}\left(\sigma_{i}\mathbf{d}_{0,i}^{\top}\mathbf{d}_{1,i}\right)^{r}=\sum_{i=1}^{k}\sigma_{i}^{r}
\end{equation}
when the first $k$ rows of $\mathbf{D}_{0}$ and $\mathbf{D}_{1}$
are identity matrix.

\section{Variational principle of reversible Markov processes\label{sec:Variational-principle-reversible}}

The variational principle of reversible Markov processes can be summarized
as follows: If the Markov process $\{\mathbf{x}_{t}\}$ is time-reversible
with respect to stationary distribution $\mu$ and all eigenvalues
of $\mathcal{K}_{\tau}$ is nonnegative, then
\begin{align}
\sum_{i=1}^{k}\lambda_{i}^{r} & =\max\sum_{i=1}^{k}\left\langle f_{i},\mathcal{K}_{\tau}f_{i}\right\rangle _{\mu}^{r}\nonumber \\
s.t. & \left\langle f_{i},f_{j}\right\rangle _{\mu}=1_{i=j}
\end{align}
for $r\ge1$ and the maximal value is achieved with $f_{i}=\psi_{i}$,
where $\psi_{i}$ denotes the eigenfunction with the $i$th largest
eigenvalue $\lambda_{i}$. The proof is trivial by using variational
principle of general Markov processes and considering that the eigendecomposition
of $\mathcal{K}_{\tau}$ is equivalent to its SVD if $\{\mathbf{x}_{t}\}$
is time-reversible and $\rho_{0}=\rho_{1}=\mu$.

\section{Analysis of estimation algorithms}

\subsection{Correctness of feature TCCA\label{sec:Correctness-of-feature-TCCA}}

We show in this appendix that the feature TCCA algorithm described
in Section \ref{subsec:feature-TCCA} solves the optimization problem
\eqref{eq:vamp-matrix}.

Let $\mathbf{U}^{\prime}=\mathbf{C}_{00}^{\frac{1}{2}}\mathbf{U}=(\mathbf{u}_{1}^{\prime},\ldots,\mathbf{u}_{k}^{\prime})$
and $\mathbf{V}^{\prime}=\mathbf{C}_{00}^{\frac{1}{2}}\mathbf{V}=(\mathbf{v}_{1}^{\prime},\ldots,\mathbf{v}_{k}^{\prime})$,
\eqref{eq:vamp-matrix} can be equivalently expressed as
\begin{align}
\max_{\mathbf{U}^{\prime},\mathbf{V}^{\prime}} & \sum_{i=1}^{k}\left(\mathbf{u}_{i}^{\prime\top}\mathbf{C}_{00}^{-\frac{1}{2}}\mathbf{C}_{01}\mathbf{C}_{11}^{-\frac{1}{2}}\mathbf{v}_{i}^{\prime}\right)^{r}\nonumber \\
\mathrm{s.t.} & \mathbf{U}^{\prime\top}\mathbf{U}^{\prime}=\mathbf{I}\nonumber \\
 & \mathbf{V}^{\prime\top}\mathbf{V}^{\prime}=\mathbf{I}.
\end{align}
According to the Cauchy-Schwarz inequality and the conclusion in Section
I.3.C of \cite{marshall1979inequalities}, we have
\begin{eqnarray}
\sum_{i=1}^{k}\left(\mathbf{u}_{i}^{\prime\top}\mathbf{C}_{00}^{-\frac{1}{2}}\mathbf{C}_{01}\mathbf{C}_{11}^{-\frac{1}{2}}\mathbf{v}_{i}^{\prime}\right)^{r} & \le & \sum_{i=1}^{k}\left|\mathbf{u}_{i}^{\prime\top}\mathbf{C}_{00}^{-\frac{1}{2}}\mathbf{C}_{01}\mathbf{C}_{11}^{-\frac{1}{2}}\mathbf{v}_{i}^{\prime}\right|^{r}\nonumber \\
 & \le & \sum_{i=1}^{k}s_{i}^{r}
\end{eqnarray}
under the constraints $\mathbf{U}^{\prime\top}\mathbf{U}^{\prime}=\mathbf{I},\mathbf{V}^{\prime\top}\mathbf{V}^{\prime}=\mathbf{I}$,
where $s_{i}$ is the $i$th largest singular value of $\mathbf{C}_{00}^{-\frac{1}{2}}\mathbf{C}_{01}\mathbf{C}_{11}^{-\frac{1}{2}}$.
Considering the equalities hold in the above when $\mathbf{U}^{\prime},\mathbf{V}^{\prime}$
are the first $k$ left and right singular vectors of $\mathbf{C}_{00}^{-\frac{1}{2}}\mathbf{C}_{01}\mathbf{C}_{11}^{-\frac{1}{2}}$,
we get
\begin{align}
\max_{\mathbf{U},\mathbf{V}} & \mathcal{R}_{r}(\mathbf{U},\mathbf{V})=\sum_{i=1}^{k}s_{i}^{r}\nonumber \\
\mathrm{s.t.} & \mathbf{U}^{\top}\mathbf{C}_{00}\mathbf{U}=\mathbf{I}\nonumber \\
 & \mathbf{V}^{\top}\mathbf{C}_{11}\mathbf{V}=\mathbf{I},\label{eq:max-Rr}
\end{align}
and the correctness of the feature TCCA can then be proved.

Furthermore, if $k=\min\{\mathrm{dim}\left(\boldsymbol{\chi}_{0}\right),\mathrm{dim}\left(\boldsymbol{\chi}_{1}\right)\}$,
we can get
\begin{equation}
\max_{\mathbf{U},\mathbf{V}}\mathcal{R}_{r}(\mathbf{U},\mathbf{V})=\left\Vert \mathbf{C}_{00}^{-\frac{1}{2}}\mathbf{C}_{01}\mathbf{C}_{11}^{-\frac{1}{2}}\right\Vert _{r}^{r}\label{eq:max-Rr-matrix-norm}
\end{equation}
under the orthonormality constraints.

\subsection{Feature TCCA of projected Koopman operators\label{subsec:Feature-TCCA-proj}}

Define projection operators
\begin{eqnarray}
\mathcal{Q}_{\boldsymbol{\chi}_{0}}f & \triangleq & \argmin_{f^{\prime}\in\mathrm{span}\{\chi_{0,1},\chi_{0,2},\ldots\}}\left\langle f^{\prime}-f,f^{\prime}-f\right\rangle _{\rho_{0}}\nonumber \\
 & = & \left\langle f,\boldsymbol{\chi}_{0}^{\top}\right\rangle _{\rho_{0}}\mathbf{C}_{00}^{-1}\boldsymbol{\chi}_{0},\\
\mathcal{Q}_{\boldsymbol{\chi}_{1}}g & \triangleq & \argmin_{g^{\prime}\in\mathrm{span}\{\chi_{1,1},\chi_{1,2},\ldots\}}\left\langle g^{\prime}-g,g^{\prime}-g\right\rangle _{\rho_{1}}\nonumber \\
 & = & \left\langle g,\boldsymbol{\chi}_{1}^{\top}\right\rangle _{\rho_{1}}\mathbf{C}_{11}^{-1}\boldsymbol{\chi}_{1},
\end{eqnarray}
and let $\mathcal{K}_{\tau}^{\mathrm{proj}}=\mathcal{Q}_{\boldsymbol{\chi}_{0}}\mathcal{K}_{\tau}\mathcal{Q}_{\boldsymbol{\chi}_{1}}$
be the projection of the Koopman operator $\mathcal{K}_{\tau}$ onto
the subspaces of $\boldsymbol{\chi}_{0},\boldsymbol{\chi}_{1}$. Then
for any $f=\mathbf{u}^{\top}\boldsymbol{\chi}_{0}\in\mathrm{span}\{\chi_{0,1},\chi_{0,2},\ldots\}$
and $g=\mathbf{v}^{\top}\boldsymbol{\chi}_{1}\in\mathrm{span}\{\chi_{1,1},\chi_{1,2},\ldots\}$,
\begin{eqnarray}
\left\langle f,\mathcal{K}_{\tau}^{\mathrm{proj}}g\right\rangle _{\rho_{0}} & = & \left\langle g,\boldsymbol{\chi}_{1}^{\top}\right\rangle _{\rho_{1}}\mathbf{C}_{11}^{-1}\mathbf{C}_{01}^{\top}\mathbf{C}_{00}^{-1}\left\langle \boldsymbol{\chi}_{0},f\right\rangle _{\rho_{0}}\nonumber \\
 & = & \mathbf{u}^{\top}\mathbf{C}_{01}\mathbf{v}\nonumber \\
 & = & \left\langle f,\mathcal{K}_{\tau}g\right\rangle _{\rho_{0}},
\end{eqnarray}
which implies that Eq.~\eqref{eq:vamp-matrix} can also be interpreted
as the variational problem for the feature TCCA of $\mathcal{K}_{\tau}^{\mathrm{proj}}$.

By ignoring the statistical noise, we can conclude from Theorem \ref{thm:VAMP-variational-principle}
that the $\{(s_{i},f_{i},g_{i})\}$ provided by the feature TCCA are
exactly the singular components of $\mathcal{K}_{\tau}^{\mathrm{proj}}$,
and the optimality of the estimation result is therefore invariant
for any choice of $r\ge1$. In addition, the sum over the $r$'th
power of all singular values of $\mathcal{K}_{\tau}^{\mathrm{proj}}$
is
\begin{equation}
\sum_{i}s_{i}^{r}=\left\Vert \mathbf{C}_{00}^{-\frac{1}{2}}\mathbf{C}_{01}\mathbf{C}_{11}^{-\frac{1}{2}}\right\Vert _{r}^{r}.\label{eq:sum-sir-matrix-norm}
\end{equation}

\subsection{An example of nonlinear TCCA\label{subsec:example-nonlinear-TCCA}}

Consider a stochastic system
\begin{equation}
x_{t+1}=\frac{1}{2}x_{t}+u_{t},
\end{equation}
where $u_{t}$ is Gaussian white noise with mean zero and variance
$1$. By setting
\begin{equation}
\rho_{0}(x)=\rho_{1}(x)=\mathcal{N}\left(x|0,\frac{4}{3}\right)
\end{equation}
to be the stationary distribution and basis functions
\begin{equation}
\boldsymbol{\chi}_{0}(x)=\boldsymbol{\chi}_{1}(x)=\left(1,\exp(-wx^{2})-\sqrt{\frac{3}{8w+3}},x\exp(-(1-w^{\frac{1}{10}})x^{2})\right)^{\top}
\end{equation}
with parameter $w\in[0.01,1]$, we can obtain
\begin{eqnarray}
\mathbf{C}_{00}=\mathbf{C}_{11} & = & \mathrm{diag}\left(1,\left(\frac{16}{3}w+1\right)^{-\frac{1}{2}}-\frac{3}{8w+3},4\sqrt{3}\left(-16w^{\frac{1}{10}}+19\right)^{-\frac{3}{2}}\right),\nonumber \\
\mathbf{C}_{01} & = & \mathrm{diag}\Bigg(1,\left(\frac{16}{3}w^{2}+\frac{16}{3}w+1\right)^{-\frac{1}{2}}-\frac{3}{8w+3},\nonumber \\
 &  & \quad\quad\quad2\sqrt{3}\left(16(1-w^{\frac{1}{10}})^{2}-16w^{\frac{1}{10}}+19\right)^{-\frac{3}{2}}\Bigg)
\end{eqnarray}
The maximal VAMP-$r$ score for a given $w$ can then be analytically
computed by
\begin{equation}
\mathcal{R}_{r}(w)=\mathrm{tr}\left[\left(\mathbf{C}_{00}(w)^{-1}\mathbf{C}_{01}(w)\right)^{r}\right]
\end{equation}
according to \eqref{eq:Rr-w}. We evaluate $\mathcal{R}_{r}(w)$ at
$9901$ equally spaced points of $w$ in the interval $[0.01,1]$
for $r=1,2$, and the maximal values of $\mathcal{R}_{1},\mathcal{R}_{2}$
are achieved at $w=0.3157$ and $w=0.7069$ respectively.

\section{Implementation of estimation algorithms}

\subsection{De-correlation of basis functions\label{sec:Proof-of-Proposition-linear-variational-bound}}

For convenience of notation, here we define
\begin{eqnarray}
\mathbf{X} & = & \left(\boldsymbol{\chi}_{0}(\mathbf{x}_{1}),\ldots,\boldsymbol{\chi}_{0}(\mathbf{x}_{T-\tau})\right)^{\top}\\
\mathbf{Y} & = & \left(\boldsymbol{\chi}_{1}(\mathbf{x}_{1+\tau}),\ldots,\boldsymbol{\chi}_{0}(\mathbf{x}_{T})\right)^{\top}.
\end{eqnarray}
In this paper, we utilize principal component analysis (PCA) to explicitly
reduce correlations between basis functions as follows: First, we
compute the empirical means of basis functions and the covariance
matrices of mean-centered basis functions:
\begin{eqnarray}
\boldsymbol{\pi}_{0} & = & \frac{1}{T-\tau}\mathbf{X}^{\top}\mathbf{1}\label{eq:pi0}\\
\boldsymbol{\pi}_{1} & = & \frac{1}{T-\tau}\mathbf{Y}^{\top}\mathbf{1}\label{eq:pi1}\\
\mathrm{COV}_{0} & = & \frac{1}{T-\tau}\mathbf{X}^{\top}\mathbf{X}-\boldsymbol{\pi}_{0}\boldsymbol{\pi}_{0}^{\top}\label{eq:cov0}\\
\mathrm{COV}_{1} & = & \frac{1}{T-\tau}\mathbf{Y}^{\top}\mathbf{Y}-\boldsymbol{\pi}_{1}\boldsymbol{\pi}_{1}^{\top}.\label{eq:cov1}
\end{eqnarray}
Next, perform the truncated eigen decomposition of the covariance
matrices as
\begin{eqnarray}
\mathrm{COV}_{0} & \approx & \mathbf{Q}_{0,d}^{\top}\mathbf{S}_{0,d}\mathbf{Q}_{0,d}\\
\mathrm{COV}_{1} & \approx & \mathbf{Q}_{1,d}^{\top}\mathbf{S}_{1,d}\mathbf{Q}_{1,d},
\end{eqnarray}
where the diagonal of matrices $\mathbf{S}_{0,d},\mathbf{S}_{1,d}$
contain all positive eigenvalues that are larger than $\epsilon_{0}$
and absolute values of all negative eigenvalues ($\epsilon_{0}=10^{-10}$
in our applications). Last, the new basis functions are given by
\begin{equation}
\boldsymbol{\chi}_{0}^{\mathrm{new}}=\left[\begin{array}{c}
\mathbf{Q}_{0,d}^{\top}\mathbf{S}_{0,d}^{\frac{1}{2}}\left(\boldsymbol{\chi}_{0}-\boldsymbol{\pi}_{0}\right)\\
\mathbbm1
\end{array}\right],\quad\boldsymbol{\chi}_{1}^{\mathrm{new}}=\left[\begin{array}{c}
\mathbf{Q}_{1,d}^{\top}\mathbf{S}_{1,d}^{\frac{1}{2}}\left(\boldsymbol{\chi}_{1}-\boldsymbol{\pi}_{1}\right)\\
\mathbbm1
\end{array}\right]\label{eq:decorrelation}
\end{equation}
We denote the transformation \eqref{eq:decorrelation} by
\begin{equation}
\boldsymbol{\chi}_{0}^{\mathrm{new}},\boldsymbol{\chi}_{1}^{\mathrm{new}}=\mathrm{DC}\left[\boldsymbol{\chi}_{0},\boldsymbol{\chi}_{1}|\boldsymbol{\pi}_{0},\boldsymbol{\pi}_{1},\mathrm{COV}_{0},\mathrm{COV}_{1}\right]
\end{equation}
Then the feature TCCA algorithm with de-correlation of basis functions
can be summarized as:
\begin{enumerate}
\item Compute $\boldsymbol{\pi}_{0},\boldsymbol{\pi}_{1}$ and $\mathrm{COV}_{0},\mathrm{COV}_{1}$
by (\ref{eq:pi0}-\ref{eq:cov1}).
\item Let $\boldsymbol{\chi}_{0},\boldsymbol{\chi}_{1}:=\mathrm{DC}\left[\boldsymbol{\chi}_{0},\boldsymbol{\chi}_{1}|\boldsymbol{\pi}_{0},\boldsymbol{\pi}_{1},\mathrm{COV}_{0},\mathrm{COV}_{1}\right]$,
and recalculate $\mathbf{X}$ and $\mathbf{Y}$ according to the new
basis functions.
\item Compute covariance matrices $\mathbf{C}_{00},\mathbf{C}_{01},\mathbf{C}_{11}$
by
\begin{eqnarray*}
\mathbf{C}_{00} & = & \frac{1}{T-\tau}\mathbf{X}^{\top}\mathbf{X}\\
\mathbf{C}_{01} & = & \frac{1}{T-\tau}\mathbf{X}^{\top}\mathbf{Y}\\
\mathbf{C}_{11} & = & \frac{1}{T-\tau}\mathbf{Y}^{\top}\mathbf{Y}
\end{eqnarray*}
\item Perform the truncated SVD $\mathbf{C}_{00}^{-\frac{1}{2}}\mathbf{C}_{01}\mathbf{C}_{11}^{-\frac{1}{2}}=\mathbf{U}_{k}^{\prime}\hat{\boldsymbol{\Sigma}}_{k}\mathbf{V}_{k}^{\prime\top}$.
\item Output estimated singular components $\hat{\boldsymbol{\Sigma}}_{k}=\mathrm{diag}(\hat{\sigma}_{1},\ldots,\hat{\sigma}_{k})$,
$\mathbf{U}_{k}^{\top}\boldsymbol{\chi}_{0}=(\hat{\psi}_{1},\ldots,\hat{\psi}_{k})^{\top}$
and $\mathbf{V}_{k}^{\top}\boldsymbol{\chi}_{1}=(\hat{\phi}_{1},\ldots,\hat{\phi}_{k})^{\top}$
with $\mathbf{U}_{k}=\mathbf{C}_{00}^{-\frac{1}{2}}\mathbf{U}_{k}^{\prime}$
and $\mathbf{V}_{k}=\mathbf{C}_{11}^{-\frac{1}{2}}\mathbf{V}_{k}^{\prime}$.
\end{enumerate}
Notice that the estimated $\mathbf{C}_{00}$, $\mathbf{C}_{01}$ and
$\mathbf{C}_{11}$ in the above algorithm satisfy
\begin{eqnarray}
\left[\begin{array}{cc}
\mathbf{C}_{00} & \mathbf{C}_{01}\\
\mathbf{C}_{01}^{\top} & \mathbf{C}_{11}
\end{array}\right] & = & \frac{1}{T-\tau}\left[\begin{array}{cc}
\mathbf{X}^{\top}\mathbf{X} & \mathbf{X}^{\top}\mathbf{Y}\\
\mathbf{Y}^{\top}\mathbf{X} & \mathbf{Y}^{\top}\mathbf{Y}
\end{array}\right]\nonumber \\
 & = & \frac{1}{T-\tau}\left(\mathbf{X},\mathbf{Y}\right)^{\top}\left(\mathbf{X},\mathbf{Y}\right)\nonumber \\
 & \succeq & 0
\end{eqnarray}
where $\mathbf{C}\succeq0$ means $\mathbf{C}$ is a positive semi-definite
matrix. According to the Schur complement lemma, we have
\begin{eqnarray}
\mathbf{C}_{01}\mathbf{C}_{11}^{-1}\mathbf{C}_{01}^{\top} & \preceq & \mathbf{C}_{00}\nonumber \\
\Rightarrow\left(\mathbf{C}_{00}^{-\frac{1}{2}}\mathbf{C}_{01}\mathbf{C}_{11}^{-\frac{1}{2}}\right)\left(\mathbf{C}_{00}^{-\frac{1}{2}}\mathbf{C}_{01}\mathbf{C}_{11}^{-\frac{1}{2}}\right)^{\top} & \preceq & \mathbf{I}
\end{eqnarray}
where $\mathbf{I}$ denotes an identity matrix of appropriate size.
So the estimated $\sigma_{1}\le1$.

Furthermore, since $\mathbf{v}_{0}^{\top}\boldsymbol{\chi}_{0}=\mathbf{v}_{1}^{\top}\boldsymbol{\chi}_{1}=\mathbbm1$
for $\mathbf{v}_{0}=(0,\ldots,0,1)^{\top}$ and $\mathbf{v}_{1}=(0,\ldots,0,1)^{\top}$,

\begin{eqnarray}
\left(\mathbf{C}_{00}^{-\frac{1}{2}}\mathbf{C}_{01}\mathbf{C}_{11}^{-\frac{1}{2}}\right)\left(\mathbf{C}_{00}^{-\frac{1}{2}}\mathbf{C}_{01}\mathbf{C}_{11}^{-\frac{1}{2}}\right)^{\top}\mathbf{C}_{00}^{\frac{1}{2}}\mathbf{v}_{0} & = & \mathbf{C}_{00}^{\frac{1}{2}}\left(\mathbf{X}^{\top}\mathbf{X}\right)^{-1}\mathbf{X}^{\top}\mathbf{Y}\left(\mathbf{Y}^{\top}\mathbf{Y}\right)^{-1}\mathbf{Y}^{\top}\mathbf{X}\mathbf{v}_{0}\nonumber \\
 & = & \mathbf{C}_{00}^{\frac{1}{2}}\mathbf{X}^{+}\mathbf{Y}\mathbf{Y}^{+}\mathbf{1}\nonumber \\
 & = & \mathbf{C}_{00}^{\frac{1}{2}}\mathbf{v}_{0}
\end{eqnarray}
which implies that $1$ is the largest singular value of $\mathbf{C}_{00}^{-\frac{1}{2}}\mathbf{C}_{01}\mathbf{C}_{11}^{-\frac{1}{2}}$.

\subsection{Parameter optimization in nonlinear TCCA\label{subsec:Parameter-optimization-nonlinear-TCCA}}

The optimization problem
\begin{equation}
\max_{\mathbf{w}}\mathcal{R}_{r}(\mathbf{w})=\left\Vert \mathbf{C}_{00}\left(\mathbf{w}\right)^{-\frac{1}{2}}\mathbf{C}_{01}\left(\mathbf{w}\right)\mathbf{C}_{11}\left(\mathbf{w}\right)^{-\frac{1}{2}}\right\Vert _{r}^{r}
\end{equation}
can be solved by direct search as in our examples (see Appendix \ref{subsec:app-One-dimensional-system}).
But for a high-dimensional parameter vector $\mathbf{w}$, it is more
efficient to perform the optimization by the gradient descent method
in the form of
\begin{equation}
\mathbf{w}\leftarrow\mathbf{w}+\eta\frac{\partial\mathcal{R}_{r}(\mathbf{w})}{\partial\mathbf{w}},
\end{equation}
where $\eta$ is the step size. When $r=2$, the gradient of $\mathcal{R}_{r}$
with respect to an element $w_{i}$ in $\mathbf{w}$ can be written
as
\begin{eqnarray}
\frac{\partial\mathcal{R}_{r}}{\partial w_{i}} & = & \frac{2}{T-\tau}\mathrm{tr}\left[\mathbf{C}_{00}^{-1}\mathbf{C}_{01}\mathbf{C}_{11}^{-1}\left(\mathbf{Y}-\mathbf{C}_{01}^{\top}\mathbf{C}_{00}^{-1}\mathbf{X}\right)\left(\frac{\partial\mathbf{X}}{\partial w_{i}}\right)^{\top}\right]\nonumber \\
 &  & +\frac{2}{T-\tau}\mathrm{tr}\left[\mathbf{C}_{11}^{-1}\mathbf{C}_{01}^{\top}\mathbf{C}_{00}^{-1}\left(\mathbf{X}-\mathbf{C}_{01}\mathbf{C}_{11}^{-1}\mathbf{Y}\right)\left(\frac{\partial\mathbf{Y}}{\partial w_{i}}\right)^{\top}\right],
\end{eqnarray}
where $\mathbf{X},\mathbf{Y}$ have the same definitions as in Appendix
\ref{sec:Proof-of-Proposition-linear-variational-bound}. If the data
size is too large, we can approximate the gradient based on a random
subset of data in each iteration, and update $\mathbf{w}$ in a stochastic
gradient descent manner \cite{andrew2013deep,vampnet}.

Like feature TCCA, the nonlinear TCCA also suffers from the numerical
singularity when $\mathbf{C}_{00}$ or $\mathbf{C}_{11}$ is not full
rank. This problem can be addressed by the de-correlation of basis
functions when performing direct search. For the gradient descent
method (or stochastic gradient descent method), we can replace the
objective function $\mathcal{R}_{r}(\mathbf{w})$ by a regularized
one
\begin{equation}
\mathcal{R}_{r}(\mathbf{w};\epsilon)=\left\Vert \left(\mathbf{C}_{00}\left(\mathbf{w}\right)+\epsilon\mathbf{I}\right)^{-\frac{1}{2}}\mathbf{C}_{01}\left(\mathbf{w}\right)\left(\mathbf{C}_{11}\left(\mathbf{w}\right)+\epsilon\mathbf{I}\right)^{-\frac{1}{2}}\right\Vert _{r}^{r},
\end{equation}
where $\epsilon>0$ is a hyperparameter and can be selected by the
cross-validation.

\section{Relationship between VAMP and EDMD\label{sec:Relationship-between-VAMP-EDMD}}

The proof of \eqref{eq:chi-K-chi} is trivial. Here, we only show
that the eigenvalue problem of $\hat{\mathcal{K}}_{\tau}$ given by
the feature TCCA is equivalent to that of matrix $\mathbf{K}_{\chi}$
as
\begin{equation}
\hat{\mathcal{K}}_{\tau}g=\lambda g\Longleftrightarrow\mathbf{K}_{\chi}\mathbf{b}=\lambda\mathbf{b}\text{ with }g=\mathbf{b}^{\top}\boldsymbol{\chi}\label{eq:eigenvalue-problem}
\end{equation}
under the assumption that $\boldsymbol{\chi}_{0}=\boldsymbol{\chi}_{1}=\boldsymbol{\chi}$
and $\mathbf{C}_{00}$ is invertible, which is consistent with the
spectral approximation theory in EDMD. First, if $g$ and $\lambda$
satisfy $\mathcal{K}_{\tau}g=\lambda g$, there must exist vector
$\mathbf{b}$ so that $g=\mathbf{b}^{\top}\boldsymbol{\chi}$. Then
\begin{eqnarray}
\hat{\mathcal{K}}_{\tau}g & = & \lambda g\nonumber \\
\Rightarrow\mathbf{b}^{\top}\mathbf{K}_{\chi}^{\top}\boldsymbol{\chi} & = & \lambda\mathbf{b}^{\top}\boldsymbol{\chi}\nonumber \\
\Rightarrow\mathbf{b}^{\top}\mathbf{K}_{\chi}^{\top}\mathbf{C}_{00} & = & \lambda\mathbf{b}^{\top}\mathbf{C}_{00}\nonumber \\
\Rightarrow\mathbf{K}_{\chi}\mathbf{b} & = & \lambda\mathbf{b}.
\end{eqnarray}
Second, if $\mathbf{K}_{\chi}\mathbf{b}=\lambda\mathbf{b}$,
\begin{eqnarray}
\hat{\mathcal{K}}_{\tau}\mathbf{b}^{\top}\boldsymbol{\chi} & = & \mathbf{b}^{\top}\mathbf{K}_{\chi}^{\top}\boldsymbol{\chi}\nonumber \\
 & = & \lambda\mathbf{b}^{\top}\boldsymbol{\chi}.
\end{eqnarray}

\section{Analysis of the VAMP-E score \label{sec:Analysis-VAMP-E}}

\subsection{Proof of \eqref{eq:error}}

Here we define
\begin{eqnarray}
\mathbf{C}_{ff} & = & \left\langle \mathbf{f},\mathbf{f}^{\top}\right\rangle _{\rho_{0}}=\mathbf{U}^{\top}\mathbf{C}_{00}\mathbf{U},\\
\mathbf{C}_{gg} & = & \left\langle \mathbf{g},\mathbf{g}^{\top}\right\rangle _{\rho_{1}}=\mathbf{V}^{\top}\mathbf{C}_{11}\mathbf{V},\\
\mathbf{C}_{fg} & = & \left\langle \mathbf{f},\mathcal{K}_{\tau}\mathbf{g}^{\top}\right\rangle _{\rho_{1}}=\mathbf{U}^{\top}\mathbf{C}_{01}\mathbf{V}.
\end{eqnarray}
Considering $\{\phi_{i}\}$ is an orthonormal basis of $\mathcal{L}_{\rho_{1}}^{2}$,
we have
\begin{eqnarray}
\left\Vert \hat{\mathcal{K}}_{\tau}\right\Vert _{\mathrm{HS}}^{2} & = & \sum_{j}\left\langle \hat{\mathcal{K}}_{\tau}\phi_{j},\hat{\mathcal{K}}_{\tau}\phi_{j}\right\rangle _{\rho_{0}}\nonumber \\
 & = & \sum_{j}\left\langle \left\langle \phi_{j},\mathbf{g}^{\top}\right\rangle _{\rho_{1}}\mathbf{K}\mathbf{f},\mathbf{f}^{\top}\mathbf{K}\left\langle \mathbf{g},\phi_{j}\right\rangle _{\rho_{1}}\right\rangle _{\rho_{0}}\nonumber \\
 & = & \sum_{j}\left\langle \phi_{j},\mathbf{g}^{\top}\right\rangle _{\rho_{1}}\mathbf{K}\left\langle \mathbf{f},\mathbf{f}^{\top}\right\rangle _{\rho_{0}}\mathbf{K}\left\langle \mathbf{g},\phi_{j}\right\rangle _{\rho_{1}}\nonumber \\
 & = & \mathrm{tr}\left[\mathbf{K}\left\langle \mathbf{f},\mathbf{f}^{\top}\right\rangle _{\rho_{0}}\mathbf{K}\sum_{j}\left\langle \mathbf{g},\phi_{j}\right\rangle _{\rho_{1}}\left\langle \phi_{j},\mathbf{g}^{\top}\right\rangle _{\rho_{1}}\right]\nonumber \\
 & = & \mathrm{tr}\left[\mathbf{K}\left\langle \mathbf{f},\mathbf{f}^{\top}\right\rangle _{\rho_{0}}\mathbf{K}\left\langle \sum_{j}\left\langle \mathbf{g},\phi_{j}\right\rangle _{\rho_{1}}\phi_{j},\mathbf{g}^{\top}\right\rangle _{\rho_{1}}\right]\nonumber \\
 & = & \mathrm{tr}\left[\mathbf{K}\left\langle \mathbf{f},\mathbf{f}^{\top}\right\rangle _{\rho_{0}}\mathbf{K}\left\langle \mathbf{g},\mathbf{g}^{\top}\right\rangle _{\rho_{1}}\right]\nonumber \\
 & = & \mathrm{tr}\left[\mathbf{K}\mathbf{C}_{ff}\mathbf{K}\mathbf{C}_{gg}\right]
\end{eqnarray}
and
\begin{eqnarray}
\left\langle \hat{\mathcal{K}}_{\tau},\mathcal{K}_{\tau}\right\rangle _{\mathrm{HS}} & = & \sum_{j}\left\langle \hat{\mathcal{K}}_{\tau}\phi_{j},\mathcal{K}_{\tau}\phi_{j}\right\rangle _{\rho_{0}}\nonumber \\
 & = & \sum_{j}\left\langle \left\langle \phi_{j},\mathbf{g}^{\top}\right\rangle _{\rho_{1}}\mathbf{S}\mathbf{f},\sigma_{j}\psi_{j}\right\rangle _{\rho_{0}}\nonumber \\
 & = & \sum_{j}\sigma_{j}\left\langle \phi_{j},\mathbf{g}^{\top}\right\rangle _{\rho_{1}}\mathbf{S}\left\langle \mathbf{f},\psi_{j}\right\rangle _{\rho_{0}}\nonumber \\
 & = & \mathrm{tr}\left[\mathbf{K}\sum_{j}\sigma_{j}\left\langle \mathbf{f},\psi_{j}\right\rangle _{\rho_{0}}\left\langle \phi_{j},\mathbf{g}^{\top}\right\rangle _{\rho_{1}}\right]\nonumber \\
 & = & \mathrm{tr}\left[\mathbf{K}\left\langle \mathbf{f},\sum_{j}\sigma_{j}\psi_{j}\left\langle \phi_{j},\mathbf{g}^{\top}\right\rangle _{\rho_{1}}\right\rangle _{\rho_{0}}\right]\nonumber \\
 & = & \mathrm{tr}\left[\mathbf{K}\left\langle \mathbf{f},\mathcal{K}_{\tau}\mathbf{g}^{\top}\right\rangle _{\rho_{0}}\right]\nonumber \\
 & = & \mathrm{tr}\left[\mathbf{K}\mathbf{C}_{fg}\right],
\end{eqnarray}
where $\left\langle \cdot,\cdot\right\rangle _{\mathrm{HS}}$ denotes
the Hilbert-Schmidt inner product of operators. Then, according to
the definition of Hilbert-Schmidt norm,
\begin{eqnarray}
\left\Vert \hat{\mathcal{K}}_{\tau}-\mathcal{K}_{\tau}\right\Vert _{\mathrm{HS}}^{2} & = & \left\Vert \hat{\mathcal{K}}_{\tau}\right\Vert _{\mathrm{HS}}^{2}-2\sum_{j}\left\langle \hat{\mathcal{K}}_{\tau},\mathcal{K}_{\tau}\right\rangle _{\mathrm{HS}}+\left\Vert \mathcal{K}_{\tau}\right\Vert _{\mathrm{HS}}^{2}\nonumber \\
 & = & \mathrm{tr}\left[\mathbf{K}\mathbf{C}_{ff}\mathbf{K}\mathbf{C}_{gg}-2\mathbf{K}\mathbf{C}_{fg}\right]+\left\Vert \mathcal{K}_{\tau}\right\Vert _{\mathrm{HS}}^{2}
\end{eqnarray}

\subsection{Relationship between VAMP-2 and VAMP-E}

We first show that the feature TCCA algorithm maximizes VAMP-E. Notice
that
\begin{eqnarray}
\mathcal{R}_{E}(\mathbf{K},\mathbf{U},\mathbf{V}) & = & \mathrm{tr}\left[2\left(\mathbf{C}_{00}^{\frac{1}{2}}\mathbf{U}\mathbf{K}\mathbf{V}^{\top}\mathbf{C}_{11}^{\frac{1}{2}}\right)^{\top}\left(\mathbf{C}_{00}^{-\frac{1}{2}}\mathbf{C}_{01}\mathbf{C}_{11}^{-\frac{1}{2}}\right)-\left(\mathbf{C}_{00}^{\frac{1}{2}}\mathbf{U}\mathbf{K}\mathbf{V}^{\top}\mathbf{C}_{11}^{\frac{1}{2}}\right)^{\top}\left(\mathbf{C}_{00}^{\frac{1}{2}}\mathbf{U}\mathbf{K}\mathbf{V}^{\top}\mathbf{C}_{11}^{\frac{1}{2}}\right)\right]\nonumber \\
 & = & -\left\Vert \mathbf{C}_{00}^{\frac{1}{2}}\mathbf{U}\mathbf{K}\mathbf{V}^{\top}\mathbf{C}_{11}^{\frac{1}{2}}-\mathbf{C}_{00}^{-\frac{1}{2}}\mathbf{C}_{01}\mathbf{C}_{11}^{-\frac{1}{2}}\right\Vert _{F}^{2}+\left\Vert \mathbf{C}_{00}^{-\frac{1}{2}}\mathbf{C}_{01}\mathbf{C}_{11}^{-\frac{1}{2}}\right\Vert _{F}^{2}\nonumber \\
 & = & -\left\Vert \mathbf{U}^{\prime}\mathbf{K}\mathbf{V}^{\prime\top}-\mathbf{C}_{00}^{-\frac{1}{2}}\mathbf{C}_{01}\mathbf{C}_{11}^{-\frac{1}{2}}\right\Vert _{F}^{2}+\left\Vert \mathbf{C}_{00}^{-\frac{1}{2}}\mathbf{C}_{01}\mathbf{C}_{11}^{-\frac{1}{2}}\right\Vert _{F}^{2},\label{eq:VAMP-E-decomposition}
\end{eqnarray}
where $\left\Vert \cdot\right\Vert _{F}$ denotes the Frobenius norm
and $\mathbf{U}^{\prime}=\mathbf{C}_{00}^{\frac{1}{2}}\mathbf{U}$,
$\mathbf{V}^{\prime}=\mathbf{C}_{11}^{\frac{1}{2}}\mathbf{V}$. It
can be seen that the feature TCCA algorithm maximizes the first term
on the right-hand side of \eqref{eq:VAMP-E-decomposition} and therefore
maximizes VAMP-E.

For the optimal model generated by the nonlinear TCCA, the first term
on the right-hand side of \eqref{eq:VAMP-E-decomposition} is equal
to zero and the second term is maximized as a function of $\mathbf{w}$.
Thus, the nonlinear TCCA also maximizes VAMP-E.

In addition, for $\mathbf{K},\mathbf{U},\mathbf{V}$ provided by both
feature TCCA and nonlinear TCCA,
\begin{eqnarray}
\mathcal{R}_{E}(\mathbf{K},\mathbf{U},\mathbf{V}) & = & -\left\Vert \mathbf{U}^{\prime}\mathbf{K}\mathbf{V}^{\prime\top}-\mathbf{C}_{00}^{-\frac{1}{2}}\mathbf{C}_{01}\mathbf{C}_{11}^{-\frac{1}{2}}\right\Vert _{F}^{2}+\left\Vert \mathbf{C}_{00}^{-\frac{1}{2}}\mathbf{C}_{01}\mathbf{C}_{11}^{-\frac{1}{2}}\right\Vert _{F}^{2}\nonumber \\
 & = & -\sum_{i=k+1}^{\min\{m,n\}}K_{ii}^{2}+\sum_{i=1}^{\min\{m,n\}}K_{ii}^{2}\nonumber \\
 & = & \sum_{i=1}^{k}K_{ii}^{2}\nonumber \\
 & = & \mathcal{R}_{2}(\mathbf{U},\mathbf{V}).
\end{eqnarray}

\section{Subspace variational principle\label{sec:Singular-space-based-vamp}}

The variational principle proposed in Section \ref{subsec:Variational-principle}
can be further extended to singular subspaces of the Koopman operator
as follows:
\begin{equation}
\sum_{i=1}^{k}\sigma_{i}^{r}\ge\mathcal{R}_{r}^{\mathrm{space}}\left[\mathbf{f},\mathbf{g}\right]=\left\Vert \mathbf{C}_{ff}^{-\frac{1}{2}}\mathbf{C}_{fg}\mathbf{C}_{gg}^{-\frac{1}{2}}\right\Vert _{r}^{r}\label{eq:variational-space}
\end{equation}
for $r\ge1$, and the equality holds if $\mathrm{span}\{\psi_{1},\ldots,\psi_{k}\}=\mathrm{span}\{f_{1},\ldots,f_{k}\}$
and $\mathrm{span}\{\phi_{1},\ldots,\phi_{k}\}=\mathrm{span}\{g_{1},\ldots,g_{k}\}$,
where $\mathbf{C}_{ff}=\left\langle \mathbf{f},\mathbf{f}^{\top}\right\rangle _{\rho_{0}}$,
$\mathbf{C}_{fg}=\left\langle \mathbf{f},\mathcal{K}_{\tau}\mathbf{g}^{\top}\right\rangle _{\rho_{0}}$
and $\mathbf{C}_{gg}=\left\langle \mathbf{g},\mathbf{g}^{\top}\right\rangle _{\rho_{1}}$.
This statement can be proven by implementing the feature TCCA algorithm
with feature functions $\mathbf{f}$ and $\mathbf{g}$.

The $\mathcal{R}_{r}^{\mathrm{space}}\left[\mathbf{f},\mathbf{g}\right]$
is a relaxation of VAMP-$r$, which measures the consistency between
the subspaces spanned by $\mathbf{f},\mathbf{g}$ and the dominant
singular spaces, and we call it the subspace VAMP-$r$ score. $\mathcal{R}_{r}^{\mathrm{space}}\left[\mathbf{f},\mathbf{g}\right]$
is invariant with respect to the invertible linear transformations
of $\mathbf{f}$ and $\mathbf{g}$, i.e., $\mathcal{R}_{r}^{\mathrm{space}}\left[\mathbf{f},\mathbf{g}\right]=\mathcal{R}_{r}^{\mathrm{space}}\left[\mathbf{A}_{f}\mathbf{f},\mathbf{A}_{g}\mathbf{g}\right]$
for any invertible matrices $\mathbf{A}_{f},\mathbf{A}_{g}$.

In the cross-validation for feature TCCA, we can utilize $\mathcal{R}_{r}^{\mathrm{space}}$
to calculate the validation score by
\begin{eqnarray}
\mathrm{CV}\left(\mathbf{K},\mathbf{U},\mathbf{V}|\mathcal{D}_{\mathrm{test}}\right) & = & \mathcal{R}_{r}^{\mathrm{space}}\left(\mathbf{U},\mathbf{V}|\mathcal{D}_{\mathrm{test}}\right)\nonumber \\
 & = & \mathcal{R}_{r}^{\mathrm{space}}\left[\mathbf{U}^{\top}\boldsymbol{\chi}_{0},\mathbf{V}^{\top}\boldsymbol{\chi}_{1}|\mathcal{D}_{\mathrm{test}}\right]\nonumber \\
 & = & \left\Vert \left(\mathbf{U}^{\top}\mathbf{C}_{00}^{\mathrm{test}}\mathbf{U}\right)^{-\frac{1}{2}}\left(\mathbf{U}^{\top}\mathbf{C}_{01}^{\mathrm{test}}\mathbf{V}\right)\left(\mathbf{V}^{\top}\mathbf{C}_{11}^{\mathrm{test}}\mathbf{V}\right)^{-\frac{1}{2}}\right\Vert _{r}^{r}.
\end{eqnarray}

We now analyze the difficulties of applying $\mathcal{R}_{r}^{\mathrm{space}}$
to the cross-validation. First, for given basis functions $\boldsymbol{\chi}_{0},\boldsymbol{\chi}_{1}$,
$\mathcal{R}_{r}^{\mathrm{space}}\left(\mathbf{U},\mathbf{V}|\mathcal{D}_{\mathrm{test}}\right)$
is monotonically increasing with respect to $k$ and
\begin{equation}
\mathcal{R}_{r}^{\mathrm{space}}\left(\mathbf{U}_{k},\mathbf{V}_{k}|\mathcal{D}_{\mathrm{test}}\right)=\left\Vert \left(\mathbf{C}_{00}^{\mathrm{test}}\right)^{-\frac{1}{2}}\mathbf{C}_{01}^{\mathrm{test}}\left(\mathbf{C}_{11}^{\mathrm{test}}\right)^{-\frac{1}{2}}\right\Vert _{r}^{r}
\end{equation}
is independent of the estimated singular components if $k=\max\{\mathrm{dim}(\boldsymbol{\chi}_{0}),\mathrm{dim}(\boldsymbol{\chi}_{1})\}$.
Therefore, $k$ is a new hyper-parameter that cannot be determined
by the cross-validation. Second, for training set, $\mathbf{U}_{k}^{\top}\mathbf{C}_{00}^{\mathrm{train}}\mathbf{U}_{k}=\mathbf{V}_{k}^{\top}\mathbf{C}_{11}^{\mathrm{train}}\mathbf{V}_{k}=\mathbf{I}$.
But for test set, $\mathbf{U}_{k}^{\top}\mathbf{C}_{00}^{\mathrm{test}}\mathbf{U}_{k}$
and $\mathbf{V}_{k}^{\top}\mathbf{C}_{11}^{\mathrm{test}}\mathbf{V}_{k}$
are possibly singular and the validation score cannot be reliably
computed.

\section{Computation of $\hat{\mathcal{K}}_{\tau}^{n}$\label{sec:Calculation-Ktau-n}}

The approximate Koopman operator in the form of \eqref{eq:finite-rank-approximation}
can also be written as
\begin{equation}
\hat{\mathcal{K}}_{\tau}g=\left\langle g,\mathbf{g}^{\top}\right\rangle _{\rho_{1}}\mathbf{K}\mathbf{f}.
\end{equation}
Hence,

\begin{equation}
\hat{\mathcal{K}}_{\tau}^{n}g=\left\langle g,\mathbf{g}^{\top}\right\rangle _{\rho_{1}}\mathbf{K}\left(\mathbf{R}^{n-1}\right)^{\top}\mathbf{f},
\end{equation}
and we have
\begin{equation}
\left\langle f,\hat{\mathcal{K}}_{\tau}^{n}g\right\rangle _{\rho_{0}(n\tau)}=\left\langle f,\mathbf{f}^{\top}\right\rangle _{\rho_{0}(n\tau)}\mathbf{R}^{n-1}\mathbf{K}\left\langle \mathbf{g},g\right\rangle _{\rho_{1}}\label{eq:cov-fg-app}
\end{equation}
and

\begin{eqnarray}
\hat{p}_{n\tau}(\mathbf{x},\mathbf{y}) & = & \hat{\mathcal{K}}_{\tau}^{n}\delta_{\mathbf{y}}(\mathbf{x})\nonumber \\
 & = & \mathbf{f}(\mathbf{x})^{\top}\mathbf{R}^{n-1}\mathbf{K}\mathbf{g}(\mathbf{y})\rho_{1}(\mathbf{y}),
\end{eqnarray}
where
\begin{equation}
\mathbf{R}=\mathbf{K}\left\langle \mathbf{g},\mathbf{f}^{\top}\right\rangle _{\rho_{1}}.
\end{equation}

Notice that substituting $\mathbf{f}=\mathbf{U}^{\top}\boldsymbol{\chi}_{0},\mathbf{g}=\mathbf{V}^{\top}\boldsymbol{\chi}_{1}$
into \eqref{eq:cov-fg-app} yields \eqref{eq:cov-pred}.

\section{Details of numerical examples}

\subsection{One-dimensional system\label{subsec:app-One-dimensional-system}}

For convenience of analysis and computation, we partition the state
space $[-20,20]$ into $2000$ bins $S_{1},\ldots,S_{2000}$ uniformly,
and discretize the one-dimensional dynamical system described in Example
\ref{exa:one-dimensional-example-svd} as
\begin{equation}
\mathbb{P}(x_{t+1}\in S_{j}|x_{t}\in S_{i})\propto\mathcal{N}\left(s_{j}|\frac{s_{i}}{2}+\frac{7s_{i}}{1+0.12s_{i}^{2}}+6\cos s_{i},10\right),\label{eq:one-dimensional-discretization}
\end{equation}
where $s_{i}$ is the center of the bin $S_{i}$, and the local distribution
of $x_{t}$ within any bin is always uniform distribution. All numerical
computations and simulations in Examples \ref{exa:one-dimensional-example-svd},
\ref{exa:one-dimensional-example-estimation} and \ref{exa:one-dimensional-example-cv}
are based on \eqref{eq:one-dimensional-discretization}, and the initial
state $x_{0}$ is distributed according to the stationary distribution
$\rho_{0}=\rho_{1}=\mu$.

In Example \ref{exa:one-dimensional-example-svd}, the the stationary
distribution and singular components of the Koopman operator are analytically
computed by the feature TCCA with basis functions $\chi_{0,i}(x)=\chi_{1,i}(x)=1_{x\in S_{i}}$
as follows:
\begin{enumerate}
\item Compute the transition matrix $\mathbf{P}=[P_{ij}]=[\mathbb{P}(x_{t+1}\in S_{j}|x_{t}\in S_{i})]$
and the stationary vector $\boldsymbol{\pi}=[\pi_{i}]$ satisfying
\[
\boldsymbol{\pi}^{\top}\mathbf{P}=\boldsymbol{\pi}^{\top},\quad\sum_{i}\pi_{i}=1.
\]
\item Compute covariance matrices $\mathbf{C}_{00}=\mathbf{C}_{11}=\mathrm{diag}(\boldsymbol{\pi})$
and $\mathbf{C}_{01}=\mathrm{diag}(\boldsymbol{\pi})\mathbf{P}$.
\item Perform the SVD
\[
\bar{\mathbf{K}}=\mathbf{C}_{00}^{-\frac{1}{2}}\mathbf{C}_{01}\mathbf{C}_{11}^{-\frac{1}{2}}=\mathbf{U}^{\prime}\mathbf{K}\mathbf{V}^{\prime\top}
\]
with $\mathbf{K}=\mathrm{diag}(\sigma_{1},\ldots,\sigma_{2000})$
and $\sigma_{1}\ge\sigma_{2}\ge\ldots\ge\sigma_{2000}$.
\item Compute $\mathbf{U}=[U_{ij}]=\mathbf{C}_{00}^{-\frac{1}{2}}\mathbf{U}^{\prime}$
and $\mathbf{V}=[V_{ij}]=\mathbf{C}_{11}^{-\frac{1}{2}}\mathbf{V}^{\prime}$.
\item Output the stationary distribution $\mu(x)=\sum_{i}50\pi_{i}\cdot1_{x\in S_{i}}$
and singular components $(\sigma_{i},\psi_{i}(x),\phi_{i}(x))=(\sigma_{i},\sum_{j}U_{ji}\cdot1_{x\in S_{j}},\sum_{j}V_{ji}\cdot1_{x\in S_{j}})$.
\end{enumerate}
The transition density of the projected Koopman operator $\hat{\mathcal{K}}_{\tau}=\sum_{i=1}^{k}\sigma_{i}\left\langle \cdot,\phi_{i}\right\rangle _{\rho_{1}}\psi_{i}$
is obtained by
\begin{eqnarray}
\hat{p}_{\tau}(x,y) & = & \hat{\mathcal{K}}_{\tau}\delta_{y}(x)\nonumber \\
 & = & \sum_{i=1}^{k}\sigma_{i}\psi_{i}(x)\phi_{i}(y)\mu(y)
\end{eqnarray}
 (see Appendix \ref{subsec:ptau-and-Koopman}) and the corresponding
approximate transition matrix is
\begin{equation}
\hat{\mathbf{P}}=\mathbf{U}_{k}^{\top}\mathbf{K}_{k}\mathbf{V}_{k}\mathrm{diag}(\boldsymbol{\pi}),
\end{equation}
where $\mathbf{U}_{k},\mathbf{V}_{k}$ consist of the first $k$ columns
of $\mathbf{U},\mathbf{V}$, and $\mathbf{K}_{k}=\mathrm{diag}(\sigma_{1},\ldots,\sigma_{k})$.
Then the relative error of $\hat{\mathcal{K}}_{\tau}$ in Fig.~\ref{fig:one-dimensional-svd}e
can be calculated by
\begin{equation}
\frac{\Vert\hat{\mathcal{K}}_{\tau}-\mathcal{K}_{\tau}\Vert_{\mathrm{HS}}}{\Vert\mathcal{K}_{\tau}\Vert_{\mathrm{HS}}}=\frac{\sqrt{\sum_{i=k+1}^{2000}\sigma_{i}^{2}}}{\sqrt{\sum_{i=1}^{2000}\sigma_{i}^{2}}},
\end{equation}
the long-time transition density in Fig.~\ref{fig:one-dimensional-evolution}
is given by
\begin{equation}
\hat{p}_{n\tau}(x,y)=50\sum_{j}\left[\hat{\mathbf{P}}^{n}\right]_{ij}\cdot1_{y\in S_{j}},
\end{equation}
and the cumulative error of $\hat{p}_{n\tau}(x,y)$ is
\begin{eqnarray}
\mathrm{error} & = & \sum_{n=1}^{256}\int\mu(y)^{-1}\left(\hat{p}_{n\tau}(x,y)-p_{n\tau}(x,y)\right)^{2}\mathrm{d}y\nonumber \\
 & = & \sum_{n=1}^{256}\sum_{j=1}^{2000}\pi_{j}^{-1}\left(\left[\hat{\mathbf{P}}^{n}\right]_{ij}-\left[\mathbf{P}^{n}\right]_{ij}\right)^{2}
\end{eqnarray}
for $x\in S_{i}$.

In Examples \ref{exa:one-dimensional-example-estimation} and \ref{exa:one-dimensional-example-cv},
the smoothing parameter $w$ are optimized by the golden-section search
algorithm \cite{press2007numerical} as follows for nonlinear TCCA:
\begin{enumerate}
\item Let $a=-6$, $b=6$, $c=0.618a+0.382b$, $d=0.382a+0.618b$.
\item Compute $\mathcal{R}_{2}(\exp a)$, $\mathcal{R}_{2}(\exp b)$, $\mathcal{R}_{2}(\exp c)$
and $\mathcal{R}_{2}(\exp d)$, where $\mathcal{R}_{2}(w)=\left\Vert \mathbf{C}_{00}\left(w\right)^{-\frac{1}{2}}\mathbf{C}_{01}\left(w\right)\mathbf{C}_{11}\left(w\right)^{-\frac{1}{2}}\right\Vert _{F}^{2}$
and $\Vert\cdot\Vert_{F}$ denotes the Frobenius norm.
\item If $\max\{\mathcal{R}_{2}(\exp a),\mathcal{R}_{2}(\exp b),\mathcal{R}_{2}(\exp c)\}>\max\{\mathcal{R}_{2}(\exp b),\mathcal{R}_{2}(\exp c),\mathcal{R}_{2}(\exp d)\}$,
let $(a,b,c,d):=(a,d,0.618a+0.382d,c)$. Otherwise, let $(a,b,c,d):=(c,b,d,0.618b+0.382c)$.
\item If $|a-b|<10^{-3}$, output $\log w\in\{a,b,c,d\}$ with the largest
value of $\mathcal{R}_{2}(w)$. Otherwise, go back to Step 2.
\end{enumerate}
Furthermore, $w$ is computed in the same way when perform nonlinear
TCCA in Sections \ref{subsec:Double-gyre-system} and \ref{subsec:Stochastic-Lorenz-system}.

\subsection{Double-gyre system\label{subsec:app-Double-gyre-system}}

For the double-gyre system in Section \ref{subsec:Double-gyre-system},
we first perform the temporal discretization by the Euler\textendash Maruyama
scheme as
\begin{eqnarray}
\mathbb{P}(x_{t+\Delta}|\mathbf{x}_{t}) & = & \mathcal{N}(x_{t+\Delta}|x_{t}-\pi A\sin(\pi x_{t})\cos(\pi y_{t})\Delta,\epsilon^{2}(x_{t}/4+1)),\nonumber \\
\mathbb{P}(y_{t+\Delta}|\mathbf{x}_{t}) & = & \mathcal{N}(y_{t+\Delta}|y_{t}+\pi A\cos(\pi x_{t})\sin(\pi y_{t})\Delta,\epsilon^{2}),\label{eq:double-gyre-1}
\end{eqnarray}
where $\mathbf{x}_{t}=(x_{t},y_{t})^{\top}$ and $\Delta=0.02$ is
the step size. Then perform the spatial discretization as
\begin{eqnarray}
\mathbb{P}(\mathbf{x}_{t+\Delta}\in S_{j}|\mathbf{x}_{t}\in S_{i}) & \propto & \mathcal{N}(s_{j,x}|s_{i,x}-\pi A\sin(\pi s_{i,x})\cos(\pi s_{i,y})\Delta,\epsilon^{2}(s_{i,x}/4+1))\nonumber \\
 &  & \cdot\mathcal{N}(s_{j,y}|s_{i,y}+\pi A\cos(\pi s_{i,x})\sin(\pi s_{i,y})\Delta,\epsilon^{2}).\label{eq:double-gyre-discretization}
\end{eqnarray}
Here $S_{1},\ldots,S_{1250}$ are $50\times25$ bins which form a
uniform partition of the state space $[0,2]\times[0,1]$ and $(s_{i,x},s_{i,y})$
represents the center of $S_{i}$. Simulation data and the ``true''
singular components are all computed by using \eqref{eq:double-gyre-discretization}
with the initial distribution of $(x_{0},y_{0})$ being the stationary
one.

In Fig.~\ref{fig:double-gyre-evolution}, the transition density
of lag time $n\tau$ is computed from the estimated singular components
$(\mathbf{K},\mathbf{U}^{\top}\boldsymbol{\chi}_{0},\mathbf{V}^{\top}\boldsymbol{\chi}_{1})$
as
\begin{equation}
\hat{p}_{n\tau}(\mathbf{x},\mathbf{y})=625\sum_{j}\left[\hat{\mathbf{P}}^{n}\right]_{ij}\cdot1_{\mathbf{y}\in S_{j}},\quad\text{for }x\in S_{i}
\end{equation}
where
\begin{equation}
\hat{\mathbf{P}}=\mathbf{U}^{\top}\mathbf{K}\mathbf{V}\mathrm{diag}(\boldsymbol{\rho}_{1})
\end{equation}
is the approximate transition matrix, and $\boldsymbol{\rho}_{1}=[\boldsymbol{\rho}_{1i}]$
with
\begin{equation}
\boldsymbol{\rho}_{1i}=\frac{1}{T-\tau}\sum_{t=1}^{T-\tau}1_{\mathbf{x}_{t+\tau}\in S_{i}}.
\end{equation}

\clearpage\bibliographystyle{plain}
\addcontentsline{toc}{section}{\refname}\bibliography{all,hwu,own,new}

\begin{thebibliography}{10}

\bibitem{andrew2013deep}
Galen Andrew, Raman Arora, Jeff Bilmes, and Karen Livescu.
\newblock Deep canonical correlation analysis.
\newblock In {\em International Conference on Machine Learning}, pages
  1247--1255, 2013.

\bibitem{ArlotCelisse_StatSurv10_CVReview}
S.~Arlot and A.~Celisse.
\newblock A survey of cross-validation procedures for model selection.
\newblock {\em Stat. Surv.}, 4:40--79, 2010.

\bibitem{bollt2013applied}
Erik~M Bollt and Naratip Santitissadeekorn.
\newblock {\em Applied and Computational Measurable Dynamics}.
\newblock SIAM, 2013.

\bibitem{BoninsegnaEtAl_JCTC15_VariationalDM}
L.~Boninsegna, G.~Gobbo, F.~No{\'e}, and C.~Clementi.
\newblock Investigating molecular kinetics by variationally optimized diffusion
  maps.
\newblock {\em J. Chem. Theory Comput.}, 11:5947--5960, 2015.

\bibitem{BowmanPandeNoe_MSMBook}
G.~R. Bowman, V.~S. Pande, and F.~No{\'e}, editors.
\newblock {\em An Introduction to Markov State Models and Their Application to
  Long Timescale Molecular Simulation.}, volume 797 of {\em Advances in
  Experimental Medicine and Biology}.
\newblock Springer Heidelberg, 2014.

\bibitem{brunton2016koopman}
Steven~L Brunton, Bingni~W Brunton, Joshua~L Proctor, and J~Nathan Kutz.
\newblock Koopman invariant subspaces and finite linear representations of
  nonlinear dynamical systems for control.
\newblock {\em PloS one}, 11(2):e0150171, 2016.

\bibitem{brunton2016discovering}
Steven~L Brunton, Joshua~L Proctor, and J~Nathan Kutz.
\newblock Discovering governing equations from data by sparse identification of
  nonlinear dynamical systems.
\newblock {\em Proceedings of the National Academy of Sciences},
  113(15):3932--3937, 2016.

\bibitem{chekroun2011stochastic}
Micka{\"e}l~D Chekroun, Eric Simonnet, and Michael Ghil.
\newblock Stochastic climate dynamics: Random attractors and time-dependent
  invariant measures.
\newblock {\em Physica D: Nonlinear Phenomena}, 240(21):1685--1700, 2011.

\bibitem{ChoderaNoe_COSB14_MSMs}
J.~D. Chodera and F~No{\'e}.
\newblock Markov state models of biomolecular conformational dynamics.
\newblock {\em Curr. Opin. Struc. Biol.}, 25:135--144, 2014.

\bibitem{conrad2016finding}
Natasa~Djurdjevac Conrad, Marcus Weber, and Christof Sch{\"u}tte.
\newblock Finding dominant structures of nonreversible markov processes.
\newblock {\em Multiscale Modeling \& Simulation}, 14(4):1319--1340, 2016.

\bibitem{dellnitz2001algorithms}
Michael Dellnitz, Gary Froyland, and Oliver Junge.
\newblock The algorithms behind gaio - set oriented numerical methods for
  dynamical systems.
\newblock In {\em Ergodic theory, analysis, and efficient simulation of
  dynamical systems}, pages 145--174. Springer, 2001.

\bibitem{DeuflhardWeber_LAA05_PCCA+}
P.~Deuflhard and M.~Weber.
\newblock Robust perron cluster analysis in conformation dynamics.
\newblock In M.~Dellnitz, S.~Kirkland, M.~Neumann, and C.~Sch{\"u}tte, editors,
  {\em Linear Algebra Appl.}, volume 398C, pages 161--184. Elsevier, New York,
  2005.

\bibitem{friedman2001elements}
Jerome Friedman, Trevor Hastie, and Robert Tibshirani.
\newblock {\em The elements of statistical learning}.
\newblock Springer, New York, 2001.

\bibitem{froyland2013analytic}
Gary Froyland.
\newblock An analytic framework for identifying finite-time coherent sets in
  time-dependent dynamical systems.
\newblock {\em Physica D: Nonlinear Phenomena}, 250:1--19, 2013.

\bibitem{froyland2016optimal}
Gary Froyland, Cecilia Gonz{\'a}lez-Tokman, and Thomas~M Watson.
\newblock Optimal mixing enhancement by local perturbation.
\newblock {\em SIAM Review}, 58(3):494--513, 2016.

\bibitem{froyland2014computational}
Gary Froyland, Georg~A Gottwald, and Andy Hammerlindl.
\newblock A computational method to extract macroscopic variables and their
  dynamics in multiscale systems.
\newblock {\em SIAM Journal on Applied Dynamical Systems}, 13(4):1816--1846,
  2014.

\bibitem{froyland2009almost}
Gary Froyland and Kathrin Padberg.
\newblock Almost-invariant sets and invariant manifolds -- connecting
  probabilistic and geometric descriptions of coherent structures in flows.
\newblock {\em Physica D: Nonlinear Phenomena}, 238(16):1507--1523, 2009.

\bibitem{froyland2014almost}
Gary Froyland and Kathrin Padberg-Gehle.
\newblock Almost-invariant and finite-time coherent sets: directionality,
  duration, and diffusion.
\newblock In {\em Ergodic Theory, Open Dynamics, and Coherent Structures},
  pages 171--216. Springer, 2014.

\bibitem{hardoon2004canonical}
David~R Hardoon, Sandor Szedmak, and John Shawe-Taylor.
\newblock Canonical correlation analysis: An overview with application to
  learning methods.
\newblock {\em Neural Computation}, 16(12):2639--2664, 2004.

\bibitem{harmeling2003kernel}
Stefan Harmeling, Andreas Ziehe, Motoaki Kawanabe, and Klaus-Robert M{\"u}ller.
\newblock Kernel-based nonlinear blind source separation.
\newblock {\em Neural Computation}, 15(5):1089--1124, 2003.

\bibitem{hsing2015theoretical}
Tailen Hsing and Randall Eubank.
\newblock {\em Theoretical foundations of functional data analysis, with an
  introduction to linear operators}.
\newblock John Wiley \& Sons, 2015.

\bibitem{Klus_Arxiv16_Tensor}
Stefan Klus, Patrick Gel{\ss}, Sebastian Peitz, and Christof Sch{\"u}tte.
\newblock Tensor-based dynamic mode decomposition.
\newblock {\em Nonlinearity}, 31(7):3359, 2018.

\bibitem{KlusKoltaiSchuette_ApproximationKoopman}
Stefan Klus, P{\'e}ter Koltai, and Christof Sch{\"u}tte.
\newblock On the numerical approximation of the perron-frobenius and koopman
  operator.
\newblock {\em arXiv:1512.05997}, 2015.

\bibitem{KlusSchuette_Arxiv15_Tensor}
Stefan Klus and Christof Sch{\"u}tte.
\newblock Towards tensor-based methods for the numerical approximation of the
  perron-frobenius and koopman operator.
\newblock {\em arXiv:1512.06527}, 2015.

\bibitem{koltai2018optimal}
P.~Koltai, H.~Wu, F.~Noe, and C.~Sch{\"u}tte.
\newblock Optimal data-driven estimation of generalized markov state models for
  non-equilibrium dynamics.
\newblock {\em Computation}, 6(1):22, 2018.

\bibitem{konrad2001markov}
Almudena Konrad, Ben~Y Zhao, Anthony~D Joseph, and Reiner Ludwig.
\newblock A markov-based channel model algorithm for wireless networks.
\newblock In {\em Proceedings of the 4th ACM international workshop on
  Modeling, analysis and simulation of wireless and mobile systems}, pages
  28--36. ACM, 2001.

\bibitem{Koopman_PNAS31_Koopman}
B.O. Koopman.
\newblock Hamiltonian systems and transformations in hilbert space.
\newblock {\em Proc. Natl. Acad. Sci. USA}, 17:315--318, 1931.

\bibitem{korda2017convergence}
Milan Korda and Igor Mezi{\'c}.
\newblock On convergence of extended dynamic mode decomposition to the
  {K}oopman operator.
\newblock {\em Journal of Nonlinear Science}, 28(2):687--710, 2018.

\bibitem{kurebayashi370optimal}
W.~Kurebayashi, S.~Shirasaka, and H.~Nakao.
\newblock Optimal parameter selection for kernel dynamic mode decomposition.
\newblock In {\em Proc. Int. Symp. NOLTA}, volume 370, page 373.

\bibitem{li2017extended}
Q.~Li, F.~Dietrich, E.~M. Bollt, and I.~G. Kevrekidis.
\newblock Extended dynamic mode decomposition with dictionary learning: A
  data-driven adaptive spectral decomposition of the {K}oopman operator.
\newblock {\em Chaos}, 27(10):103111, 2017.

\bibitem{lusch2018deep}
Bethany Lusch, J~Nathan Kutz, and Steven~L Brunton.
\newblock Deep learning for universal linear embeddings of nonlinear dynamics.
\newblock {\em Nature Communications}, 9(1):4950, 2018.

\bibitem{ma2001composite}
Yue Ma, James~J Han, and Kishor~S Trivedi.
\newblock Composite performance and availability analysis of wireless
  communication networks.
\newblock {\em IEEE Transactions on Vehicular Technology}, 50(5):1216--1223,
  2001.

\bibitem{vampnet}
Andreas Mardt, Luca Pasquali, Hao Wu, and Frank No{\'e}.
\newblock Vampnets for deep learning of molecular kinetics.
\newblock {\em Nature Communications}, 9(1):5, 2018.

\bibitem{marshall1979inequalities}
Albert~W Marshall, Ingram Olkin, and Barry~C Arnold.
\newblock {\em Inequalities: theory of majorization and its applications},
  volume 143.
\newblock Springer, 1979.

\bibitem{McGibbonPande_JCP15_CrossValidation}
R.~T. McGibbon and V.~S. Pande.
\newblock Variational cross-validation of slow dynamical modes in molecular
  kinetics.
\newblock {\em J. Chem. Phys.}, 142:124105, 2015.

\bibitem{Mezic_NonlinDyn05_Koopman}
I.~Mezi\'{c}.
\newblock Spectral properties of dynamical systems, model reduction and
  decompositions.
\newblock {\em Nonlinear Dynam.}, 41:309--325, 2005.

\bibitem{mezic2013analysis}
Igor Mezi{\'c}.
\newblock Analysis of fluid flows via spectral properties of the koopman
  operator.
\newblock {\em Annual Review of Fluid Mechanics}, 45:357--378, 2013.

\bibitem{Molgedey_94}
L.~Molgedey and H.~G. Schuster.
\newblock Separation of a mixture of independent signals using time delayed
  correlations.
\newblock {\em Phys. Rev. Lett.}, 72:3634--3637, 1994.

\bibitem{Noe_JCP08_TSampling}
F.~No{\'e}.
\newblock {Probability Distributions of Molecular Observables computed from
  Markov Models}.
\newblock {\em J. Chem. Phys.}, 128:244103, 2008.

\bibitem{NoeClementi_JCTC15_KineticMap}
F.~No\'{e} and C.~Clementi.
\newblock Kinetic distance and kinetic maps from molecular dynamics simulation.
\newblock {\em J. Chem. Theory Comput.}, 11:5002--5011, 2015.

\bibitem{NoeNueske_MMS13_VariationalApproach}
F.~No{\'e} and F.~N{\"u}ske.
\newblock A variational approach to modeling slow processes in stochastic
  dynamical systems.
\newblock {\em Multiscale Model. Simul.}, 11:635--655, 2013.

\bibitem{NueskeEtAl_JCTC14_Variational}
F.~N{\"u}ske, B.~G. Keller, G.~P{\'e}rez-Hern{\'a}ndez, A.~S. J.~S. Mey, and
  F.~No{\'e}.
\newblock Variational approach to molecular kinetics.
\newblock {\em J. Chem. Theory Comput.}, 10:1739--1752, 2014.

\bibitem{NueskeEtAl_JCP15_Tensor}
F.~N\"uske, R.~Schneider, F.~Vitalini, and F.~No\'{e}.
\newblock Variational tensor approach for approximating the rare-event kinetics
  of macromolecular systems.
\newblock {\em J. Chem. Phys.}, 144:054105, 2016.

\bibitem{otto2019linearly}
Samuel~E Otto and Clarence~W Rowley.
\newblock Linearly recurrent autoencoder networks for learning dynamics.
\newblock {\em SIAM Journal on Applied Dynamical Systems}, 18(1):558--593,
  2019.

\bibitem{Paul2018identification}
F.~Paul, H.~Wu, M.~Vossel, B.~Groot, and F.~Noe.
\newblock Identification of kinetic order parameters for non-equilibrium
  dynamics.
\newblock {\em J. Chem. Phys.}, 2018.
\newblock submitted.

\bibitem{PerezEtAl_JCP13_TICA}
G.~Perez-Hernandez, F.~Paul, T.~Giorgino, G.~{D Fabritiis}, and Frank No{\'e}.
\newblock Identification of slow molecular order parameters for markov model
  construction.
\newblock {\em J. Chem. Phys.}, 139:015102, 2013.

\bibitem{press2007numerical}
William~H Press, Saul~A Teukolsky, William~T Vetterling, and Brian~P Flannery.
\newblock {\em Numerical recipes: The art of scientific computing}.
\newblock Cambridge University Press, 2007.

\bibitem{PrinzEtAl_JCP10_MSM1}
J.-H. Prinz, H.~Wu, M.~Sarich, B.~G. Keller, M.~Senne, M.~Held, J.~D. Chodera,
  C.~Sch{\"u}tte, and F.~No{\'e}.
\newblock Markov models of molecular kinetics: Generation and validation.
\newblock {\em J. Chem. Phys.}, 134:174105, 2011.

\bibitem{renardy2004introduction}
Michael Renardy and Robert~C Rogers.
\newblock {\em An introduction to partial differential equations}.
\newblock Springer, New York, 2004.

\bibitem{RowleyEtAl_JFM09_DMDSpectral}
Clarence~W. Rowley, Igor Mezi{\'{c}}, Shervin Bagheri, Philipp Schlatter, and
  Dan~S. Henningson.
\newblock Spectral analysis of nonlinear flows.
\newblock {\em J. Fluid Mech.}, 641:115, nov 2009.

\bibitem{Schmid_JFM10_DMD}
Peter~J. Schmid.
\newblock Dynamic mode decomposition of numerical and experimental data.
\newblock {\em J. Fluid Mech.}, 656:5--28, jul 2010.

\bibitem{SchuetteFischerHuisingaDeuflhard_JCompPhys151_146}
C.~Sch\"{u}tte, A.~Fischer, W.~Huisinga, and P.~Deuflhard.
\newblock {A Direct Approach to Conformational Dynamics based on Hybrid Monte
  Carlo}.
\newblock {\em J. Comput. Phys.}, 151:146--168, 1999.

\bibitem{SchwantesPande_JCTC13_TICA}
C.~R. Schwantes and V.~S. Pande.
\newblock Improvements in markov state model construction reveal many
  non-native interactions in the folding of ntl9.
\newblock {\em J. Chem. Theory Comput.}, 9:2000--2009, 2013.

\bibitem{SchwantesPande_JCTC15_kTICA}
C.~R. Schwantes and V.~S. Pande.
\newblock Modeling molecular kinetics with tica and the kernel trick.
\newblock {\em J. Chem. Theory Comput.}, 11:600--608, 2015.

\bibitem{snoek2012practical}
Jasper Snoek, Hugo Larochelle, and Ryan~P Adams.
\newblock Practical bayesian optimization of machine learning algorithms.
\newblock In {\em Advances in neural information processing systems}, pages
  2951--2959, 2012.

\bibitem{song2013kernel}
Le~Song, Kenji Fukumizu, and Arthur Gretton.
\newblock Kernel embeddings of conditional distributions: A unified kernel
  framework for nonparametric inference in graphical models.
\newblock {\em IEEE Signal Processing Magazine}, 30(4):98--111, 2013.

\bibitem{sparrow1982lorenz}
Colin Sparrow.
\newblock {\em The {L}orenz equations: bifurcations, chaos, and strange
  attractors}.
\newblock Springer-Verlag, New York, 1982.

\bibitem{takeishi2017learning}
Naoya Takeishi, Yoshinobu Kawahara, and Takehisa Yairi.
\newblock Learning koopman invariant subspaces for dynamic mode decomposition.
\newblock In {\em Advances in Neural Information Processing Systems}, pages
  1130--1140, 2017.

\bibitem{tibshirani1996regression}
Robert Tibshirani.
\newblock Regression shrinkage and selection via the lasso.
\newblock {\em Journal of the Royal Statistical Society. Series B
  (Methodological)}, pages 267--288, 1996.

\bibitem{TuEtAl_JCD14_ExactDMD}
Jonathan~H. Tu, Clarence~W. Rowley, Dirk~M. Luchtenburg, Steven~L. Brunton, and
  J.~Nathan Kutz.
\newblock On dynamic mode decomposition: Theory and applications.
\newblock {\em J. Comput. Dyn.}, 1(2):391--421, dec 2014.

\bibitem{WilliamsKevrekidisRowley_JNS15_EDMD}
M.~O. Williams, I.~G. Kevrekidis, and C.~W. Rowley.
\newblock A data--driven approximation of the koopman operator: Extending
  dynamic mode decomposition.
\newblock {\em J. Nonlinear Sci.}, 25:1307--1346, 2015.

\bibitem{williams2014kernel}
Matthew~O Williams, Clarence~W Rowley, and Ioannis~G Kevrekidis.
\newblock A kernel-based method for data-driven {K}oopman spectral analysis.
\newblock {\em Journal of Computational Dynamics}, 2(2):247--265, 2015.

\bibitem{WuNoe_JCP15_GMTM}
H.~Wu and F.~No\'{e}.
\newblock Gaussian markov transition models of molecular kinetics.
\newblock {\em J. Chem. Phys.}, 142:084104, 2015.

\bibitem{WuEtAl_JCP17_VariationalKoopman}
Hao Wu, Feliks N{\"u}ske, Fabian Paul, Stefan Klus, Peter Koltai, and Frank
  No{\'e}.
\newblock Variational koopman models: slow collective variables and molecular
  kinetics from short off-equilibrium simulations.
\newblock {\em J. Chem. Phys.}, 146:154104, 2017.

\bibitem{ZieheMueller_ICANN98_TDSEP}
Andreas Ziehe and Klaus-Robert M\"{u}ller.
\newblock {TDSEP} {\textemdash} an efficient algorithm for blind separation
  using time structure.
\newblock In {\em {ICANN} 98}, pages 675--680. Springer Science and Business
  Media, 1998.

\end{thebibliography}

\end{document}